\PassOptionsToPackage{dvipsnames}{xcolor}
\documentclass{article}

\usepackage[table]{xcolor}

\usepackage[preprint]{neurips_2025}

\usepackage[utf8]{inputenc} 
\usepackage[T1]{fontenc}    
\usepackage{hyperref}       
\usepackage{url}            
\usepackage{booktabs}       
\usepackage{amsfonts}       
\usepackage{nicefrac}       
\usepackage{microtype}      
\usepackage{xcolor}         

\usepackage{graphicx}
\usepackage{subfig}
\usepackage{amssymb}
\usepackage{amsmath}
\usepackage{amsthm}
\usepackage{multirow}
\usepackage{enumitem}
\usepackage[font=small,labelfont=bf]{caption}
\usepackage{wrapfig}

\usepackage{listings}
\usepackage{kotex}
\usepackage{bm}
\usepackage[dvipsnames]{xcolor}

\title{
MultiTab: A Comprehensive Benchmark Suite \\ for Multi-Dimensional Evaluation in Tabular Domains
}

\author{
\makebox[\textwidth][c]{%
    \begin{tabular}{c}
    Kyungeun Lee, Moonjung Eo, Hye-Seung Cho, Dongmin Kim, \\
    Ye Seul Sim, Seoyoon Kim, Min-Kook Suh, Woohyung Lim \\
    \normalfont LG AI Research \\
    \texttt{\{kyungeun.lee,moonj,hs.cho,dmkim\}@lgresearch.ai}, \\
    \texttt{\{ysl.sim,seoyoon.kim,minkook.suh,w.lim\}@lgresearch.ai}
    \end{tabular}
}}

\begin{document}

\maketitle

\begin{abstract}
    Despite the widespread use of tabular data in real-world applications, most benchmarks rely on average-case metrics, which fail to reveal how model behavior varies across diverse data regimes. To address this, we propose \textsc{MultiTab}, a benchmark suite and evaluation framework for multi-dimensional, data-aware analysis of tabular learning algorithms. Rather than comparing models only in aggregate, \textsc{MultiTab} categorizes 196 publicly available datasets along key data characteristics, including sample size, label imbalance, and feature interaction, and evaluates 13 representative models spanning a range of inductive biases. Our analysis shows that model performance is highly sensitive to such regimes: for example, models using sample-level similarity excel on datasets with large sample sizes or high inter-feature correlation, while models encoding inter-feature dependencies perform best with weakly correlated features. These findings reveal that inductive biases do not always behave as intended, and that regime-aware evaluation is essential for understanding and improving model behavior. \textsc{MultiTab} enables more principled model design and offers practical guidance for selecting models tailored to specific data characteristics. All datasets, code, and optimization logs are publicly available at \url{https://huggingface.co/datasets/LGAI-DILab/Multitab}.
\end{abstract}

\vspace{-0.25cm}
\section{Introduction}
\label{sec:intro}
\vspace{-0.25cm}

Tabular data remains one of the most prevalent formats in real-world machine learning applications, from finance and healthcare to scientific discovery and recommendation systems~\citep{shwartz2022tabular,borisov2022deep,ulmer2020trust, somani2021deep,borisov2021robust,clements2020sequential,guo2017deepfm,zhang2019deep,ahmed2017survey,tang2020customer}. Unlike domains like vision or language, which benefit from strong structural priors and standardized modeling pipelines, tabular learning lacks a unified foundation~\citep{gorishniy2021revisiting,hollmann2022tabpfn,bonet2024hyperfast,van2024tabular}. This is largely due to the heterogeneity of tabular datasets, which vary widely in sample size, feature types, and class imbalance~\citep{zhang2023mixed,liu2024d2r2,cheng2024arithmetic,lee2024binning,eo2025representation}. As a result, no single algorithm consistently outperforms others across tasks, and model performance tends to be highly dependent on the specific data regime.

Despite this diversity, most benchmarks report model rankings using average-case performance~\citep{grinsztajn2022tree,mcelfresh2024neural,salinas2023tabrepo,ye2024closer}. While convenient, such global comparisons obscure when and why specific models succeed or fail. They offer limited insight into the interaction between model design and dataset characteristics, and often reduce nuanced behaviors to single-number summaries. These limitations highlight the need for evaluation frameworks that go beyond global rankings and instead ask: \textit{How do different modeling assumptions affect performance under varying data conditions?}

Recent work has begun to address this question by analyzing the role of individual factors such as sample size or function irregularity. For instance, \citep{grinsztajn2022tree,mcelfresh2024neural} found that gradient-boosted decision trees (GBDTs) often outperform neural networks (NNs) on smaller datasets or when data distributions are inconsistent across features. However, these studies mainly compare broad model families (\emph{e.g.}, GBDTs vs. NNs) rather than examining how specific architectural components contribute to performance in different data regimes. Moreover, previous evaluations~\citep{grinsztajn2022tree,gardner2024benchmarking,wang2024towards} often rely on narrow dataset collections or artificial distribution shifts, limiting the generalizability of their findings.

To address these limitations, we introduce \textsc{MultiTab}, a benchmark suite and evaluation framework designed to support structured, data-aware analysis of tabular learning algorithms. Rather than identifying a single best model, \textsc{MultiTab} aims to reveal how different modeling assumptions perform across well-defined dataset regimes. It includes 196 publicly available datasets covering both classification and regression tasks, and evaluates 13 representative models with diverse inductive biases. Datasets are grouped into sub-categories based on key statistical axes—such as sample size, label imbalance, feature-to-sample ratio, and feature interaction—with multiple complementary metrics used to ensure robust stratification. Each model is trained under consistent cross-validation protocols with extensive hyperparameter optimization.

By evaluating models under interpretable dataset characteristics, \textsc{MultiTab} enables fine-grained comparisons that reveal when and why particular inductive biases, such as those designed to capture inter-feature dependencies or exploit sample-level similarity, lead to performance gains or fail to do so. Based on our benchmark, we found that model performance varies significantly depending on the dataset regime. For example, models that encode inter-sample dependencies excel when the sample size is large or numerical features dominate, while models that encode inter-feature dependencies are more robust when features are weakly correlated. Tree-based models, in contrast, remain strong on regression tasks and under low class imbalance. These patterns are not visible under average-case evaluation, highlighting the need for conditional, data-aware analysis.

More than a benchmark, \textsc{MultiTab} serves as a diagnostic tool for identifying when modeling assumptions succeed or fail under specific data regimes. By identifying which modeling assumptions work best under particular conditions, \textsc{MultiTab} supports more effective model selection and informs the design of future tabular architectures better suited to real-world data. Full benchmark suite and logs are publicly available at \url{https://huggingface.co/datasets/LGAI-DILab/Multitab}.

\vspace{-0.3cm}
\section{Previous Tabular Benchmarks}
\label{sec:previous}
\vspace{-0.3cm}

Tabular data remains central to many applications, yet the debate continues over deep learning versus gradient-boosted decision trees (GBDTs)~\citep{gorishniy2021revisiting, grinsztajn2022tree, mcelfresh2024neural}. While new neural architectures emerge~\citep{gorishniy2022embeddings, yan2023t2g}, they are often evaluated on narrow benchmarks using average-case metrics, limiting insights into when and why models succeed.

Existing benchmarks typically compare broad model classes rather than specific architectural components. \citep{grinsztajn2022tree} evaluated GBDTs and NNs on 45 datasets but relied on synthetic transformations and average metrics. TabZilla~\citep{mcelfresh2024neural} expanded to 176 datasets, finding no dominant model, but their rank-based approach reduced complex behaviors to single numbers and focused mainly on classification. TabRepo~\citep{salinas2023tabrepo} provided 1,300 model predictions for ensemble analysis but lacked structured evaluation of architectural components. Other studies highlighted preprocessing impact~\citep{tschalzev2024data} or identified influential dataset features~\citep{ye2024closer, shmuel2024comprehensive}, yet still relied on global performance metrics.
Recent benchmarks have explored distributional robustness from different angles: TableShift~\citep{gardner2024benchmarking} examined how models handle label distribution shifts, TabReD~\citep{rubachev2024tabred} focused on temporal generalization in industrial settings, and \citep{kohli2024towards} revealed that many tabular datasets are outdated for modern challenges. While these studies address important real-world deployment concerns, they do not comprehensively analyze how different modeling assumptions perform across diverse data conditions.

Previous studies reveal several limitations: lack of principled dataset categorization, limited regression coverage, minimal analysis of architectural components, and absence of conditional evaluation beyond average rankings. \textsc{MultiTab} addresses these gaps by organizing diverse datasets and models along key axes—such as sample size, label imbalance, and feature interaction—to support hypothesis-driven analysis of how modeling assumptions interact with data characteristics.

\vspace{-0.2cm}
\section{\textsc{MultiTab}: Tabular Benchmark Suite for Multi-Dimensional Evaluation}
\label{sec:benchmark}
\vspace{-0.2cm}

In this study, we introduce \textsc{MultiTab}, a benchmark suite designed to support multi-dimensional, data-aware evaluation of tabular learning algorithms. 
To capture the diversity inherent to tabular data, we define a set of sub-categories based on quantitative dataset statistics that reflect key properties such as sample size and label imbalance. 
Detailed descriptions for the benchmark suite are available in Appendix~\ref{appendix:benchmark}.

\vspace{-0.2cm}
\subsection{Data Collection and Preprocessing}
\label{subsec:benchmark-1}
\vspace{-0.2cm}

We construct a benchmark suite based on the following criteria: (1) prior use in existing benchmarks or tabular deep learning studies~\citep{grinsztajn2022tree,mcelfresh2024neural,gorishniy2021revisiting,lee2024binning,eo2025representation}, (2) public availability and structured metadata from OpenML~\citep{OpenMLPython2019} or scikit-learn~\citep{scikit-learn}, and (3) quality control, excluding datasets with unclear licensing, missing metadata, or inconsistent column mappings.
To mitigate selection bias, we ensure diversity across task types, dataset sizes, and feature dimensionalities. The final benchmark includes 196 datasets with over 10 million samples.

Missing values are addressed by removing columns with more than 50\% missing entries and discarding samples with missing values in inputs or labels. We determine feature types based on available data descriptions and apply quantile transformation to all numerical features~\citep{gorishniy2021revisiting,gorishniy2022embeddings,grinsztajn2022tree,de2023choice}.
Detailed descriptions are provided in Appendix~\ref{appendix:benchmark}.

\vspace{-0.2cm}
\subsection{Algorithms}
\label{subsec:benchmark-2}
\vspace{-0.2cm}


We evaluate 13 representative models spanning diverse architectural assumptions relevant to tabular prediction: (1) \textbf{Classical method:} random forest~\citep{breiman2001random}, a popular tree-based model often used as a baseline in prior studies; (2) \textbf{Gradient-Boosted Decision Trees (GBDTs):} XGBoost~\citep{chen2016xgboost}, CatBoost~\citep{prokhorenkova2018catboost}, and LightGBM~\citep{ke2017lightgbm}, widely used for their effectiveness on tabular tasks and diverse optimization strategies; (3) \textbf{NNs without explicit structural priors (NN-Simple):} MLP, MLP-C (MLP with categorical embedding modules), MLP-CN (MLP with categorical and numerical embedding modules)~\citep{gorishniy2022embeddings}, and ResNet~\citep{gorishniy2021revisiting}, which do not explicitly model interactions between features or between samples; (4) \textbf{NNs modeling inter-feature dependency (NN-Feature):} FT-Transformer~\citep{gorishniy2021revisiting} and T2G-Former~\citep{yan2023t2g}, both of which use attention to capture interactions across input features; (5) \textbf{NNs modeling inter-sample dependency (NN-Sample):} TabR~\citep{gorishniy2023tabr} and ModernNCA~\citep{ye2024modern}, which learn inter-sample relationships through retrieval or metric learning objectives; (6) \textbf{NNs modeling both inter-feature and inter-sample dependency (NN-Both):} SAINT~\citep{somepalli2021saint}, which combines row-wise and column-wise attention to capture complex dependencies across both dimensions.
These categories reflect core inductive biases commonly used in recent tabular modeling, and allow us to analyze how structural assumptions interact with different data regimes. In this study, we exclude AutoML frameworks~\citep{erickson2020autogluon,le2020scaling,salinas2023tabrepo} to focus on individual model behavior under controlled evaluation protocols.

\vspace{-0.2cm}
\subsection{Training Protocols}
\label{subsec:benchmark-3}
\vspace{-0.2cm}

To ensure fair and robust evaluation, we apply a standardized training pipeline across all algorithms and datasets. We evaluate each algorithm using stratified $k$-fold cross-validation, with independent hyperparameter optimization performed on each fold. 

\vspace{-0.2cm}
\paragraph{Hyperparameter Optimization.}
For every algorithm-dataset pair, we perform 100 optimization trials using the Tree-structured Parzen Estimator (TPE)~\citep{bergstra2011algorithms} as implemented in the Optuna library~\citep{optuna2019}. Unlike previous benchmarks that rely on random search with time limits~\citep{grinsztajn2022tree, mcelfresh2024neural}, we impose no time constraints and adopt a consistent trial budget to ensure equal optimization effort across all models.
The optimization objective is task-dependent: validation accuracy for classification and validation RMSE for regression. 
Our search space is adapted from prior work~\citep{gorishniy2021revisiting,gorishniy2022embeddings,yan2023t2g,ye2024modern} and expanded to cover additional architectural and optimization components. 
Detailed hyperparameter search space is provided in Appendix~\ref{appendix:algorithms}, and complete optimization logs are available in the suite. 

\vspace{-0.2cm}
\paragraph{Cross-validation.}
For each $k$-fold split, we perform hyperparameter optimization independently and retrain the model using the best configuration found on that fold's validation split. For datasets with fewer than 50,000 samples, we use 10-fold cross-validation; for larger datasets, we use 3-fold to reduce computational cost. This approach ensures that each fold reflects the best possible performance within its own training context, balancing evaluation rigor with computational feasibility. Final results are obtained by averaging predictive error across folds for each model-dataset pair.

\vspace{-0.2cm}
\subsection{Evaluation Metrics}
\label{subsec:benchmark-4}
\vspace{-0.2cm}

We use \textit{normalized predictive error} as the primary evaluation metric to ensure fair comparisons across datasets with varying scales and difficulty levels. For classification tasks, we use log loss; for regression tasks, we use root mean squared error (RMSE).

To account for dataset-specific difficulty and variance, each model’s error is linearly scaled within each dataset-split pair between the best and worst error values. This yields a normalized score in $[0, 1]$, with 0 corresponding to the best model and 1 to the worst. Final results are computed by averaging these normalized errors across cross-validation folds. This approach emphasizes relative performance rather than raw error magnitudes, enabling robust comparisons across heterogeneous datasets.
In addition to our primary metric, in the suite, we also report the average rank based on raw error, averaged across all splits and seeds.

We select log loss for classification because it is sensitive to class imbalance and remains meaningful even when all labels belong to a single class. In contrast, accuracy fails to reflect class distribution, and AUROC becomes undefined in degenerate cases. Log loss provides a more stable and informative metric across diverse scenarios. Formal metric definitions are provided in Appendix~\ref{appendix:benchmark}.

\vspace{-0.2cm}
\subsection{Sub-category Construction}
\label{subsec:benchmark-5}
\vspace{-0.2cm}

\begin{table}[tb!]
\centering
    \caption{Overview of \textsc{MultiTab}: Datasets are grouped into sub-categories according to the criteria in the table. See Appendix~\ref{appendix:criteria} for full implementation details.}
    \label{tab:suite}
    \resizebox{\textwidth}{!}{
    \begin{tabular}{lll}
    \toprule
    Criterion & Metrics used to classify sub-categories & Sub-categories \\\midrule\midrule

    Task types & Task types & \begin{tabular}[c]{@{}l@{}}Binary classification (68 datasets)\\ Multiclass classification (60 datasets)\\ Regression (68 datasets)\end{tabular} \\\midrule

    Sample size & Number of samples after preprocessing & \begin{tabular}[c]{@{}l@{}}Small: Fewer than 1k samples (62 datasets)\\ Large: More than 10k samples (68 datasets)\end{tabular} \\\midrule

    \multirow{2}{*}{Feature heterogeneity} & \begin{tabular}[c]{@{}l@{}}The ratio of categorical features \\ to total features within each dataset\end{tabular} & \begin{tabular}[c]{@{}l@{}}Few: Smaller than 20\% (123 datasets) \\ Many: Greater than 60\% (33 datasets) \end{tabular} \\\cmidrule(lr){2-3}
     & Average cardinality of categorical features & \begin{tabular}[c]{@{}l@{}}Low: Average cardinality is smaller than 3 (33 datasets)\\ High: Average cardinality is larger than 10 (21 datasets) \end{tabular} \\\midrule
    
    Feature-to-sample ratio & \begin{tabular}[c]{@{}l@{}} The ratio of the feature dimensionality \\ to the sample size \end{tabular} & \begin{tabular}[c]{@{}l@{}}Low: Feature-to-sample ratio is smaller than 0.002 (64 datasets)\\ High: Feature-to-sample ratio is larger than 0.02 (61 datasets) \end{tabular} \\\midrule
    
    \multirow{3}{*}{Label imbalance} & \begin{tabular}[c]{@{}l@{}}Entropy ratio for the classification task, $\frac{\text{Entropy}(\mathbf{y})}{\log{|\mathbf{y}|}}$ \end{tabular} & \begin{tabular}[c]{@{}l@{}}Balanced: Entropy ratio is greater than 0.7 (43 datasets) \\ Imbalanced: Entropy ratio is smaller than 0.3 (61 datasets) \end{tabular} \\\cmidrule(lr){2-3}

    & Skewness for the regression task & \begin{tabular}[c]{@{}l@{}}
    Balanced: Absolute skewness smaller than 0.7 (27 datasets) \\ Imbalanced: Absolute skewness greater than 1.5 (24 datasets) \\ \end{tabular} \\\cmidrule(lr){2-3}

     & Imbalance factor~\citep{wang2024towards} & \begin{tabular}[c]{@{}l@{}}Balanced: Imbalance factor is smaller than 3 (74 datasets) \\ Imbalanced: Imbalance factor is greater than 5 (95 datasets) \end{tabular} \\
    \midrule
    
    Function irregularity & \begin{tabular}[c]{@{}l@{}} 
    Ratio of energy in high-frequency components \\ to total energy of the function~\citep{beyazit2024inductive} \end{tabular} & \begin{tabular}[c]{@{}l@{}}
    Regular: High-frequency ratio is smaller than 0.25 (38 datasets) \\ Irregular: High-frequency ratio is greater than 0.95 (43 datasets) \end{tabular} \\\midrule
    
    \multirow{2}{*}{Feature interaction} & 
    \begin{tabular}[c]{@{}l@{}}
    Average Frobenius norm of correlation matrix \\ between features
    \end{tabular}
    & \begin{tabular}[c]{@{}l@{}}Correlated: Average norm is greater than 0.03 (59 datasets)\\ Uncorrelated: Average norm is smaller than 0.005 (39 datasets) \end{tabular} \\\cmidrule(lr){2-3}

     & Minimum eigenvalue of normalized covariance matrix & \begin{tabular}[c]{@{}l@{}}Correlated: Minimum eigenvalue is smaller than 0.002 (56 datasets) \\ Uncorrelated: Minimum eigenvalue is greater than 0.1 (74 datasets) \end{tabular} \\
    
    \bottomrule     
    \end{tabular}
}
\end{table}

To enable structured analysis of model behavior, we define sub-categories based on well-defined data statistics. Unlike vision or language domains with consistent input structure, tabular datasets vary widely in sample size, feature composition, and label distributions. This diversity limits the utility of global performance averages, which often mask model-specific strengths. Sub-categorization allows us to isolate such effects and assess model performance under distinct data regimes.
In this study, we define seven core axes that frequently affect tabular learning performance, as follows:

\vspace{-0.2cm}
\begin{itemize}[leftmargin=*]
    \item \textbf{Task type:} The prediction objective of the dataset, reflecting basic differences in label structure that influence how models learn and generalize.
    \item \textbf{Sample size:} Total number of samples in the dataset, reflecting differences in data availability that can affect optimization stability and overfitting risk.
    \item \textbf{Feature heterogeneity:} Degree of variation in input types (\emph{i.e.}, numerical vs. categorical) and granularity of categorical features, which may affect model encoding strategies.
    \item \textbf{Feature-to-sample ratio:} Ratio of input dimensionality to the number of data samples. High values are associated with increased risk of overfitting and challenges related to the curse of dimensionality, making this a key factor in tabular model design.
    \item \textbf{Label imbalance:} Degree of skew in label distributions, which may impact model calibration and learning dynamics.
    \item \textbf{Function irregularity:} Degree of inconsistency in how variations in input features translate into target value changes, ranging from negligible to abrupt.
    \item \textbf{Feature interaction:} Strength of statistical dependence among input features, reflecting how strongly feature values correlate or encode redundant information.
\end{itemize}
\vspace{-0.2cm}

Each axis is translated into a quantitative metric, and datasets are partitioned into non-overlapping groups to support stratified comparisons as summarized in Table~\ref{tab:suite}. Full details are provided in Appendix~\ref{appendix:criteria}.
The benchmark also includes additional dataset properties, including the number of classes and non-linearity favorability, to support broader future analyses.

This sub-categorization framework offers several benefits. First, grounding each axis in interpretable metrics enables reproducible, hypothesis-driven evaluation of model behavior under well-defined data regimes. Second, several axes are measured using complementary definitions (\emph{e.g.}, label imbalance via both imbalance factor and entropy ratio), which supports consistent trend analysis across alternative formulations. Finally, grouping datasets into discrete regimes—rather than correlating continuous statistics with performance—reduces sensitivity to noise and supports clearer, more stable comparisons of inductive biases.
 \vspace{-0.3cm}
\section{Empirical Findings on MultiTab Benchmark}
\label{sec:findings}
\vspace{-0.3cm}

In this section, we examine how model behavior varies with key dataset characteristics in tabular learning, moving beyond aggregate metrics. Using our benchmark suite, we compare performance across distinct data regimes to identify when specific architectural choices offer advantages, enabling a more nuanced understanding of algorithmic performance. (All experiments used a single RTX 3090 GPU; additional results and cost analysis are available in Appendix~\ref{appendix:additionalresults}.)

\begin{figure}[tb!]
    \centering
    \includegraphics[width=\linewidth]{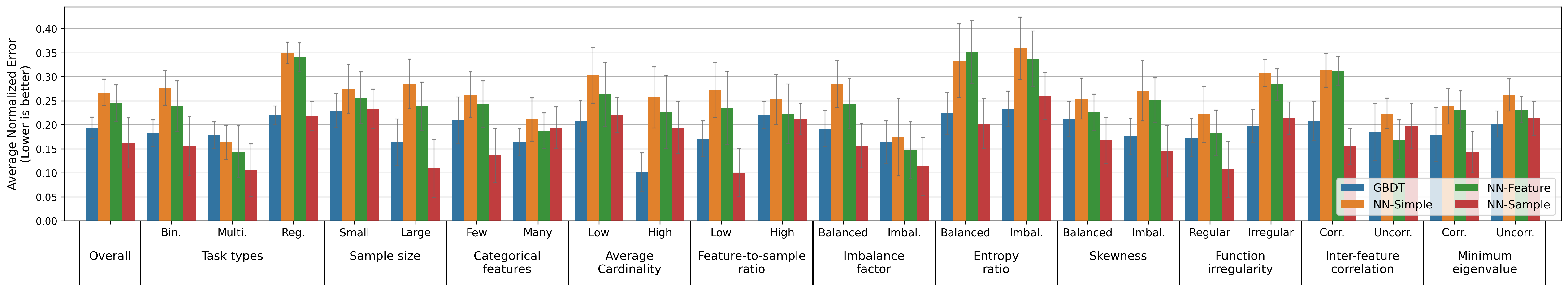}
    \caption{Average normalized predictive error across 24 sub-categories for four model classes: GBDTs, NN-Simple, NN-Sample, and NN-Feature. Lower values indicate better performance. Error bars represent 95\% confidence intervals. Unlike overall averages, model rankings vary substantially across different data regimes, highlighting the importance of conditional evaluation.}
    \vspace{-0.5cm}
    \label{fig:overview}
\end{figure}

\vspace{-0.2cm}
\subsection{Model-class-level Performance Analysis}
\vspace{-0.2cm}

We analyze performance at the model class level, grouping models into six categories based on their inductive biases over inter-feature and inter-sample dependencies as described in Section~\ref{subsec:benchmark-2}. For clarity, we focus on four strong-performing classes as summarized in Figure~\ref{fig:overview}.

On average, NN-Sample achieves the lowest normalized error, followed closely by GBDTs and NN-Feature within the confidence interval. While this suggests strong overall performance, such averages are inherently limited and may not reflect deeper trends.
For instance, while GBDTs underperform in multiclass classification, they remain dominant in datasets with high-cardinality categorical features, where their discrete splitting mechanisms are especially effective. NN-Sample performs best when sample sizes are large or feature-to-sample ratios are low. This aligns with its design, which relies on inter-sample similarity and thus benefits from larger sample sizes. In contrast, NN-Feature performs best when inter-feature correlation is low, as its attention mechanisms can more effectively identify informative features without being constrained by strong linear dependencies.

These observations show that model performance is tightly linked to dataset characteristics, with different inductive biases succeeding under different conditions. Sub-category-level analysis allows us to identify when these biases help or hinder performance, offering clearer guidance for model design and selection than aggregate metrics alone.


\vspace{-0.2cm}
\subsection{Model-level Performance Analysis}
\vspace{-0.2cm}

\begin{table}[tb!]
    \centering
    \caption{
    Average normalized predictive error across 24 data sub-categories. 
    For each sub-category, the best-performing model is shown in \textbf{bold}. 
    Models within the 95\% confidence interval of the best are highlighted in \colorbox{blue!25}{blue}. Best-performing models vary across sub-categories, reflecting sensitivity to data regimes.
    }
    \label{tab:results}
    \resizebox{\textwidth}{!}{
    \begin{tabular}{lccccccccccccccc}
    \toprule
    Criterion & Overall & \multicolumn{3}{c}{Task types} & \multicolumn{2}{c}{Sample size} & \multicolumn{2}{c}{Categorical features} & \multicolumn{2}{c}{Average cardinality} & \multicolumn{2}{c}{Feature-to-sample ratio} & \multicolumn{2}{c}{Entropy ratio} \\\cmidrule(lr){1-1}\cmidrule(lr){2-2}\cmidrule(lr){3-5}\cmidrule(lr){6-7}\cmidrule(lr){8-9}\cmidrule(lr){10-11}\cmidrule(lr){12-13}\cmidrule(lr){14-15}
    Sub-category & - & Bin & Multi & Reg & Small & Large & Few & Many & Low & High & Low & High & Balanced & Imbalanced \\\midrule\midrule
    
    Random Forest & 0.415 & 0.377 & 0.398 & 0.467 & \cellcolor{blue!25}0.327 & 0.536 & 0.447 & 0.352 & \cellcolor{blue!25}0.379 & \cellcolor{blue!25}0.316 & 0.541 & \cellcolor{blue!25}0.315 & 0.405 & 0.357 \\\midrule
    XGBoost & \cellcolor{blue!25}0.261 & \cellcolor{blue!25}\textbf{0.196} & 0.230 & 0.353 & \cellcolor{blue!25}0.320 & \cellcolor{blue!25}\textbf{0.211} & 0.272 & \cellcolor{blue!25}0.215 & \cellcolor{blue!25}\textbf{0.275} & \cellcolor{blue!25}\textbf{0.180} & \cellcolor{blue!25}\textbf{0.224} & \cellcolor{blue!25}0.278 & \cellcolor{blue!25}\textbf{0.204} & \cellcolor{blue!25}0.218 \\ 
    CatBoost & 0.460 & 0.522 & 0.643 & \cellcolor{blue!25}\textbf{0.238} & 0.557 & 0.329 & 0.502 & 0.452 & \cellcolor{blue!25}0.385 & \cellcolor{blue!25}0.307 & \cellcolor{blue!25}0.320 & 0.562 & 0.515 & 0.596 \\ 
    LightGBM & 0.304 & 0.276 & 0.277 & 0.357 & \cellcolor{blue!25}0.320 & 0.303 & 0.312 & \cellcolor{blue!25}0.292 & \cellcolor{blue!25}0.333 & \cellcolor{blue!25}0.236 & 0.318 & \cellcolor{blue!25}0.303 & 0.293 & \cellcolor{blue!25}0.266 \\\midrule
    MLP & 0.409 & 0.388 & 0.276 & 0.549 & 0.370 & 0.465 & 0.376 & 0.397 & \cellcolor{blue!25}0.427 & 0.600 & 0.436 & 0.376 & 0.401 & 0.302 \\ 
    MLP-C & 0.366 & 0.354 & 0.252 & 0.478 & \cellcolor{blue!25}0.346 & 0.407 & 0.366 & \cellcolor{blue!25}0.290 & \cellcolor{blue!25}0.395 & \cellcolor{blue!25}0.354 & 0.382 & 0.350 & 0.367 & 0.267 \\ 
    MLP-CN & 0.342 & 0.347 & 0.229 & 0.436 & \cellcolor{blue!25}0.354 & 0.325 & 0.336 & \cellcolor{blue!25}0.280 & \cellcolor{blue!25}0.360 & \cellcolor{blue!25}0.331 & \cellcolor{blue!25}0.311 & 0.354 & 0.349 & \cellcolor{blue!25}0.221 \\ 
    ResNet & 0.370 & 0.364 & 0.249 & 0.481 & 0.373 & 0.403 & 0.360 & \cellcolor{blue!25}0.298 & \cellcolor{blue!25}0.410 & \cellcolor{blue!25}0.389 & 0.383 & 0.352 & 0.369 & 0.275 \\\midrule
    FT-Transformer & \cellcolor{blue!25}0.274 & 0.265 & \cellcolor{blue!25}0.167 & 0.377 & \cellcolor{blue!25}\textbf{0.278} & \cellcolor{blue!25}0.261 & 0.266 & \cellcolor{blue!25}0.240 & \cellcolor{blue!25}0.306 & \cellcolor{blue!25}0.251 & \cellcolor{blue!25}0.255 & \cellcolor{blue!25}0.257 & \cellcolor{blue!25}0.268 & \cellcolor{blue!25}0.165 \\ 
    T2G-Former & \cellcolor{blue!25}0.271 & 0.263 & \cellcolor{blue!25}0.161 & 0.377 & \cellcolor{blue!25}0.288 & \cellcolor{blue!25}0.261 & 0.268 & \cellcolor{blue!25}\textbf{0.205} & \cellcolor{blue!25}0.294 & \cellcolor{blue!25}0.260 & \cellcolor{blue!25}0.259 & \cellcolor{blue!25}\textbf{0.246} & \cellcolor{blue!25}0.269 & \cellcolor{blue!25}\textbf{0.164} \\\midrule 
    TabR & \cellcolor{blue!25}0.283 & 0.332 & \cellcolor{blue!25}\textbf{0.158} & 0.343 & \cellcolor{blue!25}0.302 & \cellcolor{blue!25}0.224 & 0.258 & 0.313 & \cellcolor{blue!25}0.364 & \cellcolor{blue!25}0.290 & \cellcolor{blue!25}\textbf{0.224} & \cellcolor{blue!25}0.294 & 0.336 & \cellcolor{blue!25}0.166 \\ 
    ModernNCA & \cellcolor{blue!25}\textbf{0.240} & \cellcolor{blue!25}0.231 & \cellcolor{blue!25}0.213 & \cellcolor{blue!25}0.274 & \cellcolor{blue!25}0.313 & \cellcolor{blue!25}0.234 & \cellcolor{blue!25}\textbf{0.201} & 0.325 & \cellcolor{blue!25}0.349 & \cellcolor{blue!25}0.264 & \cellcolor{blue!25}0.231 & \cellcolor{blue!25}0.287 & \cellcolor{blue!25}0.240 & \cellcolor{blue!25}0.229 \\\midrule
    SAINT & 0.359 & 0.381 & 0.243 & 0.441 & \cellcolor{blue!25}0.343 & 0.374 & 0.350 & 0.305 & \cellcolor{blue!25}0.400 & \cellcolor{blue!25}0.290 & 0.368 & 0.356 & 0.390 & \cellcolor{blue!25}0.251 \\ 
    \bottomrule
    \end{tabular}
    } \\
    \resizebox{\textwidth}{!}{
    \begin{tabular}{lcccccccccc}
    \toprule
    Criterion 
    & \multicolumn{2}{c}{Skewness} & \multicolumn{2}{c}{Imbalance factor} & \multicolumn{2}{c}{Function irregularity} & \multicolumn{2}{c}{Inter-feature correlation} & \multicolumn{2}{c}{Minimum eigenvalue} 
    \\\cmidrule(lr){1-1}\cmidrule(lr){2-3}\cmidrule(lr){4-5}\cmidrule(lr){6-7}\cmidrule(lr){8-9}\cmidrule(lr){10-11}
    Sub-category & Balanced & Imbalanced & Balanced & Imbalanced & Regular & Irregular & Correlated & Uncorrelated & Correlated & Uncorrelated \\\midrule\midrule
    Random Forest & 0.474 & 0.421 & 0.429 & 0.422 & 0.573 & 0.388 & 0.400 & 0.459 & 0.365 & 0.390 \\\midrule
    
    XGBoost & 0.396 & 0.336 & \cellcolor{blue!25}0.283 & \cellcolor{blue!25}0.246 & \cellcolor{blue!25}0.199 & \cellcolor{blue!25}0.309 & \cellcolor{blue!25}0.303 & \cellcolor{blue!25}0.249 & \cellcolor{blue!25}\textbf{0.215} & \cellcolor{blue!25}0.303 \\
    CatBoost & \cellcolor{blue!25}\textbf{0.250} & \cellcolor{blue!25}\textbf{0.245} & 0.496 & 0.426 & 0.406 & \cellcolor{blue!25}0.366 & 0.438 & 0.478 & 0.534 & 0.413 \\
    LightGBM & 0.389 & 0.344 & 0.322 & 0.292 & \cellcolor{blue!25}0.312 & \cellcolor{blue!25}0.341 & 0.325 & 0.309 & \cellcolor{blue!25}0.284 & \cellcolor{blue!25}0.303 \\\midrule

    MLP & 0.519 & 0.594 & 0.384 & 0.415 & 0.395 & 0.497 & 0.403 & 0.399 & 0.375 & 0.402 \\
    MLP-C & 0.442 & 0.489 & 0.331 & 0.382 & 0.352 & 0.419 & 0.407 & 0.335 & 0.317 & 0.350 \\
    MLP-CN & 0.460 & 0.420 & 0.334 & 0.338 & \cellcolor{blue!25}0.264 & 0.391 & 0.405 & 0.292 & \cellcolor{blue!25}0.301 & 0.338 \\
    ResNet & 0.457 & 0.493 & 0.334 & 0.388 & 0.359 & 0.426 & 0.411 & 0.316 & 0.314 & 0.392 \\\midrule
    
    FT-Transformer & \cellcolor{blue!25}0.365 & 0.395 & \cellcolor{blue!25}0.249 & \cellcolor{blue!25}0.285 & \cellcolor{blue!25}0.205 & \cellcolor{blue!25}0.317 & 0.334 & \cellcolor{blue!25}0.225 & \cellcolor{blue!25}0.257 & \cellcolor{blue!25}0.260 \\
    T2G-Former & 0.392 & 0.367 & \cellcolor{blue!25}0.252 & \cellcolor{blue!25}0.279 & \cellcolor{blue!25}0.201 & \cellcolor{blue!25}0.324 & 0.347 & \cellcolor{blue!25}\textbf{0.182} & \cellcolor{blue!25}0.252 & \cellcolor{blue!25}\textbf{0.258} \\\midrule
    
    TabR & \cellcolor{blue!25}0.349 & 0.365 & 0.319 & \cellcolor{blue!25}0.252 & \cellcolor{blue!25}0.262 & \cellcolor{blue!25}0.303 & \cellcolor{blue!25}0.284 & 0.299 & 0.308 & \cellcolor{blue!25}0.290 \\
    ModernNCA & \cellcolor{blue!25}0.255 & \cellcolor{blue!25}0.302 & \cellcolor{blue!25}\textbf{0.244} & \cellcolor{blue!25}\textbf{0.221} & \cellcolor{blue!25}\textbf{0.194} & \cellcolor{blue!25}\textbf{0.272} & \cellcolor{blue!25}\textbf{0.234} & 0.289 & \cellcolor{blue!25}0.225 & \cellcolor{blue!25}0.326 \\\midrule
    
    SAINT & 0.460 & 0.417 & 0.360 & 0.362 & 0.360 & \cellcolor{blue!25}0.366 & 0.399 & 0.289 & 0.337 & 0.343 \\
    \bottomrule
    \end{tabular}
    }
    \vspace{-0.5cm}
\end{table}

\begin{figure}[t!]
    \centering
    \includegraphics[width=\textwidth]{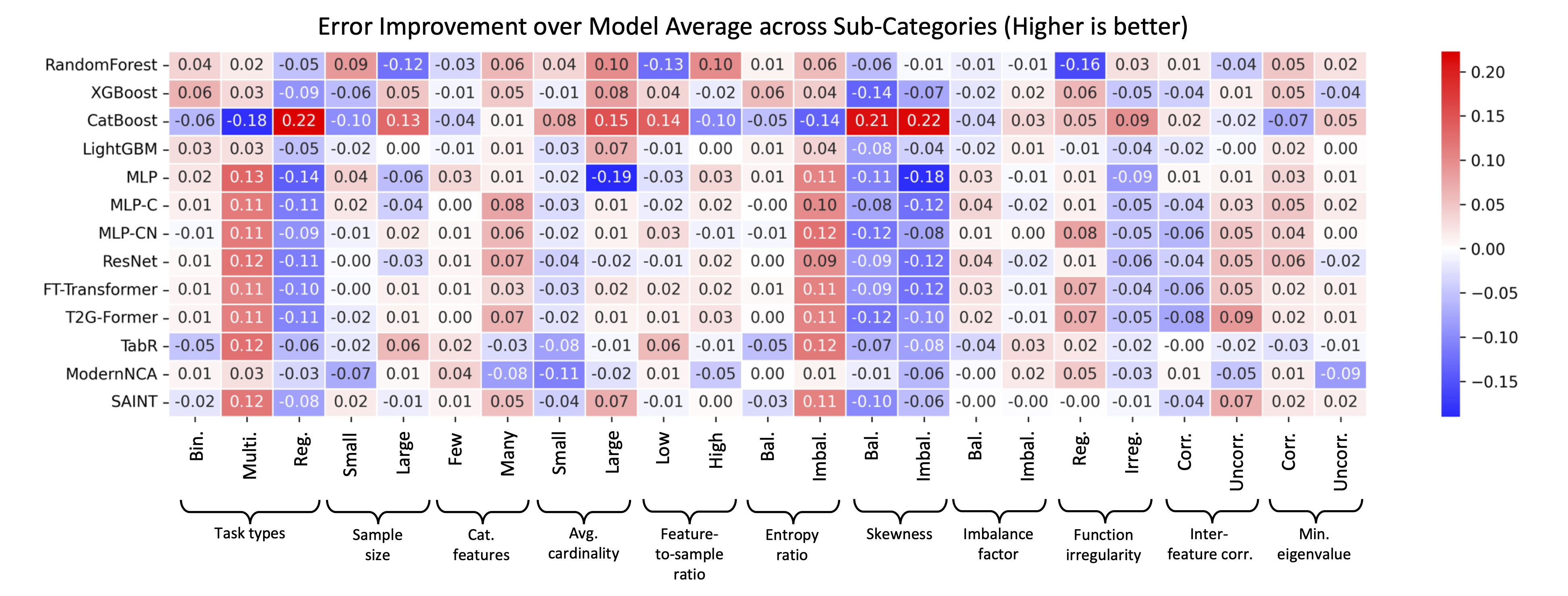}
    \vspace{-0.6cm}
    \caption{
    Deviation from overall average error across 24 sub-categories. 
    Each cell shows the difference between a model’s average error in a given sub-category and its overall mean across all datasets. 
    Blue (negative) indicates worse-than-average performance; red (positive) indicates better-than-average. 
    This highlights model-specific strengths and weaknesses relative to their overall behavior.
    }
    \label{fig:heatmap}
    \vspace{-0.6cm}
\end{figure}

We now examine performance at the level of individual algorithms across specific dataset conditions using two complementary views. Table~\ref{tab:results} provides absolute performance comparisons by reporting the average normalized error per model across 24 sub-categories, highlighting the best and statistically comparable models in each. Figure~\ref{fig:heatmap}, in contrast, shows how each model's performance deviates from its overall average, revealing which data regimes amplify or suppress a model’s relative strength.

As shown in Table~\ref{tab:results}, ModernNCA achieves the best average normalized error overall, while models like XGBoost, FT-Transformer, T2G-Former, and TabR also perform competitively within the confidence interval. These results align with prior studies~\citep{grinsztajn2022tree,mcelfresh2024neural,ye2024closer} highlighting GBDTs as strong baselines and recent neural models that close the gap through inductive bias design.

Not all neural architectures, however, meet expectations. ResNet, favoring training stability over structural adaptation, underperforms compared to simpler variants like MLP-C. Similarly, SAINT, which applies both row-wise and column-wise attention to capture complex dependencies, often lags behind models specialized for either inter-feature or inter-sample relationships. These observations highlight that model robustness depends less on architectural complexity and more on how well design choices align with the underlying data characteristics. This reinforces the need for fine-grained, regime-aware evaluation, as provided by our sub-category-based analysis.

\vspace{-0.2cm}
\subsubsection{Task types: ModernNCA is the only neural model robust across all task types}
\vspace{-0.2cm}

Only ModernNCA consistently ranks among the top-performing models across binary classification, multiclass classification, and regression tasks. This aligns with prior findings~\citep{ye2024closer}, which attribute its strength to a metric-learning-based latent space that reflects target similarity, enabling smooth generalization over both discrete and continuous outputs.

Multiclass classification, on the other hand, appears to favor neural models more broadly. Previous benchmarks~\citep{ye2024closer,chen2023excelformer} report that attention-based architectures such as FT-Transformer perform competitively in these tasks. Our results support this trend, suggesting that neural networks may be better suited to tasks involving more complex or higher-cardinality label spaces.

In summary, while neural networks have become increasingly competitive in classification tasks, their general robustness remains limited. Only ModernNCA achieves consistent performance across all task types, suggesting that most neural architectures still lack universal reliability in tabular settings.

\vspace{-0.2cm}
\subsubsection{Sample size: It alone does not determine model superiority in tabular learning}
\vspace{-0.2cm}

Model performance exhibits distinct patterns depending on sample size. In small-sample regimes, most algorithms perform similarly within overlapping confidence intervals. Surprisingly, high-capacity neural networks like FT-Transformer and ModernNCA remain competitive, with FT-Transformer achieving the best mean performance, despite the limited data. These findings challenge the common belief that neural networks require large datasets to be effective, and emphasize the importance of using statistical testing rather than relying solely on mean scores.

In large-sample regimes, performance gaps between models become more pronounced. XGBoost achieves the best average performance, but several neural models, including TabR, ModernNCA, FT-Transformer, and T2G-Former, deliver comparable results within the confidence interval. This contrasts with prior studies~\citep{mcelfresh2024neural,grinsztajn2022tree} that found neural networks to underperform as sample size grows. Our results indicate that modern neural models have closed this gap in practice. More recent meta-analyses~\citep{shmuel2024comprehensive} support this view, suggesting that sample size interacts with other factors, such as feature dimensionality, rather than being a standalone predictor of model success.

Taken together, these results suggest that while data availability is important, it does not dictate performance in isolation. Neural networks with appropriate inductive biases and regularization strategies can perform robustly across both small and large datasets.

\vspace{-0.2cm}
\subsubsection{Feature heterogeneity: Categorical embedding modules are effective when categorical features dominate}
\vspace{-0.2cm}

Tabular datasets often contain both numerical and categorical features, with the latter posing challenges due to their discrete and high-cardinality nature. Tree-based models handle them via discrete splits or target encoding, while neural networks typically rely on learned embeddings. Some variants, such as MLP-CN~\citep{gorishniy2022embeddings}, extend embeddings to numerical inputs as well.

We analyze model behavior along two axes: the proportion of categorical features and the average cardinality of categorical variables. When categorical features are few, ModernNCA performs best, benefiting from metric learning over continuous inputs. However, its performance drops sharply when categorical features dominate (see Figure~\ref{fig:heatmap}). In contrast, NN-Feature and GBDTs remain competitive, likely due to their reduced sensitivity to embedding calibration.

To isolate the effect of categorical embeddings, we compare MLP (without embeddings) to MLP-C (with embeddings). Their performance is nearly identical when categorical features are scarce (0.376 vs. 0.366), but diverges significantly when categorical features dominate (0.397 vs. 0.290), confirming that embeddings serve as effective inductive biases under the right conditions. Similarly, MLP-CN outperforms MLP-C in numerical-heavy settings, but shows limited benefit otherwise.

While the proportion of categorical features captures how much of the input space is categorical, it does not reflect their complexity. As a proxy for categorical granularity, we use average cardinality. High cardinality increases the size of embedding tables and leads to sparse input representations, which may affect model performance. 
In practice, however, average cardinality shows minimal effect. This suggests that current models do not adequately capture fine-grained variation in categorical complexity, leaving room for architectural improvement.



\vspace{-0.2cm}
\subsubsection{Feature-to-sample ratio: NN-Feature remains robust across regimes}
\vspace{-0.2cm}

The feature-to-sample ratio is an important factor in tabular learning, influencing both generalization and overfitting risk. Tree-based models handle high-dimensionality through feature subsampling and regularization, while neural networks typically rely on architectural components such as attention or embedding modules to mitigate overfitting in high-dimensional regimes. Prior work~\citep{mcelfresh2024neural} found that GBDTs tend to outperform NNs when this ratio is low.

In low-ratio settings, XGBoost and TabR achieve the best average performance, challenging earlier claims of GBDT superiority. ModernNCA, FT-Transformer, and T2G-Former also perform competitively, further supporting the emergence of neural models with effective inductive biases. MLP-CN outperforms simpler baselines, suggesting that its numerical encoder benefits from sufficient training data.
In contrast, when the number of features largely exceeds the number of samples, NN-Feature achieves the best performance, consistent with the view that attention mechanisms are effective in modeling high-dimensional inter-feature dependencies. NN-Sample, by contrast, shows a substantial drop in performance in Figure~\ref{fig:heatmap}, likely because inter-sample similarity becomes unreliable in sample-scarce regimes.

Overall, NN-Feature models exhibit robust performance across both settings, while NN-Sample models are more sensitive to the balance between features and available data. Models that rely on inter-sample similarity may require caution under data scarcity, as their effectiveness depends on reliable neighborhood structure. In contrast, attention to inter-feature dependencies appears more robust to changes in the feature-to-sample ratio.

\vspace{-0.2cm}
\subsubsection{Label imbalance: NNs are more sensitive to skewed targets than GBDTs}
\vspace{-0.2cm}

We evaluate model robustness to label imbalance using three metrics: imbalance factor~\citep{wang2024towards} (applicable to all tasks), entropy ratio (for classification), and skewness (for regression). Although the imbalance factor is task-agnostic, it captures only the ratio between the majority and minority classes, not the full class distribution. In contrast, the entropy ratio, which reflects global class skew, reveals more informative patterns in classification tasks. 
As shown in Figure~\ref{fig:heatmap}, errors tend to decrease under imbalanced settings when measured by entropy ratio. This pattern likely reflects widened gaps between stronger and weaker models, rather than universal improvement. ModernNCA performs reliably across both balanced and imbalanced conditions, while a broader set of models also show strong performance under low entropy ratio, though not tied to specific architectures.

For regression, CatBoost achieves the best average performance across both low- and high-skew settings, with ModernNCA also performing consistently well. However, while their performance is similar under low skew (0.250 vs. 0.255), the gap widens substantially in high-skew regimes (0.245 vs. 0.302), indicating that GBDTs still maintain a notable advantage when the target distribution is highly skewed. 
Interestingly, performance changes under label imbalance vary by model family. Tree-based models such as Random Forest, XGBoost, and LightGBM all show improved performance under high-skew regime. In contrast, most neural models exhibit performance degradation in high-skew settings. While recent neural models have narrowed the gap with GBDTs, a notable performance difference remains—especially under skewed regression targets. Bridging this gap is an important direction for future work on neural robustness to label imbalance.

\vspace{-0.2cm}
\subsubsection{Function irregularity: High irregularity remains challenging for all model classes}
\vspace{-0.2cm}

Function irregularity—where small input changes lead to large or inconsistent target shifts—remains a core challenge in tabular learning. We assess model robustness using a frequency-based metric~\citep{beyazit2024inductive}. Such behavior is known to favor tree-based models, while neural networks tend to be biased toward smoother functions~\citep{grinsztajn2022tree,mcelfresh2024neural,beyazit2024inductive,lee2024binning}.


We found that ModernNCA achieves the best average performance in both regular and irregular regimes, though its margin over GBDTs is not statistically significant. Surprisingly, as shown in Figure~\ref{fig:heatmap}, XGBoost exhibits one of the largest performance drops under irregular conditions, contradicting its common reputation for robustness in such settings. MLP-CN, which is specifically designed to model local nonlinearities through periodic encodings, also shows substantial degradation, suggesting a misalignment between its design objective and actual behavior.
In contrast, NN-Sample models degrade less severely, indicating that learning with sample similarity can offer robustness to irregular patterns. 
Plain MLP experiences the largest decline, suggesting that the absence of structural priors leaves neural models particularly vulnerable to irregularity.

Taken together, these results confirm that function irregularity remains a persistent challenge in tabular learning. While inductive biases—such as piecewise splitting or periodic encodings—have been proposed, we find no consistent performance gains. Our analysis suggests that existing models fall short in highly irregular regimes, highlighting the need for new architectures and training strategies tailored to such conditions.



\vspace{-0.2cm}
\subsubsection{Feature interaction: NN-Sample favors correlated features, NN-Feature favors less correlated ones}
\vspace{-0.2cm}

We assess model performance under varying inter-feature dependencies using two complementary metrics. The first is the Frobenius norm of the correlation matrix, capturing the overall magnitude of linear relationships among features. The second is the minimum eigenvalue of the standardized covariance matrix, reflecting global redundancy or near-degeneracy in feature space.

NN-Sample and NN-Feature exhibit distinct strengths depending on feature correlation. NN-Sample performs best when features are highly correlated, likely benefiting from redundancy that enhances inter-sample similarity by increasing neighborhood consistency. In contrast, NN-Feature excels under low correlation, where attention over features can more effectively capture diverse and informative feature interactions. These trends are most evident under the Frobenius norm, but also hold on average under the eigenvalue metric.

These findings provide empirical evidence that architectural inductive biases influence how models respond to feature correlation. NN-Sample benefits from highly correlated features, where redundancy reinforces sample similarity. In contrast, NN-Feature performs best under low correlation, where inter-feature attention can more effectively identify informative patterns. These results illustrate that the effectiveness of a given inductive bias depends strongly on the structure of the input data.

\vspace{-0.2cm}
\subsection{Correlation Between Dataset Statistics and Model Performance}
\vspace{-0.2cm}

\begin{wrapfigure}{r}{0.45\textwidth}
    \centering
    \vspace{-0.35cm}
    \includegraphics[width=\linewidth]{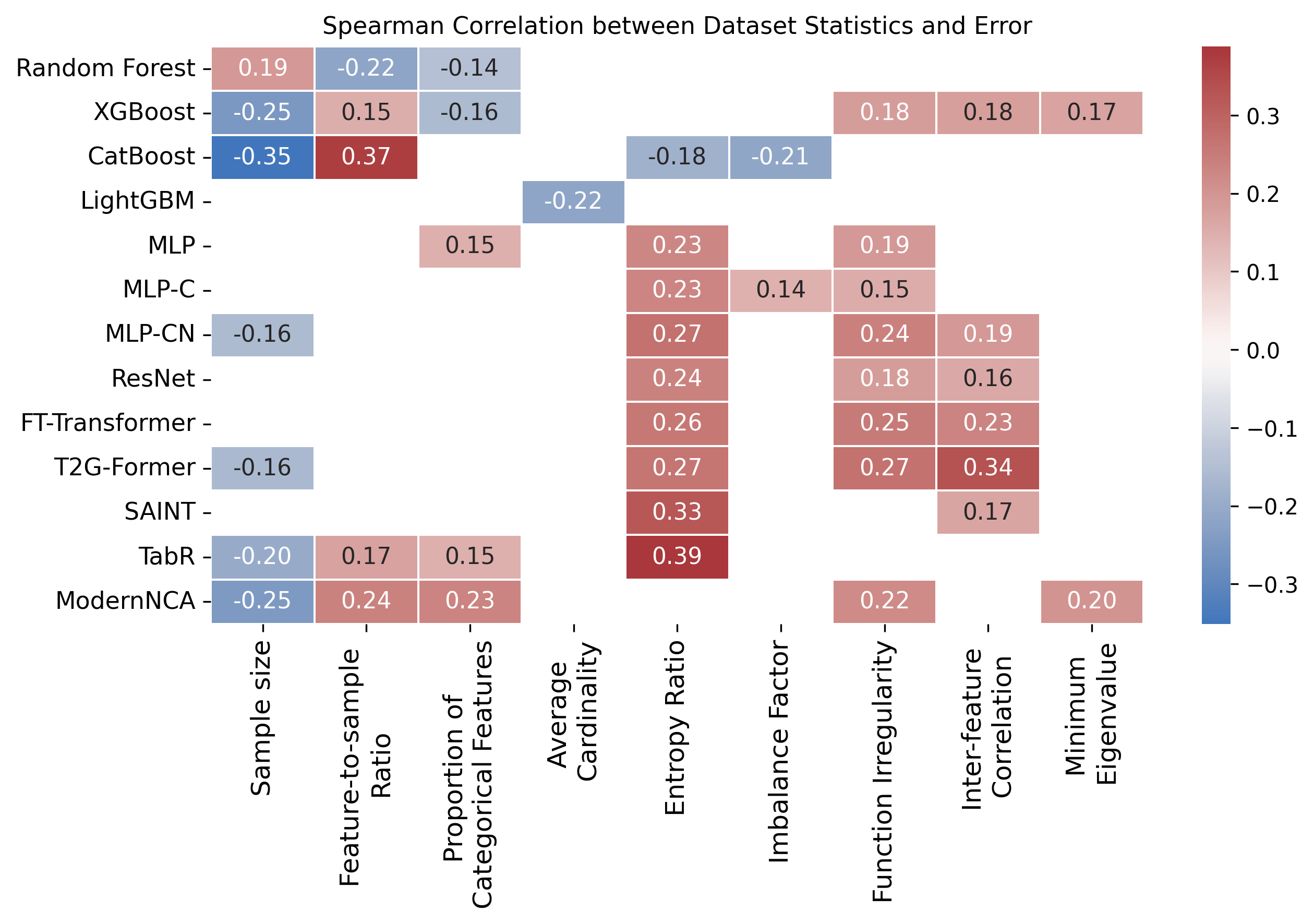}
    \caption{
    Spearman correlations between dataset statistics and model error. Only statistically significant correlations ($p < 0.05$) are shown; blank cells denote non-significance.
    }
    \vspace{-0.5cm}
    \label{fig:spearman}
\end{wrapfigure}

To understand continuous trends beyond discrete subgroups, we compute Spearman correlations between dataset statistics and model error, as summarized in Figure~\ref{fig:spearman}. Red cells denote performance degradation as the statistic increases; blue cells indicate improvement; and white cells indicate non-significance.

GBDTs show strong dependence on scale-related properties such as sample size and feature-to-sample ratio. For example, CatBoost exhibits a strong negative correlation with sample size, suggesting improved performance with more data, consistent with prior findings~\citep{mcelfresh2024neural}. 
In contrast, neural models are less affected by scale and more sensitive to structural data properties. Entropy ratio, function irregularity, and inter-feature correlation are positively correlated with error in most NNs, suggesting reduced robustness to complex or irregular patterns.
Further analysis is provided in Appendix~\ref{appendix:additionalresults-correlation}.



\vspace{-0.2cm}
\subsection{Comparison with TabPFN}
\vspace{-0.2cm}


TabPFN~\citep{hollmann2022tabpfn} has gained attention as a pretrained tabular model, but its use is limited by architectural constraints, including input dimensions, class counts, and sample size. We evaluate it on the 42 datasets that meet these conditions. Since normalized error is not applicable, we compare against ModernNCA, which showed consistently strong performance in our benchmark.

As shown in Figure~\ref{fig:tabpfn} in Appendix~\ref{appendix:additionalresults-tabpfn}, 
TabPFN outperforms ModernNCA on 29 datasets, while ModernNCA performs better on 13. This suggests TabPFN is competitive within its applicable scope but not universally dominant, with gains often small.
A Welch’s $t$-test across dataset statistics reveals that entropy ratio is the only significant factor. TabPFN tends to perform better on less balanced datasets (mean: 0.550), while ModernNCA excels in balanced ones (mean: 0.877). 
These findings suggest that TabPFN is particularly suited to imbalanced classification tasks. However, its applicability remains limited, highlighting the need for more scalable and flexible pretrained models.

\vspace{-0.2cm}
\section{Conclusion}
\label{sec:conclusion}
\vspace{-0.2cm}

This study revisits tabular model evaluation by moving beyond average-case metrics to a structured, data-aware framework. Using \textsc{MultiTab}, we analyze performance across sub-categories defined by key dataset characteristics such as sample size, label imbalance, and feature interactions. Our results show that no single model dominates: GBDTs excel in small or categorical-heavy settings, while modern neural networks are competitive on larger, more regular datasets. This underscores the importance of aligning model design with data regimes rather than relying on global rankings.

We also find that architectural components like attention and metric learning offer benefits only under specific conditions, reinforcing the need to consider dataset structure in both benchmark and model design. These findings suggest that practitioners can benefit from condition-aware model selection, and that benchmark suites should report performance across data regimes—not just on average. While our study covers a broad spectrum of datasets and models, it is limited to supervised settings; future work may extend this framework to pretraining-based or AutoML-driven scenarios. We hope \textsc{MultiTab} serves as a foundation for further research on adaptive architectures, data-driven model selection, and evaluation under distribution shift.

\bibliographystyle{unsrt}
\bibliography{reference}


\clearpage

\appendix
\appendix

\section*{Appendix}

\section{Broader Societal Impact Statement}
\label{appendix:impact}

This study introduces \textsc{MultiTab}, a large-scale benchmark suite designed to advance our understanding of how model architectures interact with data characteristics in tabular learning. Tabular data remains the backbone of decision-making in domains such as healthcare, finance, scientific research, and public services. By systematically identifying when different algorithms succeed or fail, \textsc{MultiTab} can guide practitioners in selecting more appropriate models for their data regimes, ultimately improving predictive performance, efficiency, and fairness in high-stakes applications.

In practice, this could translate to more reliable clinical risk predictions, better allocation of public resources, or more accurate detection of financial fraud. Moreover, our sub-category framework promotes a more nuanced view of benchmarking—moving beyond average-case performance toward context-aware model evaluation. This shift has the potential to influence the development of adaptive machine learning systems that are better aligned with real-world deployment settings.

At the same time, it is critical to recognize that performance improvements alone do not guarantee equitable or trustworthy systems. Models that perform well on benchmarked data may still perpetuate biases if the data reflects historical inequities. Additionally, reliance on automated predictions in sensitive domains must be accompanied by transparent reporting, interpretability, and accountability mechanisms. While \textsc{MultiTab} offers a valuable step toward better model-data alignment, responsible use requires ongoing attention to privacy, fairness, and human oversight.

In sum, we hope that \textsc{MultiTab} not only drives progress in model evaluation and selection, but also supports a more responsible and context-aware adoption of machine learning in tabular domains.

\section{Detailed Description of the Benchmark Suite}
\label{appendix:benchmark}

Our benchmark suite encompasses a comprehensive collection of 196 datasets, all of which are publicly accessible through the OpenML~\citep{OpenMLPython2019} or scikit-learn~\citep{scikit-learn} python libraries. All datasets are distributed under the CC-BY license, ensuring public availability and adherence to ethical data-sharing practices. OpenML enforces these standards through dataset-level documentation and contributor compliance. The benchmark suite and full optimization and reproduction logs are available at \url{https://huggingface.co/datasets/LGAI-DILab/Multitab}.

We provide a detailed list of OpenML dataset IDs for quick reference. Each dataset can be loaded via 
\texttt{openml.datasets.get\_dataset(DATASET\_ID)}.
One exception is the California housing dataset (ID: 999999), which is sourced from scikit-learn due to its widespread use in prior work~\citep{gorishniy2021revisiting, gorishniy2023tabr, lee2024binning}.

\begin{figure}[h!]
    \centering
    \includegraphics[width=0.4\linewidth]{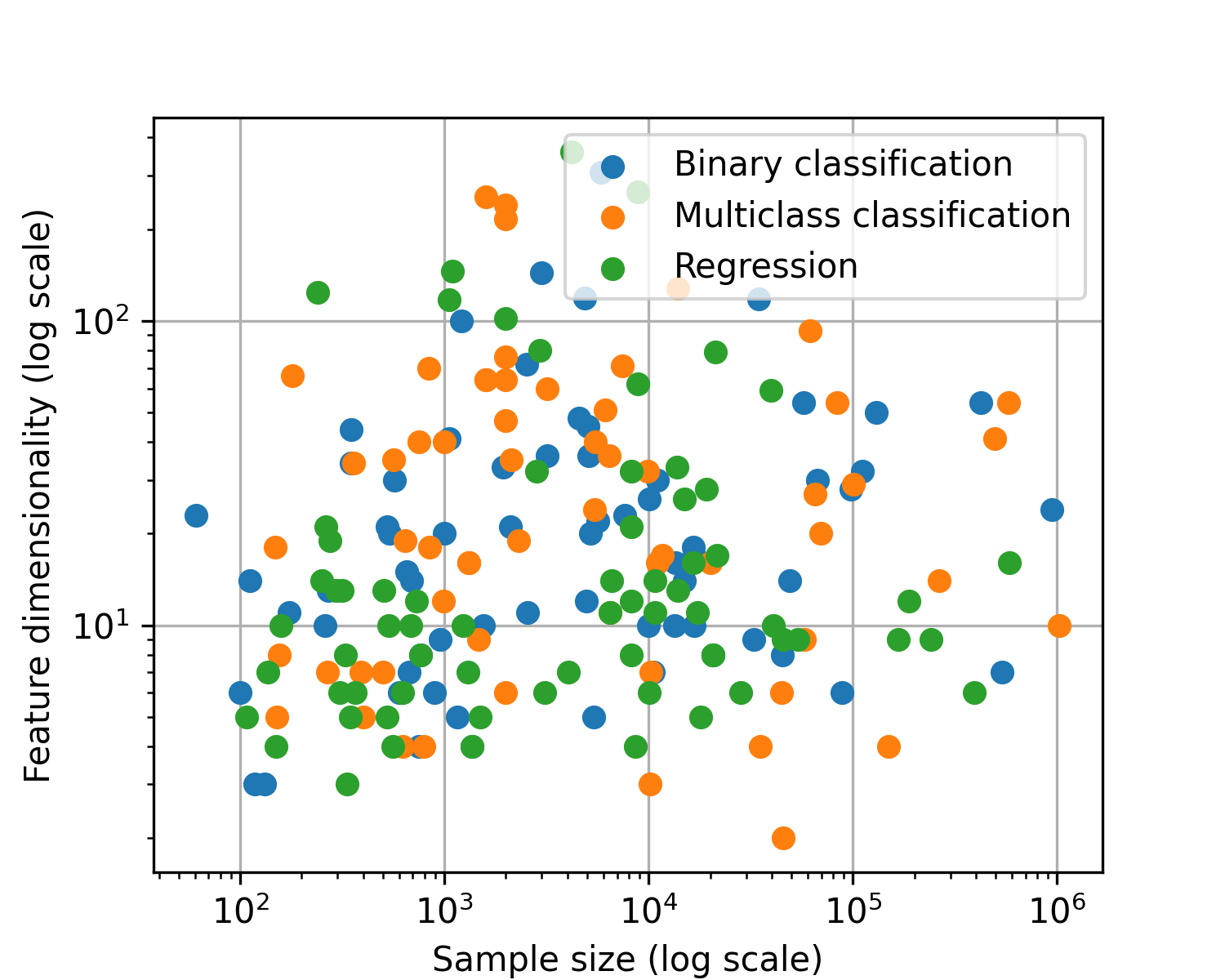}
    \caption{Distribution of datasets in our benchmark suite. Each point represents a dataset, plotted by sample size (x-axis) and feature dimensionality (y-axis), both on logarithmic scale. Colors indicate task types: binary classification, multiclass classification, and regression. The benchmark ensures broad coverage across different data scales and tasks.}
    \label{fig:data_summary}
\end{figure}

In our benchmark suite, the datasets span a diverse range of real-world domains, including medical, financial, environmental, social, and scientific contexts. To ensure broad coverage, we intentionally include a balanced number of datasets across task types (binary classification, multiclass classification, and regression), while also considering diversity in sample size and feature dimensionality during dataset selection. This ensures that the benchmark reflects diverse modeling scenarios, as illustrated in Figure~\ref{fig:data_summary}.

Here is the complete list of 196 datasets used in this study:

\begin{itemize}[leftmargin=15pt]
    \item 25, 461, 210, 466, 42665, 444, 497, 10, 1099, 48, 338, 40916, 23381, 4153, 505, 560, 51, 566, 452, 53, 524, 49, 194, 42370, 511, 509, 456, 8, 337, 59, 35, 42360, 455, 475, 40496, 531, 1063, 703, 534, 1467, 44968, 42, 1510, 334, 549, 11, 188, 29, 470, 43611, 40981, 45102, 1464, 1549, 37, 43962, 469, 458, 54, 45545, 50, 307, 1555, 31, 1494, 4544, 41702, 934, 1479, 41021, 41265, 185, 454, 1462, 43466, 23, 43919, 42931, 1501, 1493, 1492, 1504, 315, 20, 12, 14, 16, 22, 18, 1067, 1466, 36, 1487, 44091, 42727, 43926, 41143, 507, 46, 3, 44055, 44061, 1043, 44160, 44158, 40900, 44124, 1489, 1497, 40499, 24, 41145, 1475, 182, 44136, 43986, 503, 372, 44157, 558, 44132, 562, 189, 40536, 44056, 422, 44054, 4538, 45062, 44145, 44122, 1531, 1459, 44126, 44062, 42183, 32, 4534, 42734, 44125, 44123, 44137, 1476, 44005, 1471, 44133, 846, 44134, 44162, 44089, 44063, 44026, 45012, 6, 44090, 537, 999999, 44148, 44066, 44984, 4135, 1486, 45714, 44064, 344, 41027, 151, 44963, 40985, 45068, 44059, 44131, 40685, 45548, 41169, 41162, 42345, 41168, 40922, 23512, 40672, 44161, 41150, 1509, 44057, 43928, 44069, 1503, 44068, 44159, 1113, 1169, 150, 44065, 44129, 1567
\end{itemize}

\subsection{Preprocessing}

To ensure fair and consistent evaluation across models, we apply a standardized preprocessing pipeline to all datasets, following common conventions in tabular learning research~\citep{gorishniy2021revisiting,salinas2023tabrepo,mcelfresh2024neural}. The main steps are as follows:

\begin{itemize}[leftmargin=*]
    \item \textbf{Missing value handling:}
    \begin{itemize}
        \item Columns with more than 50\% missing values are dropped.
        \item Remaining rows with any missing values in input features or target labels are removed.
        \item No imputation is applied to avoid introducing artificial bias.
    \end{itemize}

    \item \textbf{Feature type detection:}
    \begin{itemize}
        \item Categorical features are defined strictly based on OpenML-provided metadata.
        \item No automatic type inference is performed.
        \item Datasets containing string-typed features incorrectly labeled as numeric, or categorical features with more than 1,000 unique values, are excluded due to frequent training failures across models.
    \end{itemize}    

    \item \textbf{Numerical feature transformation:}
    \begin{itemize}
        \item All numerical features are transformed using scikit-learn’s \texttt{QuantileTransformer} (default parameters), mapping values to a uniform distribution~\citep{gorishniy2021revisiting,gorishniy2022embeddings,grinsztajn2022tree,de2023choice}.
        \item This transformation stabilizes optimization and reduces the impact of outliers, as reported in prior work on tabular deep learning.
    \end{itemize}

    \item \textbf{Categorical feature encoding:}
    \begin{itemize}
        \item Categorical features are encoded using scikit-learn’s \texttt{LabelEncoder}, assigning a unique integer to each category.
        \item After encoding, each algorithm handles categorical inputs according to its own design: for example, CatBoost applies internal target (ordered) encoding, while neural networks use learnable embedding layers.
    \end{itemize}
\end{itemize}

\subsection{Training Protocols: Cross-validation}

To ensure reliable and fair evaluation across datasets, we adopt a stratified $k$-fold cross-validation protocol, balancing statistical rigor with computational feasibility. The fold index $k$ also serves as the random seed for each split. Each dataset-split pair is partitioned into an 8:1:1 train-validation-test ratio. The procedure is as follows:

\begin{itemize}[leftmargin=*]
    \item For datasets with fewer than 50,000 samples, we use 10-fold cross-validation.
    \item For larger datasets (50,000 samples or more), we use 3-fold cross-validation to reduce computational cost.
    \item For each fold, we select the best hyperparameter configuration using the validation split, then retrain the model on the training portion of that fold.
    \item Final performance is computed by averaging predictive error across all folds.
\end{itemize}

This setup prevents overfitting to any particular split and provides stable estimates of model performance across diverse datasets.

\subsection{Evaluation Metrics}

To enable fair performance comparison across datasets with varying scales and difficulty levels, we adopt \textit{normalized predictive error} as our primary evaluation metric. For classification tasks, the base error metric is log loss; for regression tasks, it is root mean squared error (RMSE).

To account for dataset-specific difficulty and variance, we normalize each model's error within every dataset-split pair. Let $e_{m,d}$ denote the error of model $m$ on dataset-split pair $d$, and let $e_{d}^{\min}$ and $e_{d}^{\max}$ represent the minimum and maximum errors across all models on $d$, respectively. The normalized error is defined as:
\[
\hat{e}_{m,d} = \frac{e_{m,d} - e_d^{\min}}{e_d^{\max} - e_d^{\min}}, \quad \hat{e}_{m,d} \in [0, 1],
\]
where $\hat{e}_{m,d} = 0$ indicates the best-performing model on that dataset-split pair, and $1$ indicates the worst.

The final score for each model is obtained by averaging $\hat{e}_{m,d}$ across all dataset-split pairs in the benchmark. This formulation ensures that:
\begin{itemize}[leftmargin=*]
    \item Models are evaluated relative to others on the same dataset, avoiding unfair penalization due to intrinsic task difficulty.
    \item Aggregated scores reflect relative, rather than absolute, performance—mitigating scale mismatch across datasets.
\end{itemize}

We select log loss as the default metric for classification because it remains informative under class imbalance, reflects prediction confidence, and is well-defined even when only one class is observed. In contrast, AUROC is undefined in certain edge cases, and accuracy can be misleading under skewed class distributions.

As secondary metrics, we also report average accuracy, AUROC, and raw rank across all datasets. These provide additional interpretability and complement the normalized score, although they are less sensitive to performance magnitude. These are included in the released benchmark suite for completeness but are not used in our main analysis.

\section{Detailed Description of the Sub-category Construction}
\label{appendix:criteria}

To support structured evaluation, \textsc{MultiTab} defines seven core data-centric axes that capture key properties of tabular datasets, as summarized in Table~\ref{tab:suite}. For each axis, we partition datasets into sub-categories based on prior studies or empirically determined thresholds. The resulting distributions are visualized in Figure~\ref{fig:histogram}.

Below, we provide detailed descriptions of each axis, the rationale behind the sub-category definitions, and the corresponding dataset statistics.

\subsection{Task types}

We classify the full set of 196 datasets into three sub-categories based on their task type, as defined in the OpenML metadata. This categorization reflects fundamental differences in label structure that influence model training and generalization behavior. 

Prior studies have observed consistent trends across these task types. GBDTs such as CatBoost and LightGBM typically perform strongly, particularly on regression tasks, while context-based models like ModernNCA maintain competitive results across all categories~\citep{ye2024closer,mcelfresh2024neural}. Token-based methods (\emph{e.g.}, FT-Transformer, AutoInt, ExcelFormer) tend to perform similarly, suggesting shared inductive biases based on learned embeddings~\citep{chen2023excelformer}. Among MLP variants, RealMLP~\citep{holzmuller2024better} demonstrates significantly better performance than other MLP-based baselines. Some studies also report task-specific architectural effects: for example, PLR embeddings and parametric activations significantly improve performance in regression tasks but show limited benefit in classification~\citep{holzmuller2024better}.

\begin{itemize}[leftmargin=15pt]
    \item Binary classification (68 datasets): 25, 461, 466, 42665, 444, 23381, 51, 53, 49, 337, 59, 1063, 1467, 1510, 334, 29, 470, 40981, 1464, 37, 45545, 50, 31, 1494, 934, 1479, 1462, 42931, 1504, 4135, 1486, 44161, 41150, 1067, 1487, 44091, 41143, 3, 1043, 44160, 44158, 40900, 44124, 1489, 24, 41145, 44157, 40536, 45062, 151, 45068, 44131, 44159, 1169, 44129, 44122, 44126, 4534, 44125, 44123, 1471, 846, 44162, 44089, 44090, 41162, 40922, 23512
    \item Multiclass classification (60 datasets): 10, 48, 338, 4153, 452, 35, 455, 475, 40496, 42, 11, 188, 1549, 469, 458, 54, 307, 1555, 185, 23, 1501, 1493, 1492, 20, 12, 14, 16, 22, 18, 45714, 41027, 40672, 1509, 1113, 150, 1567, 1466, 36, 46, 1497, 40499, 1475, 182, 43986, 372, 4538, 40985, 40685, 45548, 41169, 1531, 1459, 32, 42734, 1476, 6, 42345, 41168, 1503, 454
    \item Regression (68 datasets): 210, 497, 1099, 40916, 505, 560, 566, 524, 194, 42370, 511, 509, 456, 8, 42360, 531, 703, 534, 44968, 549, 43611, 45102, 43962, 4544, 41702, 41021, 41265, 43466, 43919, 315, 44984, 44064, 344, 44057, 43928, 42727, 43926, 507, 44055, 44061, 44136, 503, 558, 44132, 562, 189, 44056, 422, 44054, 44963, 44059, 44068, 44145, 44062, 42183, 44137, 44005, 44133, 44134, 44063, 44026, 45012, 537, 999999, 44148, 44066, 44069, 44065
\end{itemize}

\subsection{Sample size}

We define this axis based on the total number of samples in each dataset, which captures variation in data availability and its interaction with model capacity and optimization dynamics. Larger datasets typically provide more statistical signal, while smaller datasets may limit generalization and overfit complex architectures. Datasets are grouped into sub-categories following common thresholds used in prior work~\citep{grinsztajn2022tree,mcelfresh2024neural,hollmann2022tabpfn}, with small datasets containing fewer than 1,000 samples and large datasets containing at least 10,000.

Prior studies consistently highlight the importance of dataset size in shaping model performance. GBDTs, especially CatBoost, are known to benefit substantially from increased sample size~\citep{mcelfresh2024neural,ye2024closer}, often improving in rank as data availability increases~\citep{ye2024closer}. In contrast, neural models such as MLPs or FT-Transformer tend to degrade in performance on larger datasets, potentially due to limited inductive bias or scalability constraints~\citep{ye2024closer,holzmuller2024better}. However, context-based neural architectures like ModernNCA and large-capacity models such as TabNet may remain competitive under sufficient data~\citep{borisov2022deep}.

Interestingly, TabPFN exhibits strong performance on small datasets due to its learned prior and limited architectural capacity, but struggles to scale effectively to larger regimes~\citep{hollmann2022tabpfn}. While neural models occasionally outperform GBDTs on smaller datasets, this appears highly dependent on architectural design and task characteristics~\citep{shmuel2024comprehensive}. Overall, dataset size remains one of the most influential factors in tabular model comparison, and evaluating model behavior across sample regimes is critical for understanding scalability and generalization.

\begin{itemize}[leftmargin=15pt]
    \item Small: Fewer than 1,000 samples (62 datasets: 25, 461, 210, 466, 42665, 444, 497, 10, 1099, 48, 338, 40916, 23381, 4153, 505, 560, 51, 566, 452, 53, 524, 49, 194, 42370, 511, 509, 456, 8, 337, 59, 35, 42360, 455, 475, 40496, 531, 1063, 703, 534, 1467, 44968, 42, 1510, 334, 549, 11, 188, 29, 470, 43611, 40981, 45102, 1464, 1549, 37, 43962, 469, 458, 54, 45545, 50, 307)
    \item Large: More than 10,000 samples (68 datasets: 44984, 4135, 1486, 45714, 44064, 344, 41027, 40672, 44161, 41150, 1509, 44057, 43928, 1113, 150, 1567, 45062, 151, 44963, 40985, 45068, 44059, 44131, 40685, 45548, 41169, 44068, 44159, 1169, 44129, 44145, 44122, 1531, 1459, 44126, 44062, 42183, 32, 4534, 42734, 44125, 44123, 44137, 1476, 44005, 1471, 44133, 846, 44134, 44162, 44089, 44063, 44026, 45012, 6, 44090, 537, 999999, 44148, 44066, 41162, 42345, 41168, 40922, 23512, 44069, 1503, 44065)
\end{itemize}

The detailed histogram of sample size is provided in Figure~\ref{fig:histogram:samplesize}.

\subsection{Feature heterogeneity}

Tabular datasets often include a heterogeneous mix of feature types--most commonly, categorical and numerical--which differ in representation and modeling requirements. This axis captures how the proportion and characteristics of categorical features vary across datasets, which can influence model performance depending on the inductive biases and input encoding strategies of each algorithm. 

Despite its importance, this aspect remains relatively underexplored in prior studies. Few works explicitly analyze how feature heterogeneity affects model behavior. TabZilla~\citep{mcelfresh2024neural} has only reported that TabPFN performs best on small datasets with fewer numerical features, while ResNet variants tend to excel when the number of numerical features increases. However, no prior study has systematically examined the role of feature type composition across large-scale tabular benchmarks.

In this work, we investigate this property using two complementary axes:
\begin{itemize}[leftmargin=*]
    \item \textbf{Proportion of categorical features:} the ratio of categorical to total input features.
    \item \textbf{Average cardinality:} the mean number of unique values across all categorical features.
\end{itemize}
Sub-categories based on these metrics are described in the following subsections.

\subsubsection{Proportion of categorical features}

This metric quantifies the fraction of input features that are categorical:
\[
\text{Proportion} = \frac{\# \text{categorical features}}{\# \text{total features}}.
\]
It reflects the relative dominance of categorical inputs within the dataset, which is distinct characteristic of tabular data. High proportions may challenge models that rely heavily on continuous feature interactions (\emph{e.g.}, metric learning or raw attention), while favoring those with strong categorical handling mechanisms (\emph{e.g.}, GBDTs with target encoding or embedding-based neural networks).

We expect neural models without explicit categorical encoders (such as vanilla MLPs) to perform poorly when categorical features dominate, due to their inability to effectively represent discrete variables. In contrast, models equipped with learnable embeddings, such as FT-Transformer or MLP-C, are more likely to maintain stable performance across different proportions of categorical features. Tree-based models like CatBoost, which natively handle categorical splits through ordered or frequency-based encodings, are expected to perform consistently regardless of the proportion.

\begin{itemize}[leftmargin=15pt]
    \item Few: Less than 20\% (123 datasets: 466, 42665, 1099, 40916, 4153, 505, 560, 566, 53, 42370, 509, 8, 337, 59, 42360, 455, 40496, 531, 1063, 1467, 44968, 1510, 549, 11, 43611, 45102, 1464, 1549, 37, 43962, 458, 54, 307, 1555, 1494, 4544, 41702, 1479, 41265, 185, 1462, 43466, 43919, 42931, 1501, 1493, 1492, 1504, 315, 12, 14, 16, 22, 18, 41027, 44161, 41150, 1509, 1113, 1567, 1067, 1466, 36, 1487, 44091, 507, 1043, 40900, 44124, 1489, 1497, 40499, 41145, 1475, 182, 44136, 43986, 503, 44157, 558, 44132, 562, 189, 422, 4538, 151, 44963, 40985, 44131, 40685, 45548, 41169, 44129, 44145, 44122, 1531, 1459, 44126, 42183, 32, 44125, 44123, 44137, 1476, 44005, 1471, 44133, 846, 44134, 44089, 44026, 45012, 6, 44090, 537, 999999, 44148, 44066, 41168, 40922, 23512, 1503, 454)
    \item Large: More than 60\% samples (33 datasets: 25, 444, 10, 338, 23381, 456, 35, 475, 534, 42, 334, 469, 45545, 50, 31, 934, 23, 20, 44984, 4135, 150, 41143, 46, 3, 44055, 44061, 24, 372, 44159, 4534, 42734, 42345, 44069)
\end{itemize}

The detailed histogram of proportion of categorical features is provided in Figure~\ref{fig:histogram:cat}.

\subsubsection{Average cardinality}

This metric measures the average number of unique values across all categorical features:
\[
\text{Average cardinality} = \frac{1}{C} \sum_{i=1}^{C} \left| \mathcal{V}_i \right|,
\]
where $C$ is the number of categorical features and $\mathcal{V}_i$ denotes the set of unique values in feature $i$. This captures the granularity or resolution of categorical inputs, which can affect both model capacity and representational complexity.

High-cardinality features can pose challenges for many models. Neural networks must learn embeddings over large vocabularies, which may result in overfitting or undertraining if data is sparse. Tree-based models can also suffer from data fragmentation when splitting on rare categories. As such, models with robust categorical handling---such as CatBoost or embedding-based neural architectures---are expected to perform better under high-cardinality conditions.
In contrast, when average cardinality is low, categorical distinctions are easier to learn and most models are likely to perform similarly. Therefore, performance differences are expected to be more pronounced in the high-cardinality regime.

\begin{itemize}[leftmargin=15pt]
    \item Low: Average cardinality is less than 3 (33 datasets: 466, 497, 10, 48, 51, 524, 49, 194, 511, 534, 42, 334, 1549, 45545, 1555, 44984, 1486, 44064, 344, 44161, 150, 41143, 3, 44055, 44061, 44157, 44056, 44054, 44159, 4534, 44063, 44066, 44069)
    \item High: Average cardinality is larger than 10 (21 datasets: 40916, 23381, 566, 452, 456, 703, 188, 470, 41021, 4135, 45714, 43928, 1113, 42727, 372, 45068, 1169, 42734, 45012, 41162, 42345)
\end{itemize}

The detailed histogram of average cardinality is provided in Figure~\ref{fig:histogram:card}.

\subsection{Feature-to-sample ratio}

This axis measures the ratio between the number of input features and the number of samples:
\[
\text{Feature-to-sample ratio} = \frac{d}{n},
\]
where $d$ is the number of input features and $n$ is the number of samples. This ratio characterizes how underdetermined or high-dimensional a dataset is, relative to its sample size, and is closely tied to generalization, overfitting risk, and the curse of dimensionality. It is particularly relevant in tabular learning, where dataset size and dimensionality vary widely across tasks.

High ratios indicate settings where the input space is not sufficiently supported by data, potentially challenging for models without strong regularization. Low ratios correspond to dense regimes with abundant training examples per feature, favoring models that can efficiently leverage statistical patterns.

Prior studies suggest that GBDTs perform well when the number of samples greatly exceeds the number of features~\citep{mcelfresh2024neural}, as each decision tree split can be reliably computed. In contrast, neural networks often exhibit degraded performance when the feature-to-sample ratio is high, where limited samples per input dimension can lead to overfitting and unstable optimization, particularly in architectures like ResNet or unregularized MLPs~\citep{grinsztajn2022tree,mcelfresh2024neural}.
It is expected that neural models with attention or embedding mechanisms to show improved performance when the ratio is high, whereas GBDTs and simpler neural architectures may dominate when the ratio is low.

\begin{itemize}[leftmargin=15pt]
    \item Low: Feature-to-sample ratio is less than 0.002 (64 datasets: 44984, 4135, 45714, 44064, 344, 41027, 40672, 44161, 41150, 1509, 44057, 43928, 1113, 150, 1567, 1489, 562, 189, 44056, 45062, 151, 44963, 40985, 45068, 44059, 44131, 40685, 41169, 44068, 44159, 1169, 44129, 44145, 1531, 1459, 44126, 44062, 42183, 32, 42734, 44125, 44123, 44005, 1471, 846, 44134, 44162, 44089, 44063, 44026, 45012, 6, 44090, 537, 999999, 44066, 41162, 42345, 41168, 40922, 23512, 44069, 1503, 44065)
    \item High: Feature-to-sample ratio is larger than 0.02 (61 datasets: 25, 461, 210, 466, 42665, 444, 497, 10, 1099, 48, 338, 40916, 23381, 4153, 505, 560, 51, 566, 452, 53, 524, 49, 194, 511, 509, 337, 59, 35, 531, 1063, 1467, 42, 1510, 188, 29, 40981, 1549, 458, 54, 1555, 31, 1494, 4544, 41702, 1479, 1501, 1493, 1492, 315, 20, 12, 14, 16, 22, 1487, 43926, 41143, 44061, 41145, 40536, 422)
\end{itemize}

The detailed histogram of the feature-to-sample ratio is provided in Figure~\ref{fig:histogram:fsratio}.

\subsection{Label imbalance}

Label imbalance captures asymmetries in the distribution of target values, a common characteristic in real-world tabular datasets. This phenomenon affects both classification and regression tasks, influencing model calibration, optimization dynamics, and overall performance stability.

Despite its practical importance, label imbalance has received limited attention in large-scale tabular benchmarking. To address this gap, we assess model robustness using three complementary metrics suited to different prediction settings: entropy ratio for classification, skewness for regression, and imbalance factor for both. These metrics allow us to capture distinct aspects of imbalance, ranging from global distribution skew to extreme class dominance and target asymmetry.

In classification tasks, neural networks that do not explicitly address imbalance may struggle due to biased gradients and limited representation of minority classes. In contrast, tree-based models such as CatBoost and XGBoost often exhibit greater robustness, as they incorporate internal mechanisms like class weighting.
Models based on metric learning, such as ModernNCA, may be especially sensitive to class imbalance, since they depend on latent-space proximity to capture label structure.
For regression tasks, strong skewness in the target distribution can impair model performance--particularly for models using fixed loss functions--unless they include normalization strategies or regularization schemes that account for target variance.

To operationalize this axis, we define sub-categories based on empirical thresholds drawn from the observed distribution across datasets. Details for each metric are provided below.

\subsubsection{Entropy ratio for classification task}

The entropy ratio quantifies the global imbalance in classification tasks by comparing the empirical label entropy to the maximum possible entropy given the number of classes. It is defined as:
\[
\text{Entropy Ratio} = \frac{H(\mathbf{y})}{\log |\mathcal{Y}|},
\quad \text{where} \quad
H(\mathbf{y}) = -\sum_{c \in \mathcal{Y}} p_c \log p_c,
\]
and \( \mathcal{Y} \) is the set of unique class labels, with \( p_c \) denoting the empirical proportion of class \( c \) in the dataset.

The entropy ratio ranges from 0 (completely imbalanced, with one dominant class) to 1 (perfectly balanced, uniform class distribution). This measure captures global label distribution skew and provides a stable signal for evaluating classifier robustness under imbalance.

\begin{itemize}[leftmargin=15pt]
    \item Balanced: Entropy ratio is larger than 0.7 (61 datasets: 25, 461, 466, 42665, 444, 23381, 51, 53, 49, 337, 59, 1063, 1510, 334, 29, 470, 40981, 1464, 37, 45545, 50, 31, 1494, 934, 1479, 1462, 42931, 1504, 1486, 44161, 41150, 44091, 41143, 3, 1043, 44160, 44158, 44124, 1489, 24, 41145, 44157, 45062, 151, 45068, 44131, 44159, 1169, 44129, 44122, 44126, 4534, 44125, 44123, 1471, 846, 44162, 44089, 44090, 40922, 23512)
    \item Imbalanced: Entropy ratio is less than 0.3 (43 datasets: 10, 48, 338, 35, 455, 475, 11, 188, 469, 458, 54, 1555, 185, 23, 45714, 41027, 40672, 1509, 1113, 150, 1567, 1466, 36, 46, 40900, 1497, 40499, 1475, 182, 43986, 4538, 40985, 40685, 45548, 1531, 1459, 32, 42734, 1476, 42345, 41168, 1503, 454)
\end{itemize}

The detailed histogram of the entropy ratio is provided in Figure~\ref{fig:histogram:entropy}.

\subsubsection{Skewness for regression task}

Skewness quantifies the asymmetry of the target distribution in regression tasks. We compute it using the sample skewness estimator implemented in \texttt{scipy.stats.skew}, defined as:
\[
\text{Skewness} = \frac{1}{n} \sum_{i=1}^{n} \left( \frac{y_i - \bar{y}}{s} \right)^3,
\]
where \( y_i \) is the target value, \( \bar{y} \) is the sample mean, \( s \) is the sample standard deviation, and \( n \) is the number of samples.

Positive skewness indicates a right-tailed distribution, while negative skewness indicates a left-tailed one. Highly skewed targets can degrade regression performance, particularly for models with fixed loss functions that are sensitive to outliers or non-uniform error magnitudes.

\begin{itemize}[leftmargin=15pt]
    \item \textbf{Balanced:} Absolute skewness less than 0.7 (27 datasets: 1099, 40916, 560, 524, 509, 456, 42360, 703, 44968, 43611, 45102, 43962, 41021, 43466, 43919, 42727, 44136, 503, 189, 44056, 44963, 44059, 44068, 44062, 44026, 44066, 44065)
    \item \textbf{Imbalanced:} Absolute skewness greater than 1.5 (24 datasets: 210, 497, 566, 42370, 8, 534, 549, 4544, 41265, 315, 44984, 44057, 43928, 43926, 44055, 558, 44132, 562, 422, 44054, 44145, 42183, 44134, 45012)
\end{itemize}

A histogram of the skewness distribution is shown in Figure~\ref{fig:histogram:skew}.

\subsubsection{Imbalance factor}

Imbalance factor quantifies the degree of class imbalance by measuring the ratio between the largest and smallest class sizes~\citep{wang2024towards}. It is formally defined as:
\[
\text{Imbalance Factor} = \frac{n_1}{n_C},
\]
where \( n_1 \) is the size of the most frequent class and \( n_C \) is the size of the least frequent class.

We adopt this metric following the framework of ~\citep{wang2024towards}, who propose imbalance factor alongside the Gini coefficient and Pareto-LT ratio as key indicators of long-tailed label distributions. Since these metrics exhibit highly correlated trends, we use imbalance factor as a representative measure of class imbalance in our benchmark but other metrics are also available in the suite.

According to their interpretation, a high value suggests severe data imbalance, often necessitating mitigation strategies such as class reweighting or resampling. Conversely, when the value is low, general-purpose models tend to perform adequately without explicit imbalance handling.

\begin{itemize}[leftmargin=15pt]
    \item Balanced: Imbalance factor is less than 3 (74 datasets: 25, 461, 466, 42665, 444, 1099, 48, 40916, 23381, 51, 53, 49, 509, 337, 59, 42360, 455, 703, 1510, 334, 11, 29, 470, 43611, 40981, 37, 50, 31, 1494, 4544, 1479, 185, 1462, 43466, 23, 42931, 1504, 1486, 41027, 44161, 41150, 44091, 41143, 507, 46, 3, 44160, 44158, 44124, 1489, 24, 41145, 44157, 189, 151, 44131, 44159, 1169, 44129, 44122, 44126, 4534, 42734, 44125, 44123, 1471, 846, 44162, 44089, 44090, 42345, 40922, 23512, 454)
    \item Imbalanced: Imbalance factor is larger than 5 (95 datasets: 4153, 505, 566, 452, 524, 194, 8, 35, 40496, 531, 534, 1467, 44968, 42, 549, 1549, 43962, 469, 307, 1555, 41702, 41021, 41265, 1501, 1493, 1492, 315, 20, 12, 14, 16, 22, 18, 44984, 4135, 45714, 44064, 344, 40672, 1509, 44057, 43928, 1113, 150, 1567, 1067, 1466, 36, 1487, 43926, 44055, 44061, 40900, 40499, 1475, 182, 44136, 43986, 503, 372, 558, 44132, 562, 44056, 422, 44054, 44963, 40985, 44059, 40685, 45548, 41169, 44068, 44145, 1459, 44062, 42183, 32, 44137, 1476, 44005, 44133, 44134, 44063, 44026, 45012, 6, 537, 999999, 44148, 44066, 41162, 44069, 1503, 44065)
\end{itemize}

The detailed histogram of imbalance factor is provided in Figure~\ref{fig:histogram:if}.

\subsection{Function irregularity}

Function irregularity captures how sensitively the target value responds to small changes in the input. In tabular data, such irregular relationships are common, where even minor input perturbations can cause abrupt label shifts. This presents challenges for models that assume smooth functional mappings.

To quantify this, we follow the approach proposed by~\cite{beyazit2024inductive}, which introduces a frequency-based irregularity score. Specifically, we apply a non-uniform discrete Fourier transform (NUDFT)~\citep{fessler2003nudft} to the first principal component of each training dataset. Using the same cutoff used in~\cite{beyazit2024inductive}, we define the high-frequency region as the spectral components beyond 0.5. The irregularity score is then computed as:
\[
\text{Irregularity Score} = \frac{||\text{High-frequency components}||_2^2}{||\text{All frequency components}||_2^2}.
\]
This value ranges from 0 to 1, with higher scores indicating more high-frequency variation and, hence, greater functional irregularity. For full implementation details, see \cite{beyazit2024inductive}.

Models that rely on smooth function approximation like simple NNs, are likely to degrade on highly irregular datasets, where small input changes cause large, unpredictable shifts in the target. Tree-based models like GBDTs are expected to be more robust in such cases due to their ability to model non-smooth, piecewise constant functions. Meanwhile, architectures that incorporate attention mechanisms or sample-level similarity, such as TabR and ModernNCA in our study, may offer partial robustness by capturing localized or instance-specific variations.

\begin{itemize}[leftmargin=15pt]
    \item Regular: High-frequency ratio is less than 0.3 (58 datasets: 4135, 1486, 44161, 41150, 1567, 41143, 3, 40900, 1497, 41145, 182, 503, 44157, 562, 4538, 40985, 44131, 40685, 45548, 41169, 1169, 44129, 44122, 1531, 4534, 44125, 44123, 1471, 846, 44089, 44090, 44066, 41162, 42345, 41168, 40922, 23512, 1503)
    \item Irregular: High-frequency ratio is larger than 0.7 (43 datasets: 25, 210, 444, 1099, 40916, 505, 566, 452, 511, 509, 456, 40496, 531, 703, 534, 11, 45102, 50, 307, 4544, 41702, 41021, 41265, 43919, 42931, 44984, 45714, 344, 44057, 43928, 1113, 42727, 44055, 44061, 372, 40536, 45062, 151, 44062, 42183, 44063, 45012, 44065)
\end{itemize}

The detailed histogram of function irregularity is provided in Figure~\ref{fig:histogram:irregularity}.

\subsection{Feature interaction}

Feature interaction captures the extent to which input features exhibit linear dependence or redundancy, which is a central aspect of tabular data structure. Unlike image or sequence domains, where spatial or temporal patterns guide modeling, tabular datasets vary widely in how features relate to one another. These inter-feature dependencies influence how effectively models can learn useful representations.

To characterize this, we use two complementary metrics: (1) the Frobenius norm of the correlation matrix, which reflects overall pairwise feature dependence, and (2) the minimum eigenvalue of the normalized covariance matrix, which highlights global redundancy and near-degeneracy in the feature space.

We can expect that models that rely on feature-wise modeling, such as attention-based architectures, may perform better when features are weakly correlated, enabling them to learn interactions more freely. Conversely, in highly correlated datasets, NN-Sample may benefit from redundancy that enhances instance-level representation. GBDTs are generally robust across both regimes due to their greedy, split-based handling of features, although they may be less sensitive to subtle linear interactions.

We describe the detailed formulation of each metric in the following subsections.

\subsubsection{Average Frobenius norm of correlation matrix between features}

We define the average inter-feature correlation using the Frobenius norm of the centered correlation matrix:
\[
\rho = \frac{||\text{Corr}(\mathbf{X}) - I||_F}{d(d - 1)},
\]
where \( \text{Corr}(\mathbf{X}) \in \mathbb{R}^{d \times d} \) is the feature-wise Pearson correlation matrix, \( I \) is the identity matrix of the same size, \( ||\cdot||_F \) denotes the Frobenius norm, and \( d \) is the number of input features.

This metric captures the average pairwise linear dependence between features, normalized by the number of off-diagonal entries. The subtraction of the identity matrix removes self-correlations (which are always 1), isolating the effect of off-diagonal correlations. \(\rho \approx 0\) indicates that features are largely uncorrelated, reflecting high-dimensional and potentially more informative inputs. In contrast, larger \(\rho\) values suggest that features are more linearly dependent, possibly redundant, and may challenge models that assume independence.
In our benchmark, this measure allows us to assess whether models perform better in settings with more independent or more redundant feature structures.

\begin{itemize}[leftmargin=15pt]
    \item Correlated: $\rho$ is larger than 0.03 (59 datasets: 461, 210, 466, 42665, 444, 1099, 48, 338, 40916, 560, 452, 42370, 509, 8, 42360, 455, 40496, 531, 1063, 44968, 549, 43611, 1464, 37, 43962, 469, 54, 45545, 934, 41021, 185, 1462, 43466, 43919, 42931, 18, 45714, 1067, 507, 1489, 503, 562, 44056, 151, 44963, 44059, 40685, 44145, 1459, 44126, 44062, 44125, 1471, 44026, 44090, 537, 999999, 40922, 44069)
    \item Uncorrelated: $\rho$ is less than 0.005 (39 datasets: 456, 1467, 334, 11, 1549, 458, 1555, 4544, 41702, 1501, 315, 20, 12, 14, 16, 1486, 44064, 41027, 150, 1567, 43926, 41143, 46, 3, 44061, 1043, 41145, 372, 558, 189, 40536, 422, 44054, 40985, 44131, 45548, 44159, 41168, 454)
\end{itemize}

The detailed histogram of the Frobenius norm of inter-feature correlation is provided in Figure~\ref{fig:histogram:correlation}.

\subsubsection{Minimum eigenvalue of normalized covariance matrix}

To capture global linear dependencies among features, we follow the formulation used in TabZilla~\citep{mcelfresh2024neural} and compute the minimum eigenvalue of the standardized covariance matrix. Given an input feature matrix \( \mathbf{X} \in \mathbb{R}^{n \times d} \), we first standardize all columns to have zero mean and unit variance, and compute the covariance matrix:
\[
\Sigma = \frac{1}{n - 1} \mathbf{X}^\top \mathbf{X},
\]
where \( \Sigma \in \mathbb{R}^{d \times d} \). The minimum eigenvalue \( \lambda_{\min}(\Sigma) \) is then extracted to measure global redundancy in the feature space.

A small \( \lambda_{\min} \) (close to zero) suggests that the features lie near a lower-dimensional subspace, indicating multicollinearity or degeneracy. In contrast, larger values indicate more independent and uniformly distributed features. Unlike the Frobenius norm, which reflects average pairwise linear relationships, this metric emphasizes global structure and rank sufficiency.

\begin{itemize}[leftmargin=15pt]
    \item Correlated: Minimum eigenvalue is less than 0.002 (56 datasets: 25, 4153, 505, 566, 59, 1063, 1467, 44968, 1510, 1464, 43962, 54, 4544, 41702, 1479, 185, 42931, 1492, 1504, 315, 12, 22, 1486, 44064, 44161, 43928, 1113, 150, 1067, 1466, 36, 1487, 41143, 44061, 1043, 44160, 44158, 40499, 24, 41145, 1475, 44157, 44132, 422, 44963, 44131, 44159, 44137, 1476, 846, 44134, 44162, 44148, 41162, 44069, 454)
    \item Uncorrelated: Minimum eigenvalue is larger than 0.1 (74 datasets: 461, 210, 466, 444, 497, 10, 1099, 48, 338, 23381, 51, 452, 53, 524, 49, 194, 509, 456, 8, 42360, 475, 40496, 531, 703, 334, 549, 11, 470, 43611, 1549, 37, 469, 458, 45545, 50, 307, 1555, 31, 934, 41265, 1462, 43466, 23, 14, 44984, 4135, 344, 41027, 1509, 44057, 1567, 46, 44055, 1489, 1497, 558, 189, 44056, 45062, 151, 40985, 45068, 45548, 44068, 1169, 44129, 1531, 1459, 42734, 44089, 44026, 42345, 40922, 1503)
\end{itemize}

\begin{figure}[h!]
    \centering
    \subfloat[Sample size]{
    \includegraphics[width=0.48\textwidth]{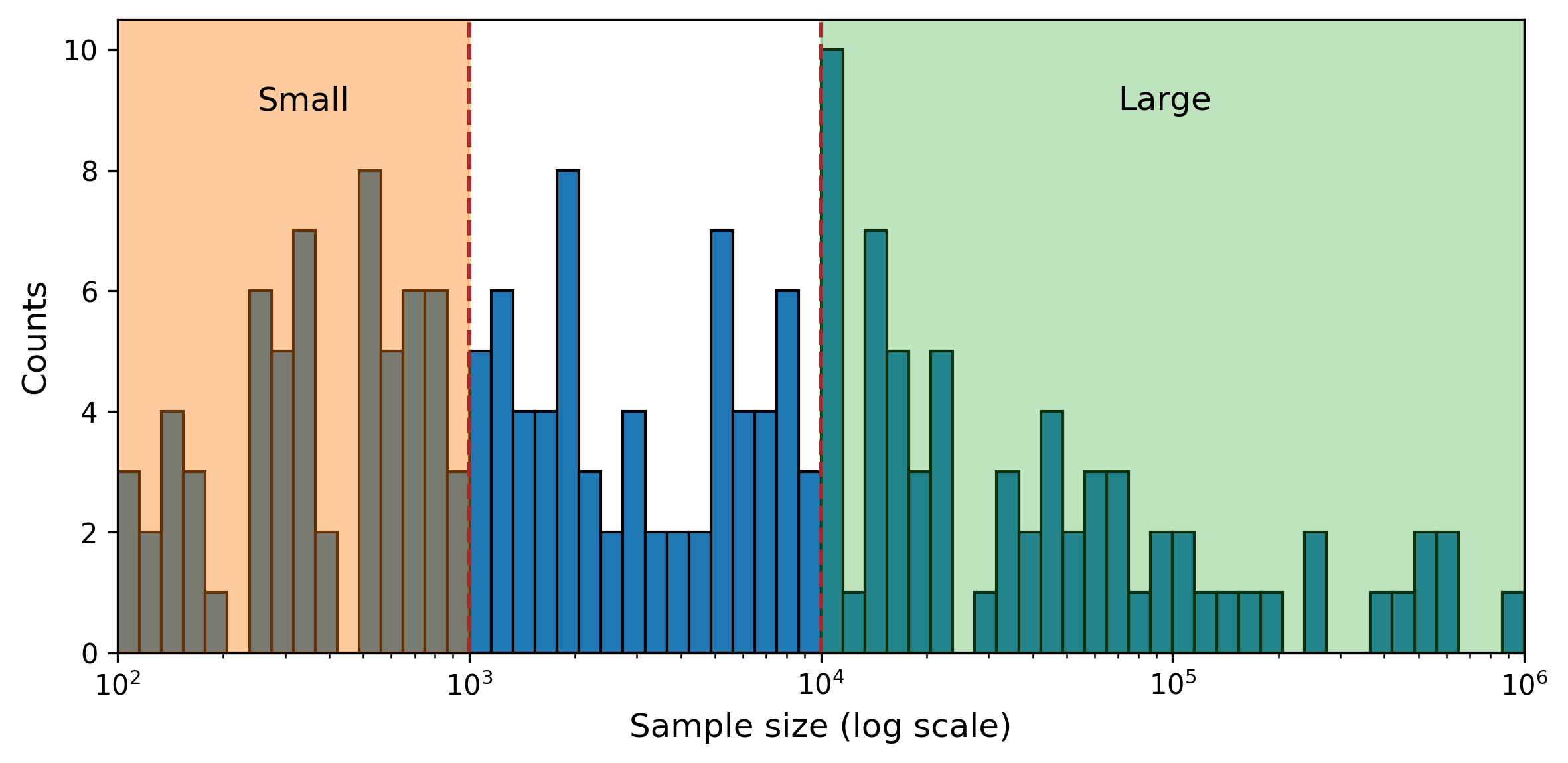}
    \label{fig:histogram:samplesize}
    }\hfill
    \subfloat[Proportion of categorical features]{
    \includegraphics[width=0.48\textwidth]{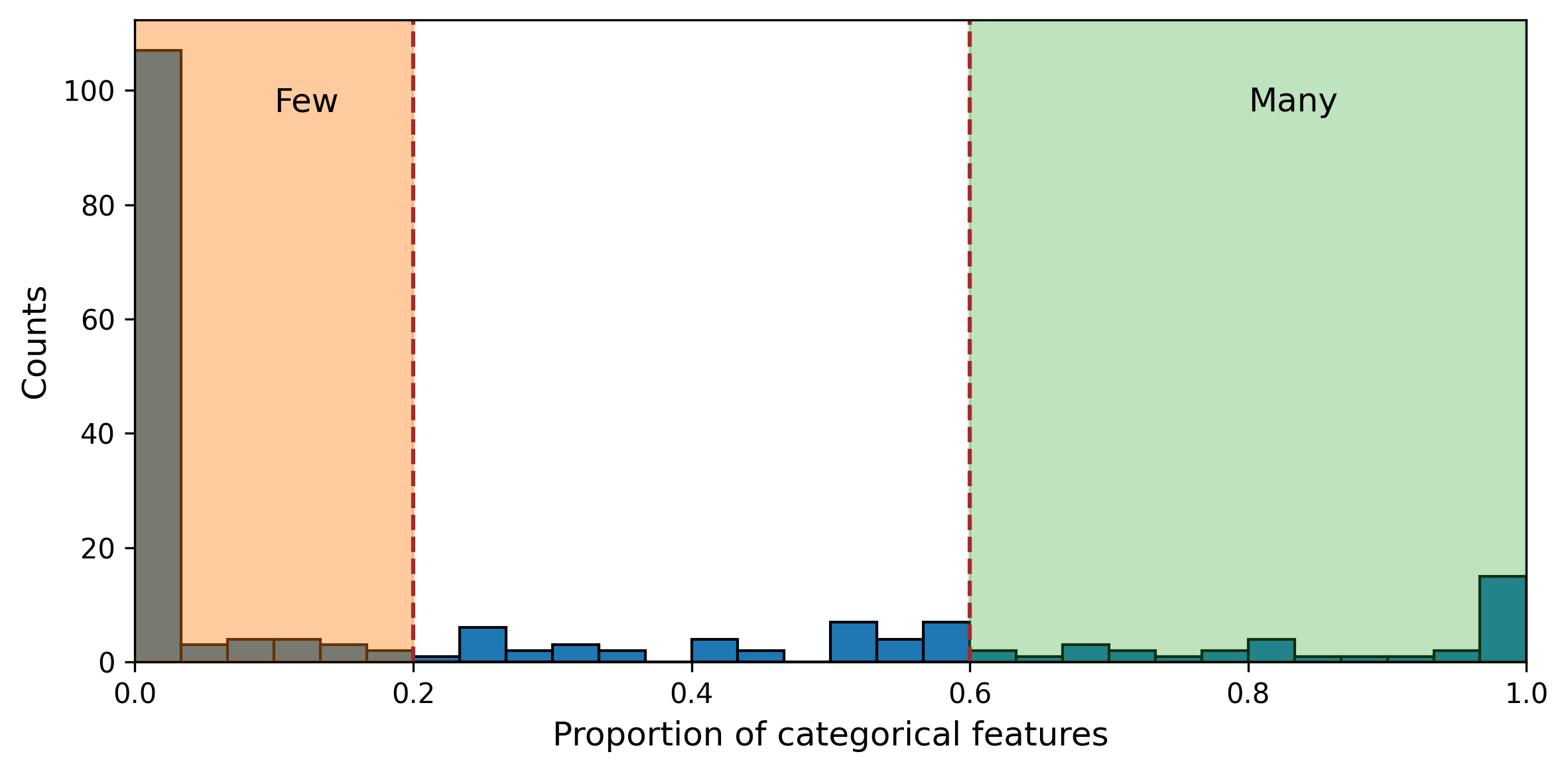}
    \label{fig:histogram:cat}
    }
    \\
    \subfloat[Average cardinality]{
    \includegraphics[width=0.48\textwidth]{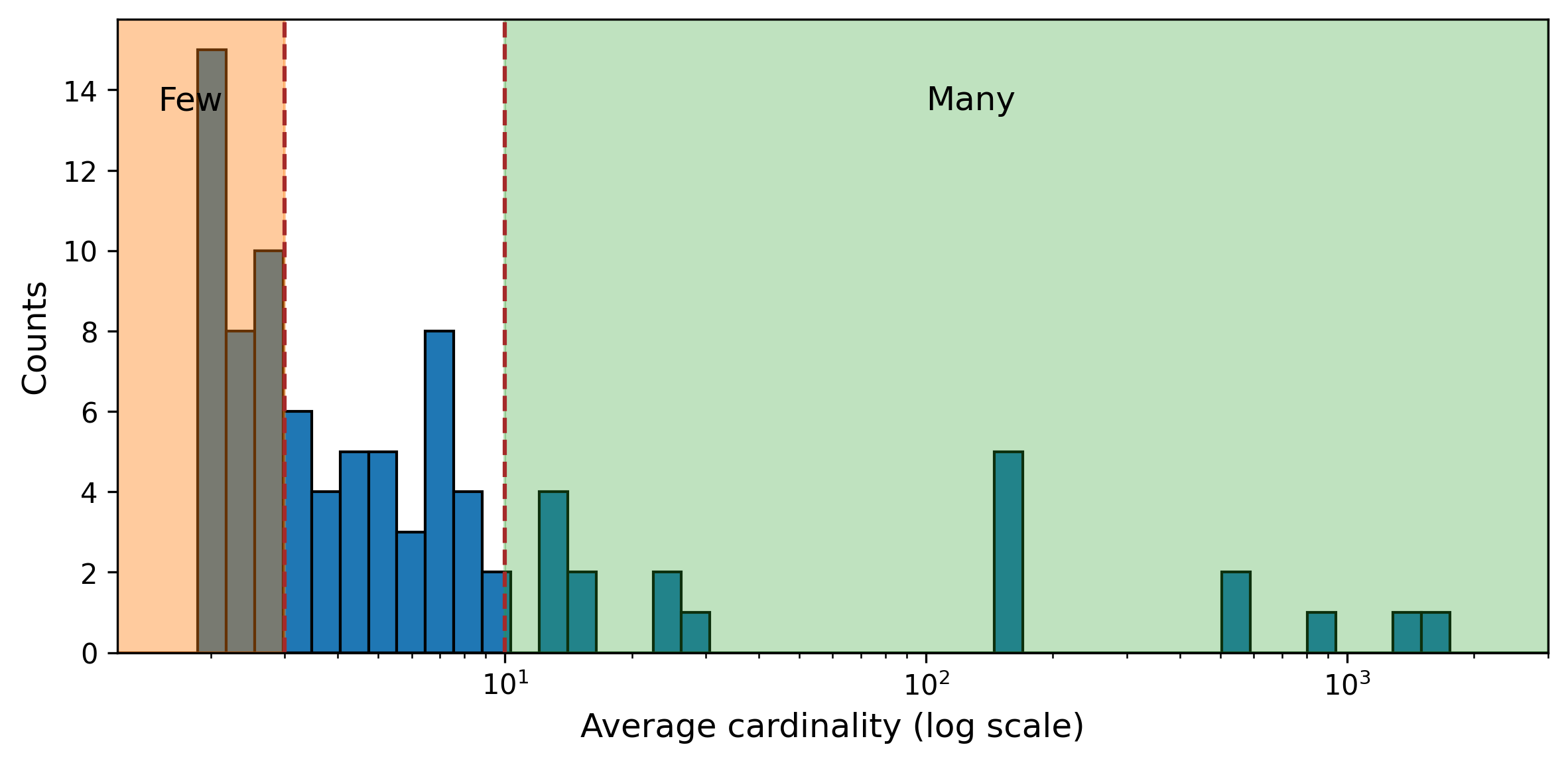}
    \label{fig:histogram:card}
    }\hfill
    \subfloat[Feature-to-sample ratio]{
    \includegraphics[width=0.48\textwidth]{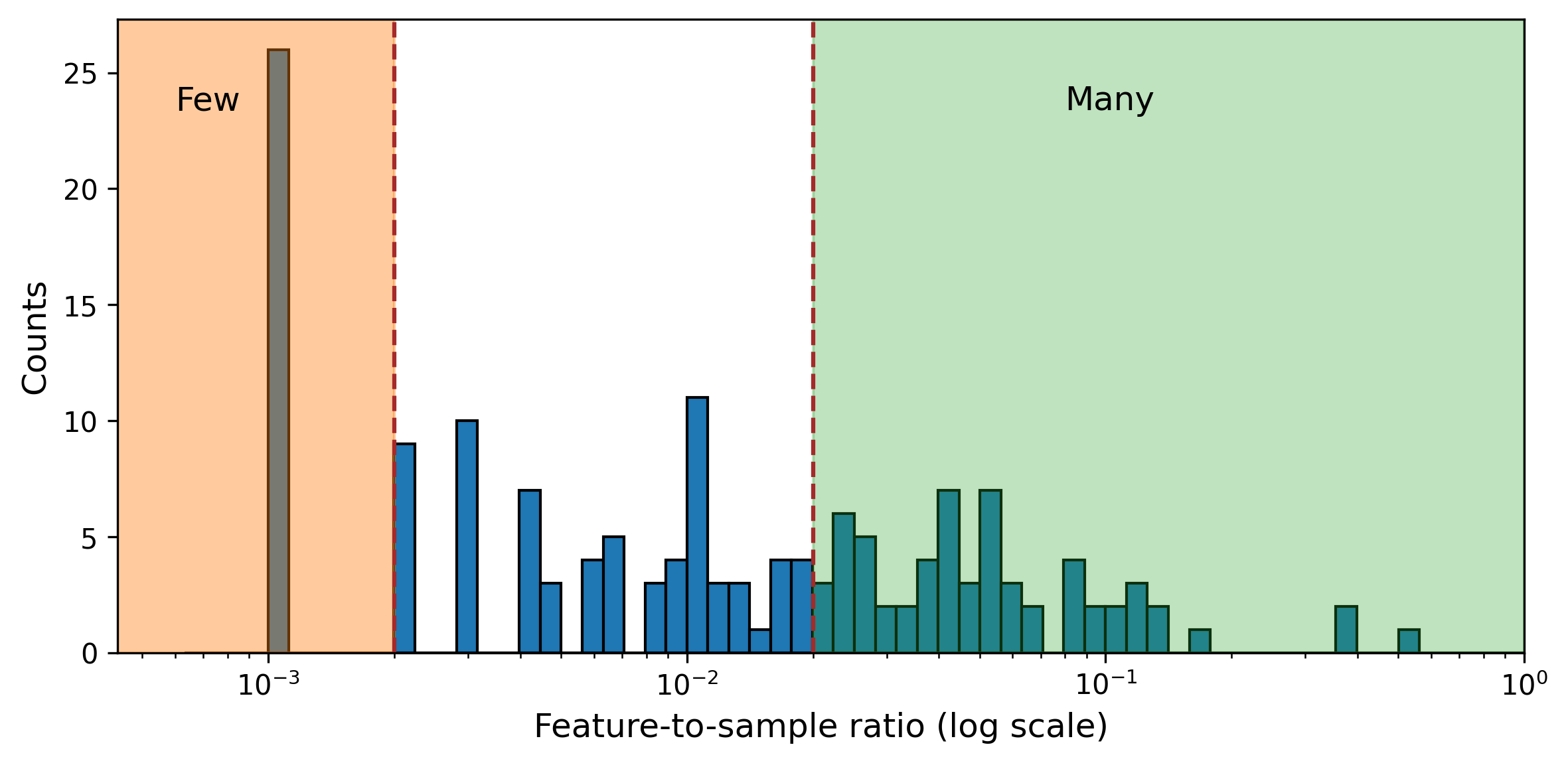}
    \label{fig:histogram:fsratio}
    } \\
    \subfloat[Entropy ratio]{
    \includegraphics[width=0.48\textwidth]{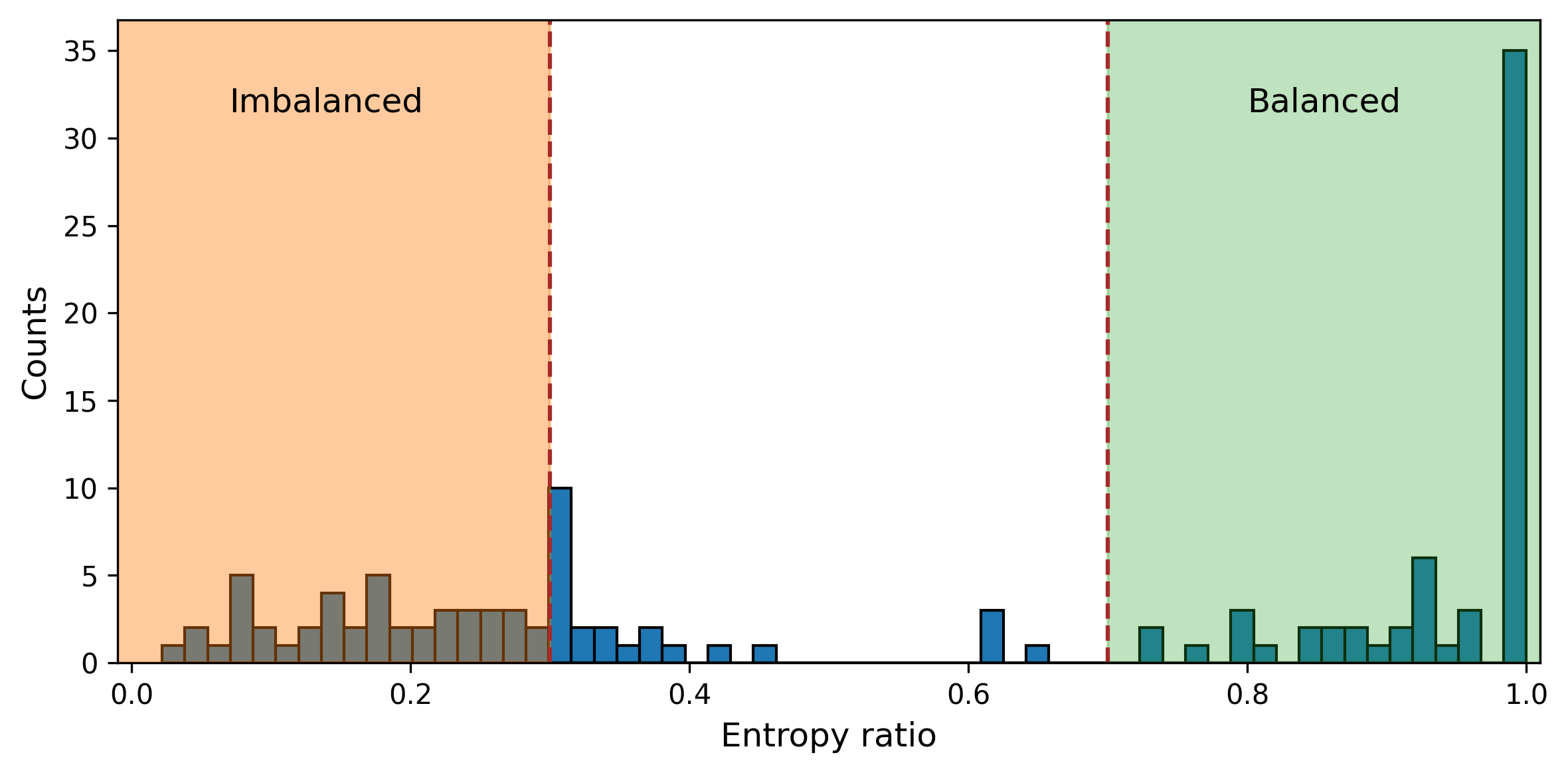}
    \label{fig:histogram:entropy}
    }\hfill
    \subfloat[Skewness]{
    \includegraphics[width=0.48\textwidth]{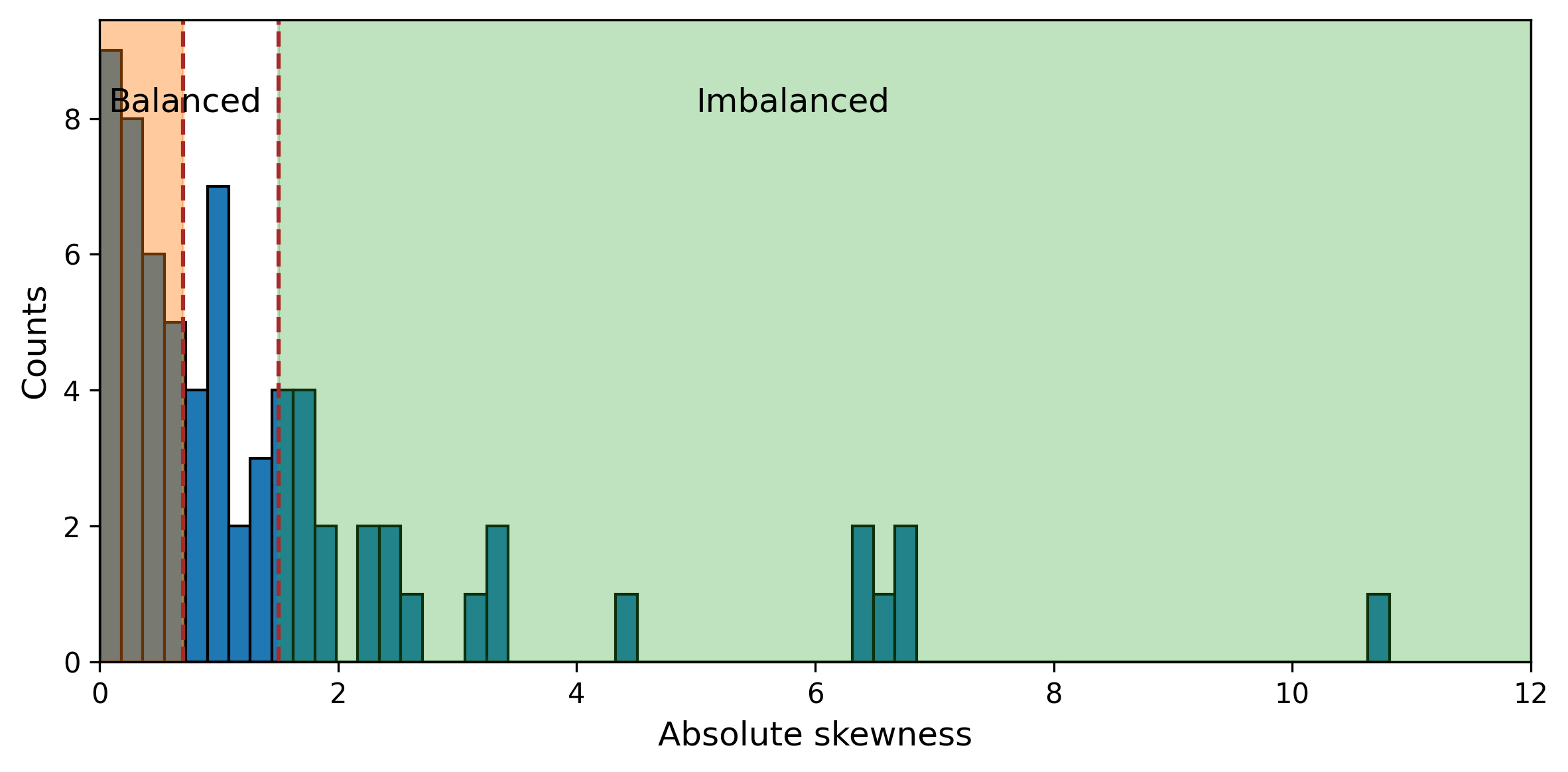}
    \label{fig:histogram:skew}
    } \\
    \subfloat[Imbalance factor]{
    \includegraphics[width=0.48\textwidth]{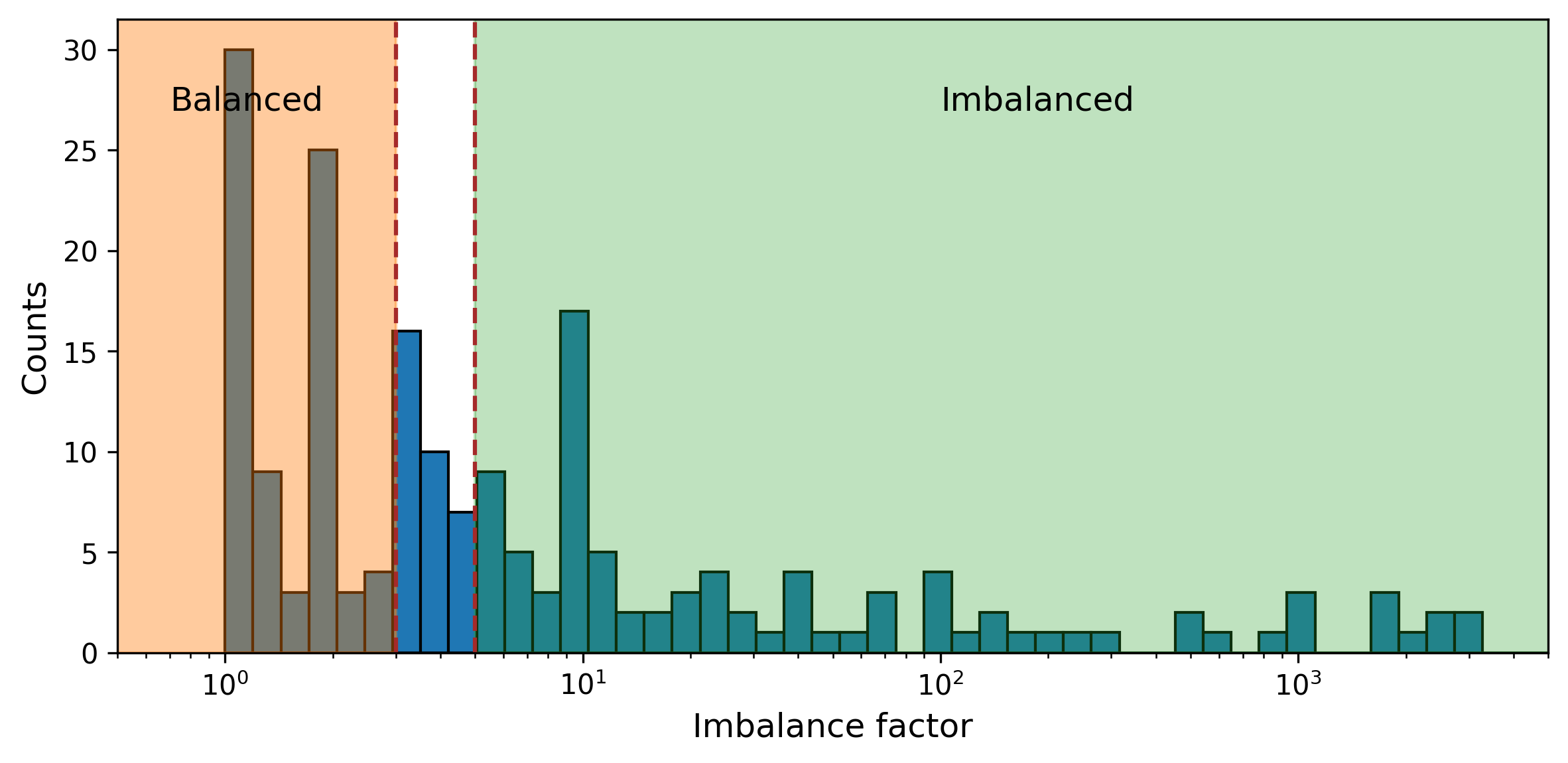}
    \label{fig:histogram:if}
    }\hfill
    \subfloat[Function irregularity]{
    \includegraphics[width=0.48\textwidth]{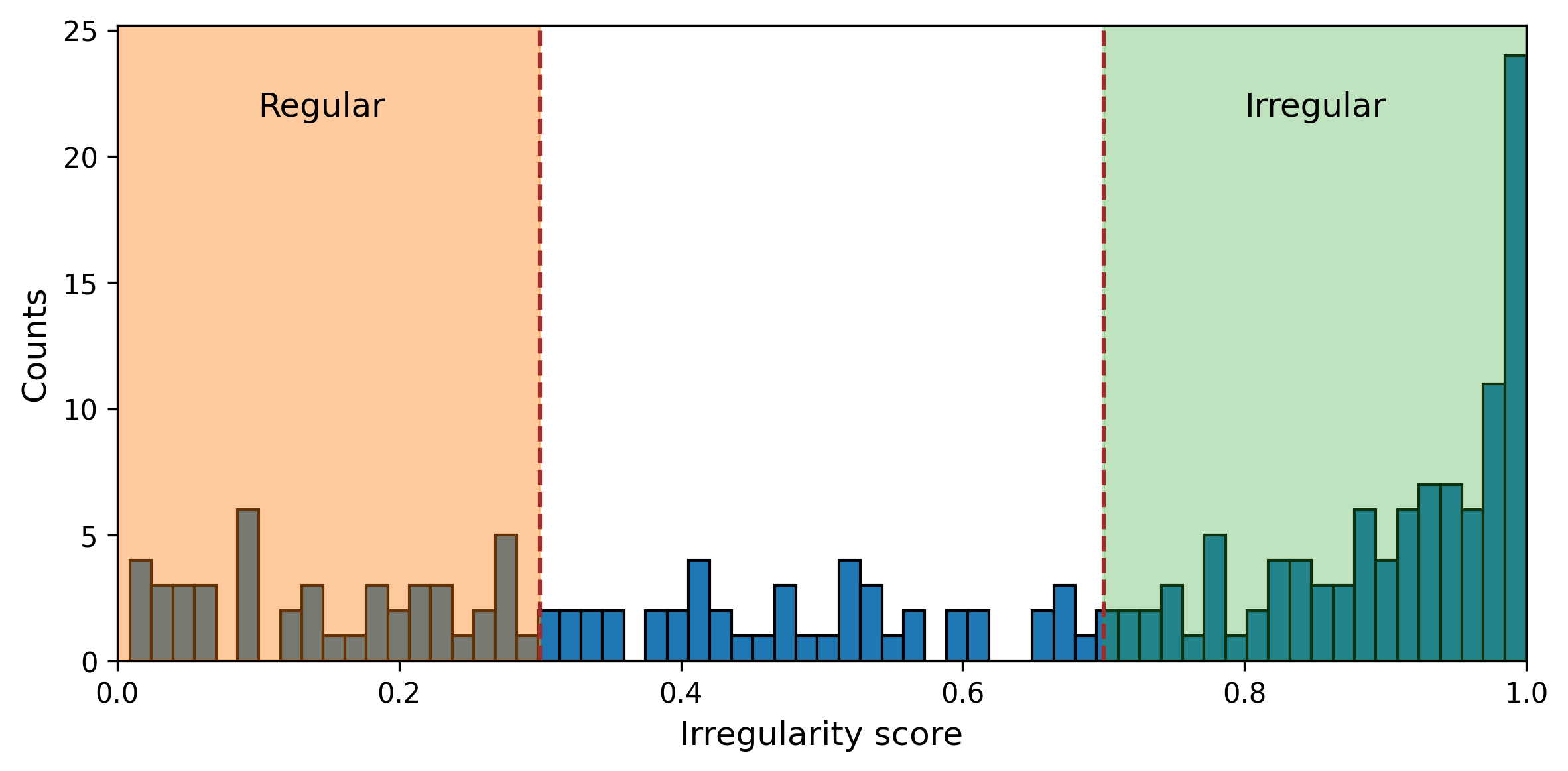}
    \label{fig:histogram:irregularity}
    } \\
    \subfloat[Inter-feature correlation]{
    \includegraphics[width=0.48\textwidth]{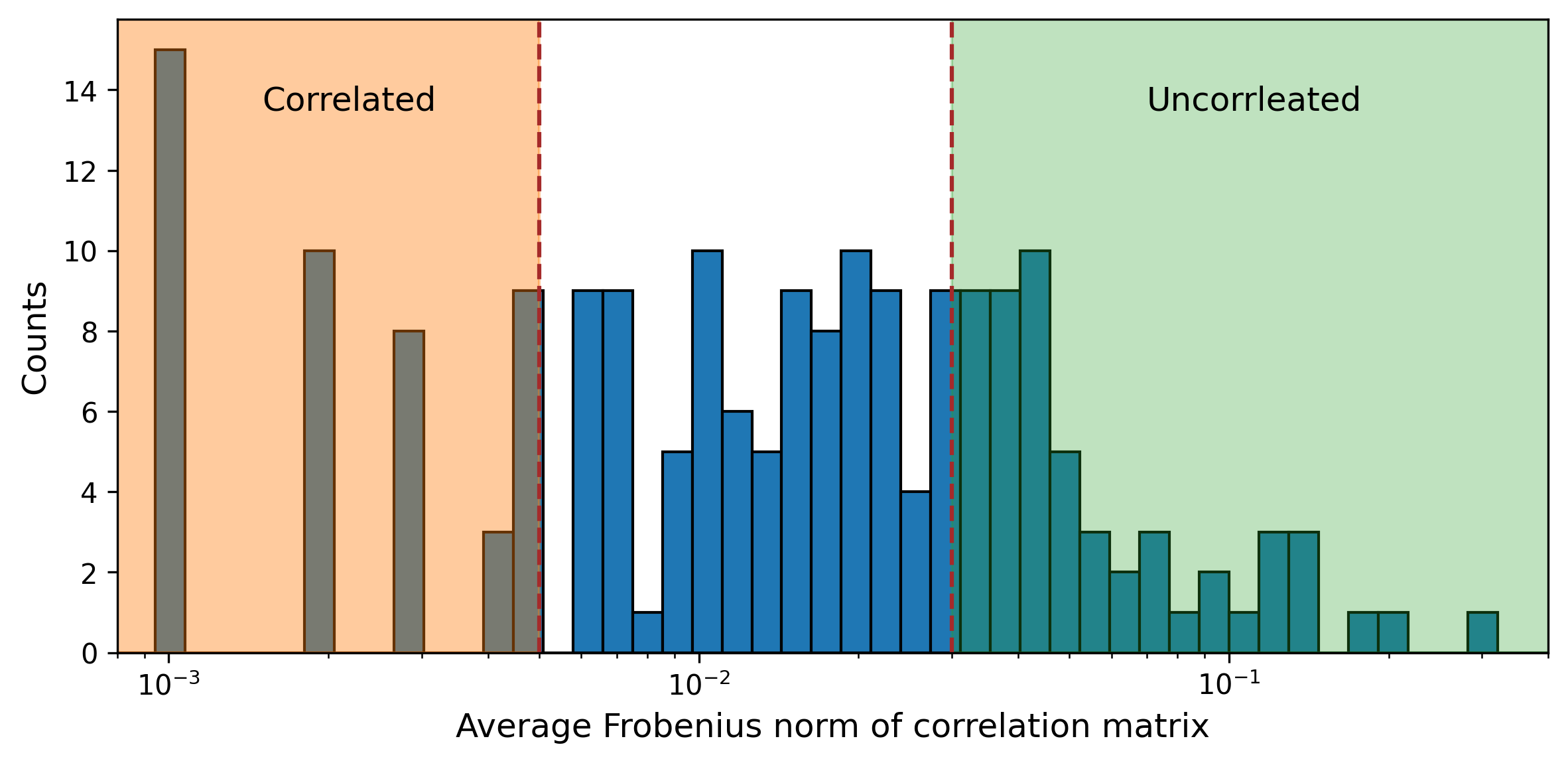}
    \label{fig:histogram:correlation}
    }\hfill
    \subfloat[Minimum eigenvalue]{
    \includegraphics[width=0.48\textwidth]{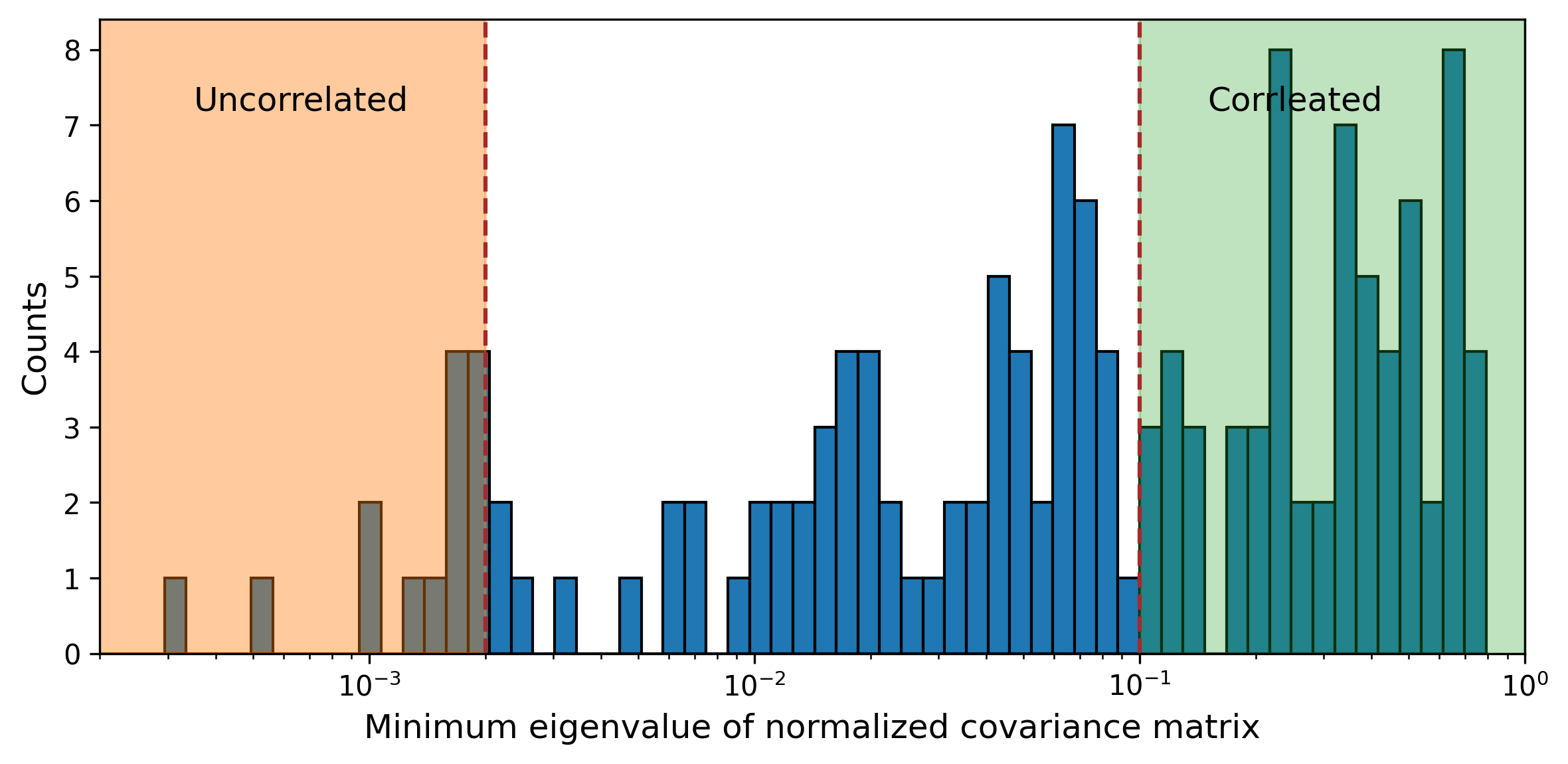}
    \label{fig:histogram:eigen}
    }
    \caption{Detailed histograms for each sub-category}
    \label{fig:histogram}
\end{figure}

The detailed histogram of minimum eigenvalue of the normalized covariance matrix is provided in Figure~\ref{fig:histogram:eigen}.

\clearpage

\clearpage

\subsection{Additional Data Statistics}

In addition to the seven main axes and 11 metrics analyzed in this study, our benchmark suite includes a variety of additional dataset statistics that may be useful for future research. These include metadata such as data domain (\emph{e.g.}, medical, financial, scientific), the number of target classes for classification tasks, and structural indicators derived from baseline model comparisons.
For example, we compute simple performance ratios between lightweight models, such as decision trees or $k$-nearest neighbors (KNN), and linear regression. These ratios serve as rough indicators of non-linearity or locality in the target function. A high decision tree-to-linear regression ratio may imply that the dataset benefits from non-linear partitioning, while a high KNN-to-linear ratio may suggest local continuity or manifold structure.

While these additional statistics are not directly analyzed in our study, we include them in the released benchmark suite. We hope that providing this enriched statistical context will facilitate deeper insights into model-data interactions in tabular learning.


\section{Detailed Description of the Algorithms}
\label{appendix:algorithms}

In this study, we evaluate 13 representative algorithms that span a wide range of architectural paradigms in tabular prediction. These models were selected to cover both classical methods and modern neural architectures, with particular emphasis on the types of inductive biases they encode and the data regimes they are likely to benefit from. Specifically, we categorize the models into six groups:

\begin{itemize}[leftmargin=15pt]
    \item \textbf{Classical method:} Random Forest~\citep{breiman2001random}, a widely used non-parametric ensemble baseline.
    \item \textbf{Gradient-Boosted Decision Trees (GBDTs):} XGBoost~\citep{chen2016xgboost}, CatBoost~\citep{prokhorenkova2018catboost}, and LightGBM~\citep{ke2017lightgbm}, known for strong performance across many tabular benchmarks.
    \item \textbf{NNs without structural priors (NN-Simple):} MLP, MLP-C, MLP-CN~\citep{gorishniy2022embeddings}, and ResNet~\citep{gorishniy2021revisiting}, which lack explicit mechanisms to model feature or sample dependencies.
    \item \textbf{NNs modeling inter-feature dependency (NN-Feature):} FT-Transformer~\citep{gorishniy2021revisiting} and T2G-Former~\citep{yan2023t2g}, which use attention to capture complex interactions among input features.
    \item \textbf{NNs modeling inter-sample dependency (NN-Sample):} TabR~\citep{gorishniy2023tabr} and ModernNCA~\citep{ye2024modern}, which leverage sample-level relationships through metric learning or retrieval objectives.
    \item \textbf{NNs modeling both feature and sample dependencies (NN-Both):} SAINT~\citep{somepalli2021saint}, which employs both row-wise and column-wise attention.
\end{itemize}

This categorization reflects diverse inductive biases, such as feature-level attention, latent similarity, and input-level encoding, which interact differently with key dataset characteristics.

To ensure fair and representative comparisons across models, we conducted extensive hyperparameter optimization for all algorithms using the Tree-structured Parzen Estimator (TPE)\citep{bergstra2011algorithms} as implemented in the Optuna Python library\citep{optuna2019}. Each algorithm was optimized over 100 trials without exception.

Our hyperparameter search spaces were primarily informed by prior work~\citep{gorishniy2021revisiting, gorishniy2022embeddings, yan2023t2g}, and extended to include key neural network parameters such as learning rate schedules, normalization strategies, activation functions, and optimizer configurations. These additions enable a more flexible and robust evaluation of architectural inductive biases.
All experiments were conducted using a single NVIDIA GeForce RTX 3090 GPU. The detailed search spaces and design motivations for each algorithm are provided following subsections.

To ensure comparability, we apply an ensemble of the top 5 hyperparameter configurations for all neural models, averaging their predictions at inference time. While GBDTs intrinsically ensemble multiple trees during training, neural networks do not benefit from this structural redundancy and are more sensitive to initialization and training stochasticity. As such, ensembling is a widely adopted practice in tabular deep learning studies~\citep{gorishniy2021revisiting, somepalli2021saint}, and provides a fairer comparison across model families.

\subsection{Classical Models: Random Forest}

Classical models serve as baseline approaches in tabular learning, providing a reference point against which the performance of more complex architectures can be evaluated. These models typically do not incorporate deep architectural inductive biases and are favored for their interpretability, stability, and low training cost. While classical linear models such as logistic regression are commonly used in traditional machine learning pipelines, we exclude them from our analysis due to consistently poor performance across benchmarks.

In this category, we include Random Forests, which are widely used ensemble-based models in tabular domains. They require minimal preprocessing, handle mixed-type inputs naturally, and are known to perform robustly under a variety of data conditions. As such, they represent a strong non-neural baseline for comparison.

\subsubsection{Random Forest}

Random Forest is a bagging-based ensemble method composed of decision trees trained on bootstrapped samples with randomized feature selection. This design promotes variance reduction while maintaining high expressivity through nonlinear decision boundaries. Despite its simplicity, Random Forest is a competitive baseline in many tabular tasks. However, it typically requires CPU-based training, which can result in longer runtimes compared to GPU-accelerated methods.

Its key inductive bias lies in its approximation of functions via piecewise constant splits, making it particularly effective for modeling irregular or discontinuous relationships in the input space. 

For implementation, we use the \texttt{RandomForestClassifier} and \texttt{RandomForestRegressor} implementations from \texttt{scikit-learn}.
Hyperparameter search space has been adopted from \cite{salinas2023tabrepo} as follows.

\begin{table}[h!]
\centering
    \caption{Hyperparameter search space for Random Forest}
    \label{hpspace:randomforest}
    \resizebox{0.7\textwidth}{!}{
    \begin{tabular}{cl}
    \toprule
    Hyperparameter        & Search space          \\\midrule
    max\_leaf\_nodes             & UniformInt(5000, 50000)     \\
    min\_samples\_leaf      & UniformCat([1, 2, 3, 4, 5, 10, 20, 40, 80]) \\
    max\_features             & UniformCat([\texttt{sqrt}, \texttt{log2}, 0.5, 0.75, 1.0])       \\
    n\_estimators          & 300 \\\bottomrule
    \end{tabular}
    }
\end{table}

\subsection{GBDTs: XGBoost, CatBoost, LightGBM}

Gradient-Boosted Decision Trees (GBDTs) are widely regarded as state-of-the-art methods for tabular data. They sequentially build an ensemble of shallow decision trees, where each new tree fits the residual error of the previous ensemble. GBDTs encode strong inductive biases such as hierarchical feature splitting, robustness to missing values, and implicit variable selection through split gain metrics. These properties make them well-suited for sparse, noisy, and heterogeneous tabular domains.

While conceptually similar, XGBoost, CatBoost, and LightGBM differ in their optimization strategies, tree construction methods, and handling of categorical features. We briefly summarize each below.

\subsubsection{XGBoost}

XGBoost~\citep{chen2016xgboost} is a highly optimized and regularized version of GBDT that employs exact or approximate split finding, supports column subsampling, and allows flexible regularization through $\ell_1$ and $\ell_2$ penalties. It is known for its speed and scalability, but does not natively handle categorical features, requiring preprocessing such as label encoding.

For implementation, we use the \texttt{xgboost} Python package.
Hyperparameter search space has been adopted from \cite{salinas2023tabrepo} as follows.

\begin{table}[h!]
\centering
    \caption{Hyperparameter search space of XGBoost}
    \label{hpspace:xgboost}
    \resizebox{0.55\textwidth}{!}{
    \begin{tabular}{cl}
    \toprule
    Hyperparameter & Search space \\\midrule
    max\_depth & UniformInt(4, 10)     \\
    min\_child\_weight & Uniform(0.5, 1.5) \\
    learning\_rate & LogUniform($5e^{-3}$, 0.1)   \\
    colsample\_bytree & Uniform(0.5, 1) \\
    enable\_categorical & UniformCat([True, False]) \\
    early\_stopping\_rounds & 20 \\
    n\_estimators & 10000 \\
    max\_iterations & 10000 \\
    \bottomrule
    \end{tabular}
    }
\end{table}

\subsubsection{CatBoost}

CatBoost~\citep{prokhorenkova2018catboost} introduces ordered boosting and efficient categorical feature encoding via target statistics, which improves stability and reduces overfitting. It is particularly effective when datasets contain many categorical features, and it does not require explicit encoding during preprocessing.
Unlike XGBoost, CatBoost includes built-in categorical processing, which we leverage directly without any additional transformation. 

For implementation, we use the \texttt{catboost} Python package.
Hyperparameter search space has been adopted from \cite{salinas2023tabrepo} and \cite{mcelfresh2024neural} as follows.

\begin{table}[h!]
    \centering
    \caption{Hyperparameter search space of CatBoost}
    \label{hpspace:catboost}
    \resizebox{0.7\textwidth}{!}{
    \begin{tabular}{cl}
    \toprule
    Hyperparameter             & Search space             \\\midrule
    max\_depth & UniformInt(4, 8) \\ 
    learning\_rate & LogUniform(5$e^{-3}$, 0.1) \\ 
    max\_ctr\_complexity & Uniform(1, 5) \\ 
    l2\_leaf\_reg & Uniform(1, 5) \\ 
    grow\_policy & UniformCat([SymmetricTree, Depthwise]) \\ 
    early\_stopping\_rounds & 20 \\ 
    iterations & 10000 \\ 
    one\_hot\_max\_size & UniformCat([2, 3, 5, 10]) \\\bottomrule 
    \end{tabular}
    }
\end{table}

\subsubsection{LightGBM}

LightGBM~\citep{ke2017lightgbm} is designed for efficiency on large datasets through histogram-based gradient computation and leaf-wise tree growth with depth constraints. It also supports categorical feature handling natively, though with less advanced treatment compared to CatBoost.
It is generally faster than XGBoost and CatBoost on large, high-dimensional datasets. 

For implementation, we use the \texttt{lightgbm} Python package.
Hyperparameter search space has been adopted from \cite{salinas2023tabrepo} and \cite{mcelfresh2024neural} as follows.

\begin{table}[h!]
    \centering
    \caption{Hyperparameter search space of LightGBM}
    \label{hpspace:lightgbm}
    \resizebox{0.45\textwidth}{!}{
    \begin{tabular}{cl}
    \toprule
    Hyperparameter & Search space \\\midrule
    learning\_rate & LogUniform(5$e^{-3}$, 0.1) \\  
    num\_leaves & UniformInt(16, 255)   \\  
    min\_data\_in\_leaf & UniformInt(2, 60) \\  
    feature\_fraction & Uniform(0.4, 1) \\  
    extra\_trees & UniformCat([False, True]) \\  
    early\_stopping\_rounds & 20 \\ 
    iterations & 10000 \\\bottomrule 
    \end{tabular}
    }
\end{table}

\subsection{NN-Simple: MLP, MLP-C, MLP-CN, ResNet}

This category includes models that do not explicitly encode feature- or sample-level interactions through architectural mechanisms. Instead, they rely on fully connected layers to model global dependencies. While simple and flexible, these models can be vulnerable to overfitting, particularly in sparse or irregular tabular regimes.

We evaluate four such models: (1) a vanilla multilayer perceptron (MLP), (2) MLP-C, which augments the MLP with categorical embeddings, (3) MLP-CN, which adds numerical embeddings to enhance robustness to irregular patterns in inputs, and (4) ResNet, which introduces residual connections for better gradient flow. These architectures serve as strong baselines in many recent tabular learning benchmarks~\citep{gorishniy2021revisiting, gorishniy2022embeddings,holzmuller2024better}.

\subsubsection{MLP}

The MLP consists of a stack of linear layers, each followed by normalization, activation, and dropout. It receives as input the concatenation of raw numerical features and label-encoded categorical features. As it lacks any inductive biases specific to tabular structure, its performance heavily depends on proper regularization and optimization.

For implementation, we mainly follow the implementation of \citep{gorishniy2021revisiting}.
Hyperparameter search space has been adopted from \cite{salinas2023tabrepo} and \cite{gorishniy2021revisiting} as follows.

\begin{table}[h!]
    \centering
    \caption{Hyperparameter search space of MLP (* indicates newly added hyperparameters in this study.)}
    \label{hpspace:mlp}
    \resizebox{0.8\textwidth}{!}{
    \begin{tabular}{cl}
    \toprule
    Hyperparameter & Search space \\\midrule
    depth & UniformInt(1, 8) \\
    width & UniformInt(1, 512) \\
    dropout & Uniform(0, 0.5) \\
    learning\_rate & Uniform(1e-5, 1e-2) \\
    lr\_scheduler$^*$ & UniformCat([True, False]) \\
    weight\_decay & Uniform(1$e^{-6}$, 1$e^{-3}$) \\
    normalization$^*$ & UniformCat([None, BatchNorm, LayerNorm]) \\
    activation$^*$ & UniformCat([ReLU, LReLU, Sigmoid, Tanh, GeLU]) \\
    n\_epochs & 100 \\
    optimizer$^*$ & UniformCat([AdamW, Adam, SGD]) \\\bottomrule
    \end{tabular}
    }
\end{table}

\subsubsection{MLP-C}

MLP-C augments the vanilla MLP by replacing label-encoded categorical features with learnable embeddings. 
We adopt the categorical embedding of FT-Transformer~\citep{gorishniy2021revisiting} to MLP to investigate the role of categorical embedding modules under the same deep architecture of MLP. 
These embeddings are concatenated with normalized numerical features and passed through the MLP backbone. The goal is to provide dense, trainable representations of discrete features, reducing sparsity and improving optimization.

We retain the same hyperparameter space as MLP, with the following additional parameter:
\begin{itemize}[leftmargin=15pt]
    \item \texttt{d\_embedding}: UniformInt(64, 512)
\end{itemize}

\subsubsection{MLP-CN}

MLP-CN further extends MLP-C by embedding both categorical and numerical features~\citep{gorishniy2022embeddings}. Numerical embeddings are implemented as learnable per-feature projections, and the resulting representations are concatenated with categorical embeddings. 
This design enhances feature-level expressivity and is particularly motivated by the need to address irregularity in tabular inputs, as discussed in \citep{gorishniy2022embeddings}.

For implementation, we applied \texttt{PeriodicEmbeddings} from \texttt{rtdl\_num\_embeddings} python library. We retain the same hyperparameter space as MLP-C, with the following additional parameter:
\begin{itemize}[leftmargin=15pt]
    \item \texttt{d\_embedding\_num}: UniformInt(1, 128)
\end{itemize}

\subsubsection{ResNet}

This model applies skip connections between linear blocks to improve optimization, following the \texttt{ResNet} design proposed in \citep{gorishniy2021revisiting}. It uses the same input representation as MLP, without feature embeddings. Residual connections help mitigate vanishing gradients and stabilize deeper architectures. This makes ResNet more stable on deep architectures, particularly when trained on large or noisy datasets.

For implementation, we mainly follow the implementation of \citep{gorishniy2021revisiting}.
Hyperparameter search space has been adopted from \cite{salinas2023tabrepo} and \cite{gorishniy2021revisiting} as follows.

\begin{table}[h!]
\centering
    \caption{Hyperparameter search space of ResNet (* indicates newly added hyperparameters in this study.)}
    \label{hpspace:resnet}
    \resizebox{0.7\textwidth}{!}{
    \begin{tabular}{cl}
    \toprule
    Hyperparameter & Search space \\\midrule
    n\_layers & UniformInt(1, 8) \\
    d & UniformInt(64, 512) \\
    d\_embedding & UniformInt(64, 512) \\
    d\_hidden\_factor & Uniform(1, 4) \\
    hidden\_dropout & Uniform(0, 0.5) \\
    residual\_dropout & Uniform(0, 0.5) \\
    learning\_rate & Uniform(1$e^{-5}$, 1$e^{-2}$) \\
    lr\_scheduler$^*$ & UniformCat([True, False]) \\
    weight\_decay & Uniform(1$e^{-6}$, 1$e^{-3}$) \\
    normalization$^*$ & UniformCat([None, BatchNorm, LayerNorm]) \\
    activation$^*$ & UniformCat([ReLU, LReLU, Sigmoid, Tanh, GeLU]) \\
    n\_epochs & 100 \\
    optimizer$^*$ & UniformCat([AdamW, Adam, SGD]) \\\bottomrule
    \end{tabular}
    }
\end{table}

\subsection{NN-Feature: FT-Transformer, T2G-Former}

This category includes neural architectures that model inter-feature dependencies through attention mechanisms. Unlike MLPs, which treat input features independently aside from shared weights, these models introduce explicit mechanisms to capture pairwise or global relationships among features. Such inductive biases can be particularly beneficial in settings with structured feature correlations or hierarchical dependencies.

\subsubsection{FT-Transformer}

FT-Transformer~\citep{gorishniy2021revisiting} applies a standard Transformer encoder to the tabular setting by embedding each input feature and applying multi-head self-attention across the embedded features. Unlike traditional Transformers used in NLP, which attend across tokens (rows), FT-Transformer attends across columns (features), enabling it to capture global dependencies between features. It uses categorical embeddings for discrete features and concatenates them with embedded numerical inputs.

This architecture is particularly effective when feature interactions are complex or nonlinear. However, it may suffer from overfitting in low-sample or highly sparse regimes due to its capacity and lack of explicit regularization.

We follow the implementation from \texttt{rtdl}~\citep{gorishniy2021revisiting} and adopt the hyperparameter space as follows.

\begin{table}[h!]
    \centering
    \caption{Hyperparameter search space of FT-Transformer (* indicates newly added hyperparameters in this study.)}
    \label{hpspace:ftt}
    \resizebox{0.75\textwidth}{!}{
    \begin{tabular}{cl}
    \toprule
    Hyperparameter & Search space \\\midrule
    token\_bias & UniformCat([True, False]) \\
    n\_layers & UniformInt(1, 4) \\
    d\_token & UniformInt(8, 64) \\
    n\_heads & 8 \\
    d\_ffn\_factor & Uniform($\frac{2}{3}$, $\frac{8}{3}$) \\
    attention\_dropout & Uniform(0, 0.5) \\
    ffn\_dropout & Uniform(0, 0.5) \\
    residual\_dropout & Uniform(0, 0.2) \\
    activation & UniformCat([reglu, geglu, Sigmoid, ReLU]) \\
    prenormalization & UniformCat([True, False]) \\
    initialization & UniformCat([Xavier, Kaiming]) \\
    kv\_compression & None \\
    kv\_compression\_sharing & None \\
    learning\_rate & Uniform(1$e^{-5}$, 1$e^{-3}$) \\
    lr\_scheduler$^*$ & UniformCat([True, False]) \\
    weight\_decay & Uniform(1$e^{-6}$, 1$e^{-3}$) \\
    n\_epochs & 100 \\
    optimizer$^*$ & UniformCat([AdamW, Adam, SGD]) \\\bottomrule
    \end{tabular}
    }
\end{table}

\clearpage

\subsubsection{T2G-Former}

T2G-Former~\citep{yan2023t2g} is a Transformer-based architecture tailored for tabular data, specifically designed to model heterogeneous feature interactions. Unlike generic attention mechanisms that treat all features equally, T2G-Former dynamically estimates relationships between features using a dedicated Graph Estimator module. This module constructs a Feature Relation Graph (FR-Graph), where edges represent learned dependencies between feature pairs.

Each block applies attention operations over the FR-Graph, allowing the model to selectively focus on informative feature pairs while suppressing spurious interactions. In addition, a Cross-Level Readout mechanism aggregates important signals from all blocks to form a global representation for final prediction.

This architecture encodes a strong inductive bias toward learning structured and selective inter-feature dependencies, making it particularly well-suited for high-dimensional or redundant tabular datasets. By combining graph-based structure learning with feature-level attention, T2G-Former seeks to overcome the limitations of standard Transformers when applied to heterogeneous and sparse tabular domains.

We adopt the implementation and hyperparameter search space proposed in \citep{yan2023t2g}.

\begin{table}[h!]
\centering
    \caption{Hyperparameter search space of T2G-Former (* indicates newly added hyperparameters in this study.)}
    \label{hpspace:t2g}
    \resizebox{0.75\textwidth}{!}{
    \begin{tabular}{cl}
    \toprule
    Hyperparameter & Search space \\\midrule
    n\_layers & UniformInt(1, 5) \\
    d\_token & UniformInt(8, 64) \\
    n\_heads & 8 \\
    residual\_dropout & Uniform(0, 0.2) \\
    attention\_dropout & Uniform(0, 0.5) \\
    ffn\_dropout & Uniform(0, 0.5) \\
    learning\_rate & LogUniform(1$e^{-5}$, 1$e^{-3}$) \\
    learning\_rate\_embed & LogUniform(5$e^{-3}$, 5$e^{-2}$) \\
    token\_bias & True \\
    kv\_compression & None \\
    kv\_compression\_sharing & None \\
    d\_ffn\_factor & Uniform(2/3, 8/3) \\
    prenormalization & UniformCat([True, False]) \\
    initialization & UniformCat([Xavier, Kaiming]) \\
    activation & UniformCat([reglu, geglu, Sigmoid, ReLU]) \\
    lr\_scheduler$^*$ & UniformCat([True, False]) \\
    weight\_decay$^*$ & Uniform(1$e^{-6}$, 1$e^{-3}$) \\
    n\_epochs & 100 \\
    optimizer$^*$ & UniformCat([AdamW, Adam, SGD]) \\\bottomrule          
    \end{tabular}
    }
\end{table}

\subsection{NN-Sample: TabR, ModernNCA}

This category includes models that explicitly model relationships between samples in the dataset, leveraging techniques such as retrieval-based attention or metric learning. Unlike architectures that focus solely on intra-feature interactions, these models aim to capture structure that emerges at the sample level, such as local neighborhoods, prototype similarity, or distance-based organization. This inductive bias is particularly beneficial in datasets where class structure or decision boundaries are governed by relational properties between instances, such as clusters or manifolds in latent space.

\subsubsection{TabR}

TabR~\citep{gorishniy2023tabr} is a retrieval-based neural network that augments classical feed-forward backbones with a learnable memory module. The core idea is to construct a retrieval set--a small set of trainable vectors that serve as prototypes or representatives for learning. During training, each sample interacts with this set via attention, effectively retrieving relevant patterns from memory to inform the prediction.

TabR applies two types of attention: (1) Sample-to-retrieval attention: the input sample queries the retrieval set to gather contextual signals; 
(2) Retrieval-to-sample attention: retrieval entries query the input to refine their embedding. 
These dual interactions allow TabR to build a bidirectional link between training examples and shared latent representations. This architecture is especially advantageous when sample-level regularities (e.g., shared structure among rare classes or local clusters) play a critical role. It is less effective in datasets where samples are independently and identically distributed without strong relational signals.

We adopt the implementation from the TALENT library and tune hyperparameters based on prior works~\citep{gorishniy2023tabr, ye2024modern}.

\begin{table}[h!]
    \centering
    \caption{Hyperparameter search space of TabR (* indicates newly added hyperparameters in this study.)}
    \label{hpspace:saint}
    \resizebox{0.75\textwidth}{!}{
    \begin{tabular}{cl}
    \toprule
    Hyperparameter & Search space \\\midrule
    d\_main & UniformInt(96, 384) \\
    context\_dropout & Uniform(0, 0.6) \\ 
    encoder\_n\_blocks & UniformInt(0, 1) \\
    predictor\_n\_blocks & UniformInt(1, 2) \\
    dropout0 & Uniform(0, 0.6) \\
    d\_embedding & UniformInt(16, 64) (UniformInt(8, 32) for large data) \\
    frequency\_scale & LogUniform(0.01, 100) \\
    n\_frequencies & UniformInt(16, 96) \\
    learning\_rate & Uniform(1$e^{-5}$, 1$e^{-3}$) \\
    lr\_scheduler$^*$ & UniformCat([True, False]) \\
    weight\_decay$^*$ & Uniform(1$e^{-6}$, 1$e^{-3}$) \\
    n\_epochs & 100 \\
    optimizer$^*$ & UniformCat([AdamW, Adam, SGD]) \\\bottomrule
    \end{tabular}
    }
\end{table}

\subsubsection{ModernNCA}

ModernNCA~\citep{ye2024modern} is a metric-learning-based model that learns to embed samples in a latent space where semantically similar instances are close together. Inspired by Neighbourhood Components Analysis (NCA), ModernNCA optimizes the probability that each sample’s embedding is close to those of the same class (for classification) or exhibits small distance to similar target values (for regression).

Unlike classical NCA, which is shallow and hard to scale, ModernNCA is implemented as a deep neural network with optional normalization, nonlinearities, and dropout. Its key inductive bias lies in assuming a smooth and clusterable latent geometry: samples from similar classes or output ranges should form compact groups in embedding space.

This assumption can be especially effective under class imbalance, weak supervision, or noisy labels, where contrastive objectives can stabilize training. However, performance may degrade when the target structure is not well captured by latent distances.

We use the official implementation and default tuning range provided in the TALENT python library~\citep{ye2024modern}.

\begin{table}[h!]
    \centering
    \caption{Hyperparameter search space of ModernNCA (* indicates newly added hyperparameters in this study.)}
    \label{hpspace:saint}
    \resizebox{0.75\textwidth}{!}{
    \begin{tabular}{cl}
    \toprule
    Hyperparameter & Search space \\\midrule
    d\_block & UniformInt(64, 1024) (UniformInt(64, 128) for large data)\\
    dim & UniformInt(64, 1024) (UniformInt(64, 128) for large data)\\ 
    dropout & Uniform(0, 0.5) \\
    n\_blocks & UniformInt(0, 2) \\
    d\_embedings & UniformInt(16, 64) (UniformInt(8, 32) for large data)\\ 
    frequency\_sclae & LogUniform(0.005, 10) \\
    n\_frequencies & UniformInt(16, 96) \\
    learning\_rate & Uniform(1$e^{-5}$, 1$e^{-3}$) \\
    lr\_scheduler$^*$ & UniformCat([True, False]) \\
    weight\_decay$^*$ & Uniform(1$e^{-6}$, 1$e^{-3}$) \\
    n\_epochs & 100 \\
    optimizer$^*$ & UniformCat([AdamW, Adam, SGD]) \\\bottomrule
    \end{tabular}
    }
\end{table}

\clearpage

\subsection{NN-Both: SAINT}

This category includes models that simultaneously model inter-feature and inter-sample dependencies, allowing for richer context modeling in both input dimensions. By combining these two axes of inductive bias, such architectures are designed to capture the complex structure of tabular data—particularly useful when dependencies exist both across columns and between rows.

\subsubsection{SAINT}

SAINT (Self-Attention and Intersample Attention Transformer)~\citep{somepalli2021saint} is a transformer-based architecture specifically designed for tabular data. It applies attention in two orthogonal directions:
(1) Column-wise attention, which models interactions across input features within a single sample (akin to standard transformers for NLP or vision), and
(2) Row-wise (inter-sample) attention, which models relationships across samples within a batch—allowing each sample to attend to others, conditioned on positional encoding and attention masks.
This hybrid mechanism allows SAINT to learn both vertical (feature-level) and horizontal (sample-level) dependencies, making it suitable for datasets where local context across rows or global input feature relationships are relevant.
However, inter-sample attention increases the computational complexity, especially for large batches or datasets, and introduces training instability under weak signal or sparse supervision.

We use the original implementation from the authors and adopt the hyperparameter tuning ranges from \citep{somepalli2021saint,salinas2023tabrepo}.

\begin{table}[h!]
    \centering
    \caption{Hyperparameter search space of SAINT (* indicates newly added hyperparameters in this study.)}
    \label{hpspace:saint}
    \resizebox{0.65\textwidth}{!}{
    \begin{tabular}{cl}
    \toprule
    Hyperparameter & Search space \\\midrule
    attn\_dropout & Uniform(0, 0.3) \\ 
    ff\_dropout & Uniform(0, 0.8) \\
    final\_mlp\_style & UniformCat([common, sep]) \\
    activation & UniformCat([reglu, geglu, sigmoid, relu]) \\
    learning\_rate & Uniform(1$e^{-5}$, 1$e^{-3}$) \\
    lr\_scheduler$^*$ & UniformCat([True, False]) \\
    weight\_decay$^*$ & Uniform(1$e^{-6}$, 1$e^{-3}$) \\
    n\_epochs & 100 \\
    optimizer$^*$ & UniformCat([AdamW, Adam, SGD]) \\\bottomrule
    \end{tabular}
    }
\end{table}

\clearpage

\section{Additional Results and Discussion}
\label{appendix:additionalresults}

This section provides extended results and analysis that could not be included in the main paper due to space limitations. These include supplementary metrics, finer-grained evaluations, and detailed findings on Spearman correlation analysis and comparison with TabPFN. While our main analysis focuses on seven dataset characteristics and normalized performance metrics, this appendix expands the discussion to cover additional insights into model behavior, robustness, and efficiency. All results, including raw scores, supplementary indicators, and complete metadata, are publicly available as part of the \textsc{MultiTab} benchmark suite at \url{https://huggingface.co/datasets/LGAI-DILab/Multitab}.

\subsection{Results with secondary metrics}
\label{appendix:additionalresults-metrics}

To complement the primary evaluation based on normalized predictive error (log loss for classification and RMSE for regression), we report additional results using secondary performance metrics--accuracy for classification and average rank across datasets. While these metrics are not used in the core analysis due to their sensitivity to dataset scale and label imbalance, they offer additional interpretability and a complementary perspective on model behavior across conditions.

\begin{figure}[h!]
    \includegraphics[width=\textwidth]{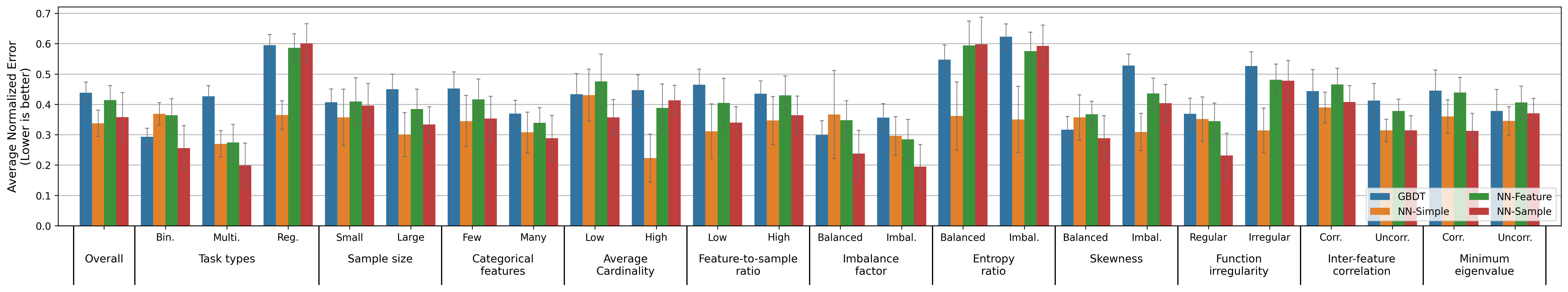}
    \caption{Average normalized error across model families and dataset sub-categories, computed based on (1 - accuracy) instead of log loss for classification tasks. Results are aggregated over all datasets and splits. Lower is better.}
    \label{fig:overview_accuracy}
\end{figure}

\begin{figure}[h!]
    \includegraphics[width=\textwidth]{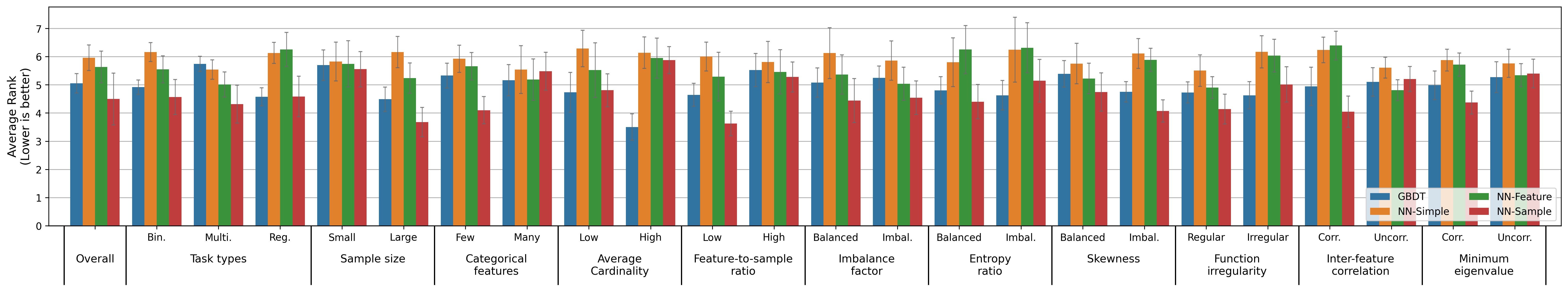}
    \caption{Average rank across model families and dataset sub-categories, computed using raw log loss and RMSE. Lower is better.}
    \label{fig:overview_rank}
\end{figure}

Figure~\ref{fig:overview_accuracy} shows the model-class-level results based on accuracy and RMSE, computed over all splits and datasets grouped by each sub-category. Figure~\ref{fig:overview_rank} presents the average rank per model family based on log loss and RMSE, providing an ordinal summary of model performance across the benchmark. These results largely align with the trends observed using normalized error and reinforce the comparative strengths of each architectural paradigm.

\subsection{Comparison between individual algorithms}
\label{appendix:additionalresults-algorithms}

Here we report comparisons among the top-performing individual algorithms: XGBoost, FT-Transformer, T2G-Former, and ModernNCA. These models consistently rank high across various dataset conditions and represent different architectural paradigms, including tree ensembles and neural networks with feature- or sample-aware components.

\begin{figure}[h!]
    \centering
    \includegraphics[width=\textwidth]{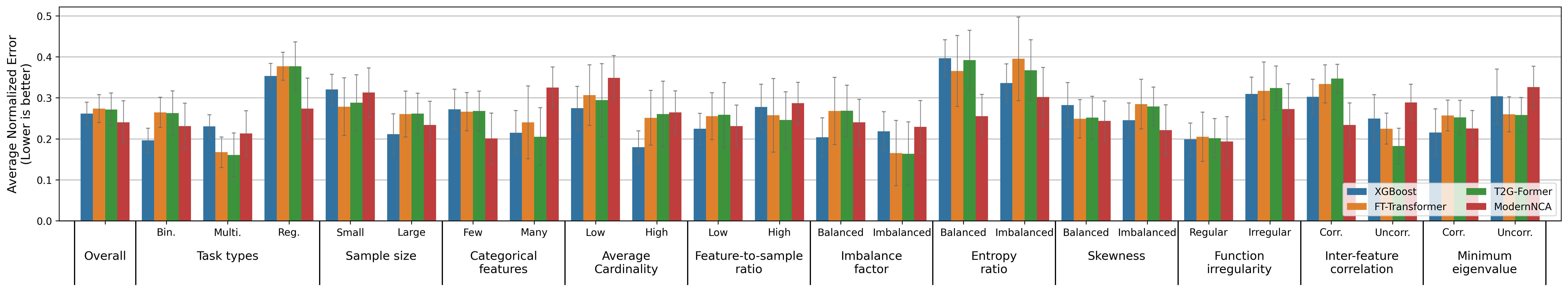}
    \caption{Normalized error of top-performing algorithms across dataset sub-categories. Lower is better.}
    \label{fig:overview_topmodels}
\end{figure}

\begin{figure}[h!]
    \centering
    \includegraphics[width=\textwidth]{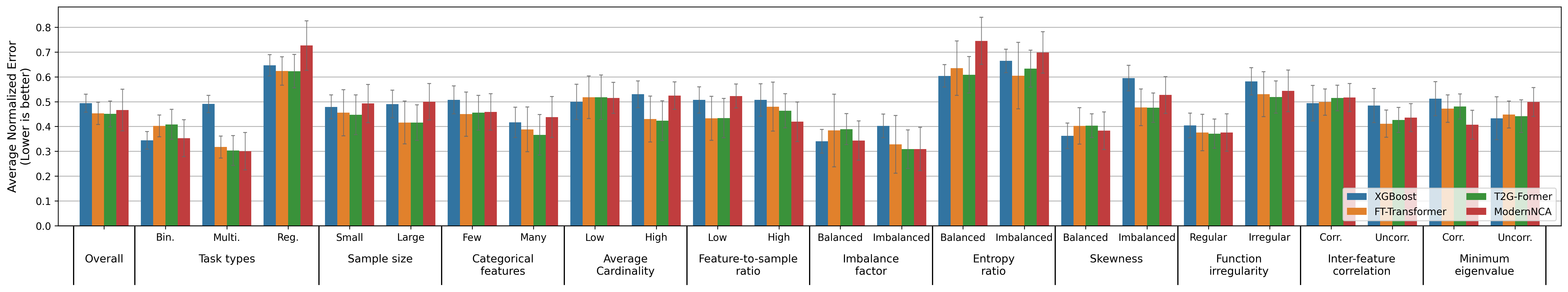}
    \caption{Normalized error of top-performing algorithms based on accuracy for classification tasks. Lower is better.}
    \label{fig:overview_accuracy_topmodels}
\end{figure}

\begin{figure}[h!]
    \centering
    \includegraphics[width=\textwidth]{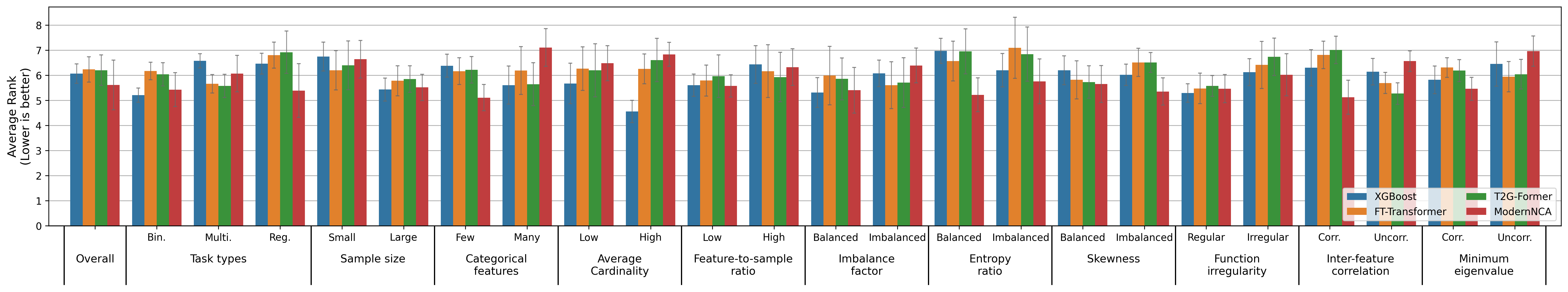}
    \caption{Average rank of top-performing algorithms across sub-categories. Lower is better.}
    \label{fig:overview_rank_topmodels}
\end{figure}

Figures~\ref{fig:overview_topmodels}, \ref{fig:overview_accuracy_topmodels}, and \ref{fig:overview_rank_topmodels} present the performance of these algorithms across all sub-categories. The plots follow the same structure as our main results.

The comparison highlights the tradeoffs among these models. For example, ModernNCA demonstrates strong performance under class imbalance and inter-sample-structured data, whereas FT-Transformer shows strength in high-cardinality and high-dimensional regimes. XGBoost remains a robust baseline across nearly all settings. 

\subsection{Correlation Between Dataset Statistics and Model Performance}
\label{appendix:additionalresults-correlation}

To understand continuous trends beyond discrete subgroups, we compute Spearman correlations between dataset statistics and model error, as summarized in Figure~\ref{fig:spearman} in the main text. Red cells denote performance degradation as the statistic increases; blue cells indicate improvement; and blank cells indicate no statistically significant correlation ($p \geq 0.05$).

\paragraph{Scale-related patterns.}
We observe that GBDTs, particularly CatBoost and XGBoost, exhibit strong negative correlations with sample size, confirming their ability to benefit from large datasets. 
Similarly, both models show positive correlations with the feature-to-sample ratio, reflecting their effectiveness in high-sample regimes where split decisions are statistically reliable.
In contrast, most neural models are relatively insensitive to sample size, with the notable exception of NN-Sample architectures such as TabR and ModernNCA. These models exhibit similarly strong negative correlations with sample size, suggesting that a large number of training examples is also crucial for learning robust inter-sample dependencies. Taken together, these findings highlight that sufficient data is particularly beneficial for models that rely on either decision branching (\emph{e.g.}, GBDTs) or learning sample-level relationships (\emph{e.g.}, NN-Sample)

\paragraph{Sensitivity to label imbalance and structural irregularity.}
Several neural models—including MLP variants, ResNet, and attention-based architectures like FT-Transformer and T2G-Former—show significant positive correlations between predictive error and entropy ratio, function irregularity, and inter-feature correlation. These patterns suggest that neural models are more sensitive to class skew, irregular label space, or entangled feature spaces. Such settings may exacerbate optimization difficulties or reduce the reliability of attention and similarity-based mechanisms. In contrast, GBDTs show little to no sensitivity to these dataset properties, reflecting their robustness to structural complexity.

\paragraph{Complementary role of correlation analysis.}
While correlation analysis has limitations--it can be sensitive to outliers, collinearity, and metric design--it nonetheless offers a complementary view to our subgroup-based evaluation. In particular, several trends observed in Figure~\ref{fig:corr} align with subgroup-level results, reinforcing key interpretations about model sensitivity to label imbalance and feature interaction. When used alongside structured partitioning, correlation analysis can help reveal global patterns that motivate or support more targeted hypotheses about model behavior under varying data conditions.

\subsection{Additional Analysis on Comparison with TabPFN}
\label{appendix:additionalresults-tabpfn}

\begin{figure}[h!]
    \centering
    \includegraphics[width=0.5\linewidth]{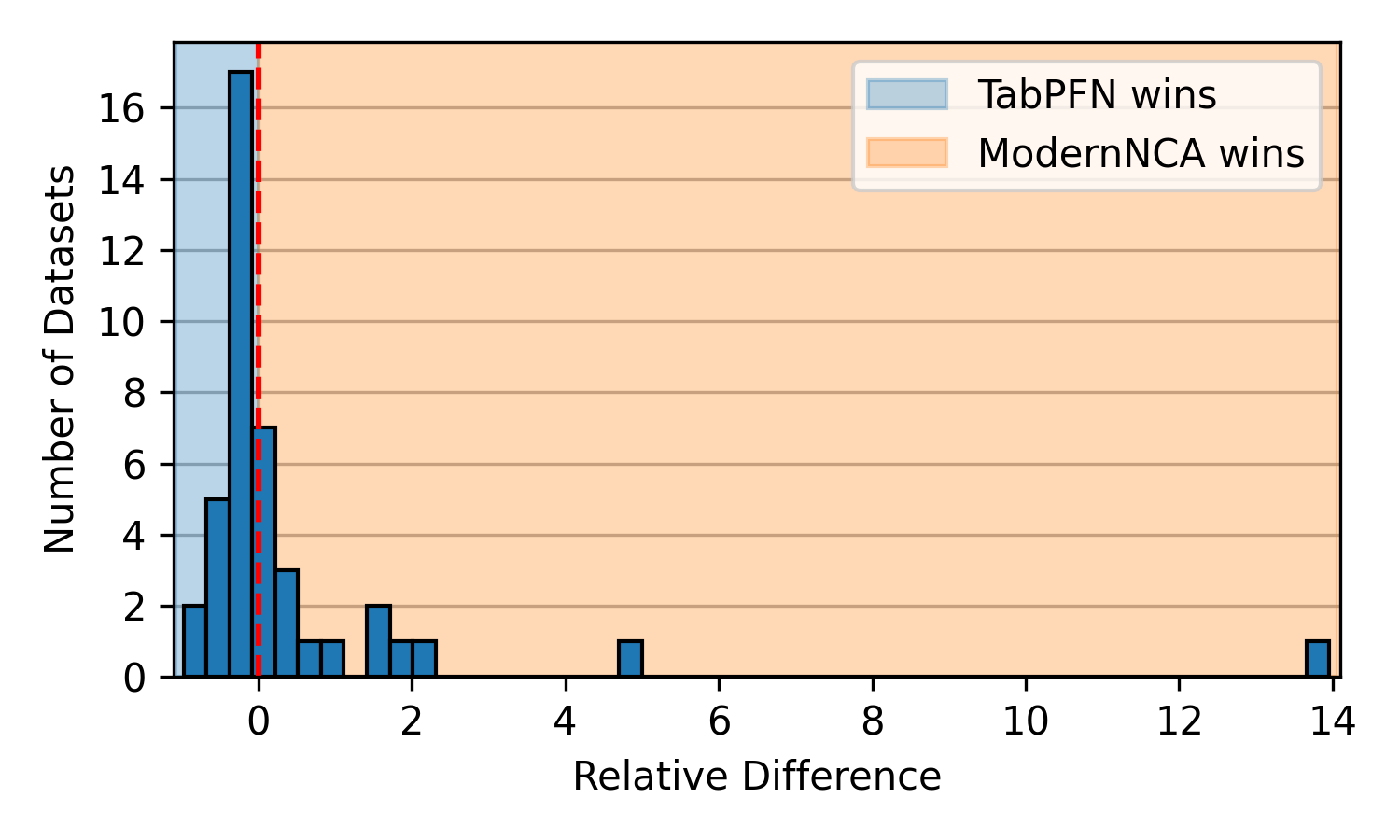}
    \caption{Relative performance comparison between TabPFN and ModernNCA across 42 datasets where both models are applicable. The $x$-axis shows the relative difference in error, with negative values indicating TabPFN wins. 
    TabPFN outperforms ModernNCA on 29 datasets (blue region), while ModernNCA performs better on 13 (orange region). 
    Although TabPFN wins more often, the margin is small in most cases.}
    \label{fig:tabpfn}
\end{figure}

TabPFN~\citep{hollmann2022tabpfn} is a pretrained transformer designed for classification on small tabular datasets via meta-learning. While promising in low-data regimes, it operates under strict architectural constraints: it supports only classification tasks with no more than 3,000 training samples, up to 1,000 input features, and a maximum of 10 classes. (While our analysis is based on TabPFN v1, we note that v2~\citep{hollmann2025accurate} introduces efficiency improvements but retains similar input constraints, and thus we expect comparable results.) As a result, in our benchmark, only 42 out of 196 datasets are eligible for TabPFN evaluation.

Because TabPFN is only applicable to a small subset of datasets, normalized error is not meaningful for comparison. Instead, we compare its performance to ModernNCA, one of the top-performing models in our benchmark, based on raw log loss and RMSE.

As shown in Figure~\ref{fig:tabpfn}, TabPFN outperforms ModernNCA on 29 out of 42 datasets, while ModernNCA achieves better performance on the remaining 13. The figure reports the relative error difference, computed as $\frac{\text{TabPFN error} - \text{ModernNCA error}}{\text{ModernNCA error}}$ where the error metric is log loss for classification and RMSE for regression. 
These results indicate that within its constrained operational domain, TabPFN is competitive but not consistently superior. In particular, when TabPFN performs better, the relative gains are typically small, where as ModernNCA often achieves larger improvements when it outperforms TabPFN.

To better understand when TabPFN has a relative advantage, we compare dataset characteristics across the two model groups. A Welch’s $t$-test reveals that only one dataset axis—entropy ratio—shows a statistically significant difference ($p < 0.05$). The average entropy ratio for datasets where TabPFN performs better is $0.550$, while it is $0.877$ for those where ModernNCA excels. This indicates that TabPFN is particularly effective on imbalanced classification datasets, possibly due to its training on synthetic tasks with class skew during pretraining. For all other dataset characteristics (\emph{e.g.}, feature-to-sample ratio, inter-feature correlation, function irregularity), no significant difference was observed between the two groups.

These results suggest that while TabPFN is an appealing solution for imbalanced and small-scale classification tasks, its limited applicability and modest gains call for future work on more generalizable pretrained tabular models. Additionally, the observed strength of ModernNCA across a wider range of settings highlights the benefits of architectures that can adaptively capture latent similarity across samples, even without pretraining.

\subsection{Discussion on Computational Costs}

To assess the computational overhead of each model class in our benchmark, we analyze three aspects: (1) average training time per model per model-hyperparameter combination, (2) total optimization time required to reach the best hyperparameter configuration, and (3) average number of trials required to reach the optimal configuration.
In all three figures, the horizontal axis represents the product of sample size and feature dimensionality (in log scale), used as a unified scale factor to reflect dataset complexity. For each model, the legend reports the average value across all evaluated datasets over the entire complexity range.

\begin{figure}[h]
    \centering
    \includegraphics[width=0.6\linewidth]{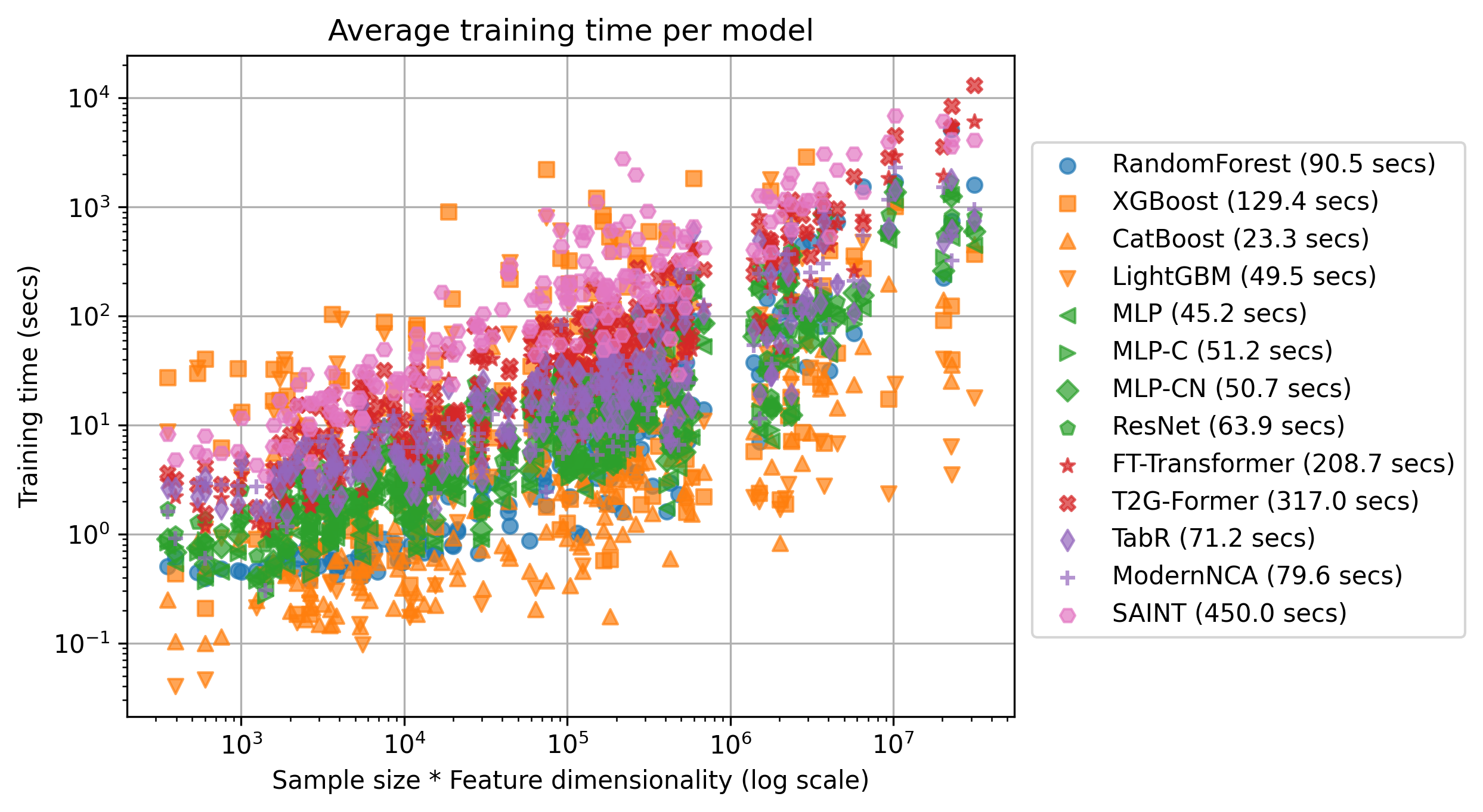}
    \caption{Average training time per model}
    \label{fig:cost-reproduce}
\end{figure}

\paragraph{Training time per model.} Figure~\ref{fig:cost-reproduce} shows the average training time for each model measured using its best-found hyperparameter configuration. 
While most models exhibit a sub-linear increase in training time relative to dataset complexity, attention-based architectures such as {SAINT} (450.0s), {T2G-Former} (317.0s), and {FT-Transformer} (208.7s) incur substantially higher per-trial costs. In contrast, tree-based models like {CatBoost} (23.3s) and {LightGBM} (49.5s) remain computationally efficient across the spectrum. Interestingly, high-performing neural models such as {TabR} (71.2s) and {ModernNCA} (79.6s) exhibit training times that are only moderately higher than classical baselines or GBDTs. Considering that many GBDT implementations do not fully support parallel training across hyperparameter configurations, these neural models can be seen as offering competitive—if not more efficient—training times under our standardized setting.

\begin{figure}[h]
    \centering
    \includegraphics[width=0.6\linewidth]{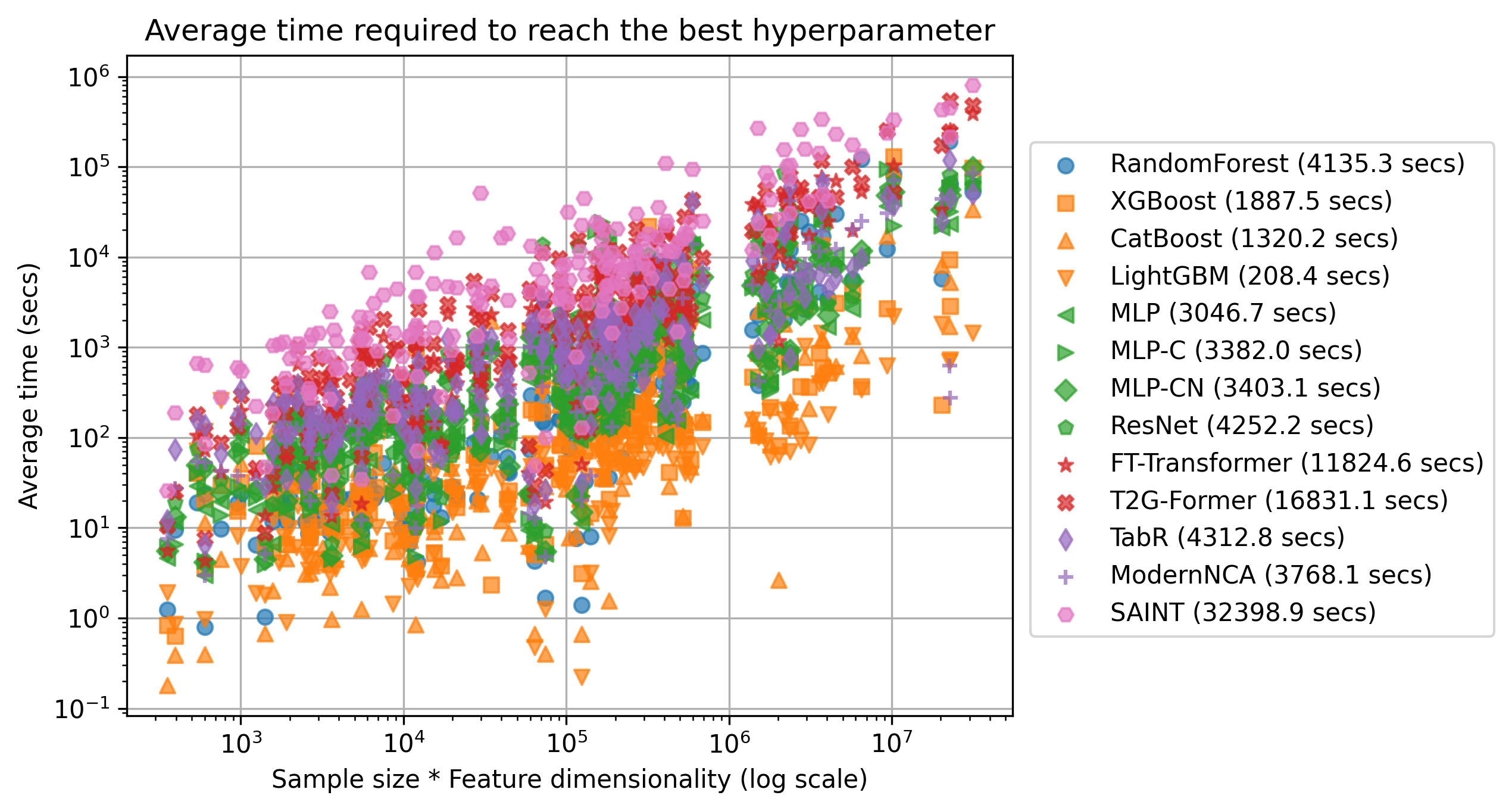}
    \caption{Average time required to reach the best hyperparameter}
    \label{fig:cost-avgtime}
\end{figure}

\paragraph{Time to best configuration.} 
Figure~\ref{fig:cost-avgtime} illustrates the average time required to reach the best hyperparameter configuration for each dataset-model pair. 
All models were tuned using a fixed budget of 100 trials; the reported values represent the cumulative time taken to reach the best-performing configuration within those trials, as described in Section~\ref{sec:benchmark}. 
A clear disparity emerges across model architectures. 
GBDTs such as XGBoost (1887.5s) and LightGBM (208.4s) tend to reach their best configurations relatively early. 
In contrast, neural models like T2G-Former (16,831.1s), SAINT (32,398.9s), and FT-Transformer (11,824.6s) generally require substantially more time to converge to their optimal settings, despite being trained under identical computational conditions. 
Notably, NN-Sample models such as TabR (4312.8s) and ModernNCA (3768.1s) offer a favorable trade-off: they deliver strong performance while converging significantly faster than other deep architectures. 
These observations highlight the practical efficiency of certain neural designs under constrained tuning budgets, and support our benchmark design choice to standardize optimization effort via a fixed trial count rather than time-based early stopping.

\begin{figure}[h]
    \centering
    \includegraphics[width=0.6\linewidth]{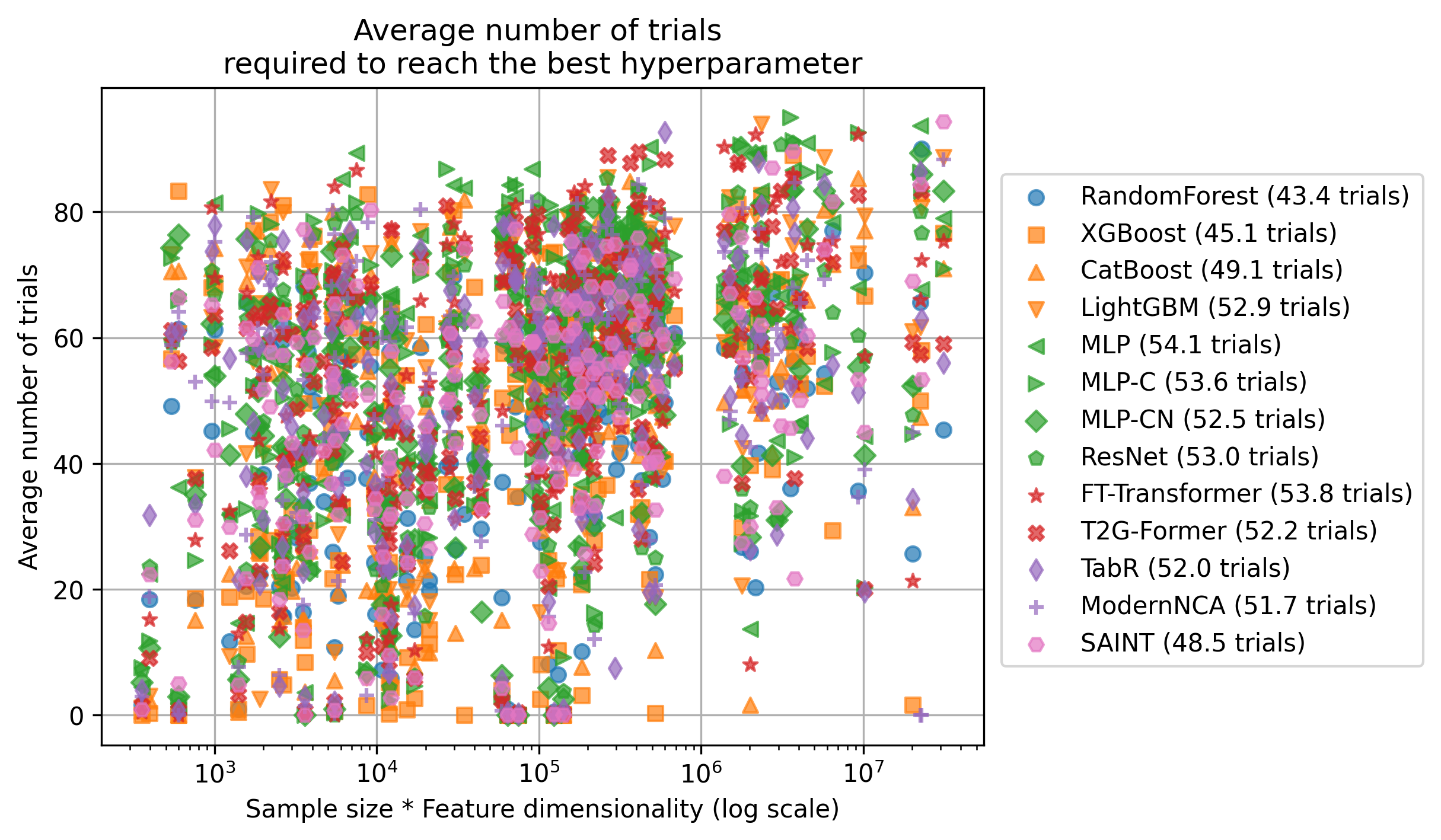}
    \caption{Average trial required to reach the best hyperparameter}
    \label{fig:cost-avgtrial}
\end{figure}

\paragraph{Trials to best configuration.}
Figure~\ref{fig:cost-avgtrial} shows the average number of trials required to reach the best-performing configuration per model. 
Despite differences in architectural complexity, most models converged within a narrow range of 50–55 trials on average. 
For example, MLP-C (53.6), FT-Transformer (53.8), and ResNet (53.0) required a similar number of trials as lighter-weight baselines like LightGBM (52.9) and CatBoost (49.1).
This consistency across models and data regimes suggests that at least 50 trials are generally necessary to reliably identify strong hyperparameter configurations.
These results reinforce our decision to fix the trial budget across models, rather than relying on time-based early stopping, thereby ensuring fair and sufficient optimization effort for all methods.

\paragraph{Most influential hyperparameters.}
To better understand which hyperparameters most strongly influence model performance, we computed feature importance scores using Optuna’s built-in analysis tools. For each model and dataset, we trained a surrogate regressor on the observed hyperparameter-performance pairs and extracted the relative importance of each parameter using the Mean Decrease Impurity method. We repeated this process across multiple seeds and datasets, then aggregated the results to compute (1) the average importance score per parameter and (2) the frequency with which each parameter ranked as the most important.

Among tree-based models, \textit{RandomForest} exhibited a clear dominance of \texttt{min\_samples\_leaf} (87.9\% top-1 frequency), indicating that limiting leaf size plays a central role in controlling overfitting. For \textit{GBDT models} (CatBoost, XGBoost, LightGBM), the most influential hyperparameters were split between tree structure parameters (\texttt{max\_depth}, \texttt{grow\_policy}) and gradient-related controls (\texttt{learning\_rate}, \texttt{colsample\_bytree}). No single parameter overwhelmingly dominated across all GBDT variants, reflecting the greater interaction between tuning knobs in these models.

In the {NN-Simple} group, all MLP variants (MLP, MLP-C, MLP-CN) consistently ranked \texttt{normalization} and \texttt{depth} as the two most important hyperparameters. These parameters together account for over 50\% of top-1 frequency in most MLP models, confirming their importance in managing both representational capacity and training stability.

For {NN-Feature} models (FT-Transformer, T2G-Former), \texttt{optimizer} emerged as the most critical hyperparameter (top-1 frequency $>$ 45\%), followed by \texttt{learning\_rate}. Despite their architectural complexity, structural hyperparameters such as \texttt{d\_token}, \texttt{d\_ffn\_factor}, or \texttt{n\_layers} showed relatively low importance. This indicates that training dynamics, rather than architectural variants, are more decisive for performance.

In the {NN-Sample} group, both {TabR} and {ModernNCA} emphasized \texttt{lr} as the dominant factor, followed closely by \texttt{frequency\_scale}, which is tied to frequency-based encoding. These models showed relatively low sensitivity to architectural depth or width parameters, reinforcing the idea that learning dynamics and signal transformation scales are more critical than structural complexity.

Finally, \textit{SAINT} also prioritized \texttt{learning\_rate} (57.6\%) and \texttt{optimizer} (29.8\%), with structural hyperparameters such as \texttt{activation}, \texttt{ff\_dropout}, or \texttt{final\_mlp\_style} contributing minimally. This supports the view that even in attention-rich hybrid architectures, optimization configuration remains the key to performance.

\begin{table}[h]
\centering
\caption{Hyperparameter importance for RandomForest}
\begin{tabular}{llcc}
\toprule
Model & Hyperparameter & Average feature importance & Top-1 frequency (\%) \\\midrule
RandomForest & min\_samples\_leaf & 0.7681 & 87.89 \\
RandomForest & max\_features & 0.1314 & 7.98 \\
RandomForest & max\_leaf\_nodes & 0.1004 & 4.14 \\\bottomrule
\end{tabular}
\end{table}

\begin{table}[h]
\centering
\caption{Hyperparameter importance for XGBoost}
\begin{tabular}{llcc}
\toprule
Model & Hyperparameter & Average feature importance & Top-1 frequency (\%) \\\midrule
XGBoost & max\_depth & 0.3270 & 40.05 \\
XGBoost & colsample\_bytree & 0.3254 & 38.79 \\
XGBoost & min\_child\_weight & 0.1651 & 12.17 \\
XGBoost & learning\_rate & 0.1643 & 8.75 \\
XGBoost & enable\_category & 0.0182 & 0.24 \\\bottomrule
\end{tabular}
\end{table}

\begin{table}[h]
\centering
\caption{Hyperparameter importance for CatBoost}
\begin{tabular}{llcc}
\toprule
Model & Hyperparameter & Average feature importance & Top-1 frequency (\%) \\\midrule
CatBoost & learning\_rate & 0.3536 & 48.84 \\
CatBoost & max\_depth & 0.2554 & 27.89 \\
CatBoost & l2\_leaf\_reg & 0.1542 & 8.43 \\
CatBoost & grow\_policy & 0.1062 & 8.25 \\
CatBoost & one\_hot\_max\_size & 0.0839 & 5.88 \\
CatBoost & max\_ctr\_complexity & 0.0468 & 0.71 \\\bottomrule
\end{tabular}
\end{table}

\begin{table}[h]
\centering
\caption{Hyperparameter importance for LightGBM}
\begin{tabular}{llcc}
\toprule
Model & Hyperparameter & Average feature importance & Top-1 frequency (\%) \\\midrule
LightGBM & learning\_rate & 0.5172 & 64.86 \\
LightGBM & extra\_trees & 0.1596 & 16.04 \\
LightGBM & min\_data\_in\_leaf & 0.1558 & 13.86 \\
LightGBM & feature\_fraction & 0.1000 & 3.83 \\
LightGBM & num\_leaves & 0.0674 & 1.42 \\\bottomrule
\end{tabular}
\end{table}

\begin{table}[h]
\centering
\caption{Hyperparameter importance for MLP}
\begin{tabular}{llcc}
\toprule
Model & Hyperparameter & Average feature importance & Top-1 frequency (\%) \\\midrule
MLP & normalization & 0.1815 & 27.91 \\
MLP & depth & 0.1643 & 23.41 \\
MLP & width & 0.1170 & 8.89 \\
MLP & optimizer & 0.1159 & 11.82 \\
MLP & activation & 0.1098 & 7.78 \\
MLP & learning\_rate & 0.1073 & 8.54 \\
MLP & dropout & 0.0978 & 6.73 \\
MLP & weight\_decay & 0.0880 & 4.86 \\
MLP & lr\_scheduler & 0.0183 & 0.06 \\\bottomrule
\end{tabular}
\end{table}

\begin{table}[h]
\centering
\caption{Hyperparameter importance for MLP-C}
\begin{tabular}{llcc}
\toprule
Model & Hyperparameter & Average feature importance & Top-1 frequency (\%) \\\midrule
MLP-C & normalization & 0.1861 & 29.78 \\
MLP-C & depth & 0.1672 & 26.35 \\
MLP-C & activation & 0.1179 & 9.71 \\
MLP-C & width & 0.1026 & 7.05 \\
MLP-C & learning\_rate & 0.0960 & 7.05 \\
MLP-C & dropout & 0.0915 & 6.57 \\
MLP-C & d\_embedding & 0.0809 & 5.03 \\
MLP-C & weight\_decay & 0.0786 & 3.73 \\
MLP-C & optimizer & 0.0630 & 4.44 \\
MLP-C & lr\_scheduler & 0.0162 & 0.30 \\\bottomrule
\end{tabular}
\end{table}

\begin{table}[h]
\centering
\caption{Hyperparameter importance for MLP-CN}
\begin{tabular}{llcc}
\toprule
Model & Hyperparameter & Average feature importance & Top-1 frequency (\%) \\\midrule
MLP-CN & depth & 0.1771 & 33.20 \\
MLP-CN & normalization & 0.1238 & 17.87 \\
MLP-CN & learning\_rate & 0.1001 & 8.08 \\
MLP-CN & width & 0.0966 & 7.49 \\
MLP-CN & dropout & 0.0876 & 6.37 \\
MLP-CN & d\_embedding\_cat & 0.0863 & 5.96 \\
MLP-CN & d\_embedding\_num & 0.0841 & 4.83 \\
MLP-CN & weight\_decay & 0.0821 & 5.13 \\
MLP-CN & optimizer & 0.0763 & 7.13 \\
MLP-CN & activation & 0.0656 & 3.24 \\
MLP-CN & lr\_scheduler & 0.0204 & 0.71 \\\bottomrule
\end{tabular}
\end{table}

\begin{table}[h]
\centering
\caption{Hyperparameter importance for ResNet}
\begin{tabular}{llcc}
\toprule
Model & Hyperparameter & Average feature importance & Top-1 frequency (\%) \\\midrule
ResNet & normalization & 0.1168 & 17.20 \\
ResNet & learning\_rate & 0.1167 & 14.17 \\
ResNet & hidden\_dropout & 0.1003 & 10.65 \\
ResNet & residual\_dropout & 0.0975 & 9.23 \\
ResNet & d & 0.0960 & 9.23 \\
ResNet & d\_embedding & 0.0921 & 9.11 \\
ResNet & d\_hidden\_factor & 0.0892 & 7.62 \\
ResNet & weight\_decay & 0.0875 & 7.98 \\
ResNet & activation & 0.0759 & 6.07 \\
ResNet & optimizer & 0.0680 & 6.25 \\
ResNet & n\_layers & 0.0430 & 2.08 \\
ResNet & lr\_scheduler & 0.0170 & 0.42 \\\bottomrule
\end{tabular}
\end{table}

\begin{table}[h]
\centering
\caption{Hyperparameter importance for FT-Transformer}
\begin{tabular}{llcc}
\toprule
Model & Hyperparameter & Average feature importance & Top-1 frequency (\%) \\\midrule
FT-Transformer & optimizer & 0.2845 & 51.38 \\
FT-Transformer & learning\_rate & 0.1159 & 12.63 \\
FT-Transformer & attention\_dropout & 0.0804 & 5.63 \\
FT-Transformer & ffn\_dropout & 0.0715 & 4.37 \\
FT-Transformer & d\_ffn\_factor & 0.0706 & 4.49 \\
FT-Transformer & residual\_dropout & 0.0701 & 4.19 \\
FT-Transformer & d\_token & 0.0688 & 4.55 \\
FT-Transformer & weight\_decay & 0.0660 & 3.29 \\
FT-Transformer & activation & 0.0658 & 5.39 \\
FT-Transformer & n\_layers & 0.0426 & 1.98 \\
FT-Transformer & token\_bias & 0.0218 & 1.50 \\
FT-Transformer & prenormalization & 0.0174 & 0.42 \\
FT-Transformer & lr\_scheduler & 0.0138 & 0.12 \\
FT-Transformer & initialization & 0.0106 & 0.06 \\\bottomrule
\end{tabular}
\end{table}

\begin{table}[h]
\centering
\caption{Hyperparameter importance for T2G-Former}
\begin{tabular}{llcc}
\toprule
Model & Hyperparameter & Average feature importance & Top-1 frequency (\%) \\\midrule
T2G-Former & optimizer & 0.2529 & 45.59 \\
T2G-Former & learning\_rate & 0.1584 & 21.76 \\
T2G-Former & attention\_dropout & 0.0750 & 5.91 \\
T2G-Former & ffn\_dropout & 0.0657 & 3.96 \\
T2G-Former & residual\_dropout & 0.0657 & 3.73 \\
T2G-Former & d\_token & 0.0652 & 3.78 \\
T2G-Former & d\_ffn\_factor & 0.0618 & 3.49 \\
T2G-Former & weight\_decay & 0.0608 & 3.13 \\
T2G-Former & learning\_rate\_embed & 0.0592 & 3.13 \\
T2G-Former & activation & 0.0519 & 2.42 \\
T2G-Former & n\_layers & 0.0461 & 1.66 \\
T2G-Former & prenormalization & 0.0159 & 0.89 \\
T2G-Former & lr\_scheduler & 0.0119 & 0.18 \\
T2G-Former & initialization & 0.0094 & 0.35 \\\bottomrule
\end{tabular}
\end{table}

\begin{table}[h]
\centering
\caption{Hyperparameter importance for TabR}
\begin{tabular}{llcc}
\toprule
Model & Hyperparameter & Average feature importance & Top-1 frequency (\%) \\\midrule
TabR & lr & 0.2817 & 45.21 \\
TabR & frequency\_scale & 0.2103 & 30.99 \\
TabR & dropout0 & 0.0798 & 4.64 \\
TabR & context\_dropout & 0.0749 & 3.54 \\
TabR & d\_main & 0.0735 & 2.73 \\
TabR & weight\_decay & 0.0698 & 3.19 \\
TabR & n\_frequencies & 0.0657 & 2.15 \\
TabR & d\_embedding & 0.0606 & 1.68 \\
TabR & encoder\_n\_blocks & 0.0579 & 5.11 \\
TabR & lr\_scheduler & 0.0154 & 0.52 \\
TabR & predictor\_n\_blocks & 0.0106 & 0.23 \\\bottomrule
\end{tabular}
\end{table}

\begin{table}[h]
\centering
\caption{Hyperparameter importance for ModernNCA}
\begin{tabular}{llcc}
\toprule
Model & Hyperparameter & Average feature importance & Top-1 frequency (\%) \\\midrule
ModernNCA & lr & 0.3026 & 49.42 \\
ModernNCA & frequency\_scale & 0.1721 & 21.23 \\
ModernNCA & n\_blocks & 0.1048 & 10.58 \\
ModernNCA & dim & 0.0804 & 3.92 \\
ModernNCA & dropout & 0.0752 & 3.68 \\
ModernNCA & d\_block & 0.0687 & 2.63 \\
ModernNCA & weight\_decay & 0.0663 & 3.10 \\
ModernNCA & n\_frequencies & 0.0610 & 2.63 \\
ModernNCA & d\_embedding & 0.0582 & 1.93 \\
ModernNCA & lr\_scheduler & 0.0155 & 0.88 \\\bottomrule
\end{tabular}
\end{table}

\begin{table}[h]
\centering
\caption{Hyperparameter importance for SAINT}
\begin{tabular}{llcc}
\toprule
Model & Hyperparameter & Average feature importance & Top-1 frequency (\%) \\\midrule
SAINT & learning\_rate & 0.3744 & 57.59 \\
SAINT & optimizer & 0.2661 & 29.84 \\
SAINT & weight\_decay & 0.1240 & 5.60 \\
SAINT & attention\_dropout & 0.1191 & 3.10 \\
SAINT & ff\_dropout & 0.0942 & 1.31 \\
SAINT & attn\_dropout & 0.0828 & 0.83 \\
SAINT & activation & 0.0421 & 0.71 \\
SAINT & lr\_scheduler & 0.0314 & 0.89 \\
SAINT & final\_mlp\_style & 0.0171 & 0.12 \\\bottomrule
\end{tabular}
\end{table}

\clearpage 

\section{Limitations of \textsc{MultiTab}}
\label{appendix:limitations}

While \textsc{MultiTab} provides a comprehensive benchmark for analyzing tabular model performance across diverse datasets and data characteristics, it is important to acknowledge several limitations of the current study:

\begin{itemize}[leftmargin=*]
    \item \textbf{Finite dataset coverage.} Our findings are derived from experiments on a finite collection of 196 datasets. Although care was taken to ensure diversity in data domains and properties, conclusions drawn may not generalize to all possible tabular datasets in the world.
    \item \textbf{Heuristic sub-category thresholds.} Sub-categories for dataset characteristics were defined using empirical thresholds based on histogram distributions. While these thresholds reflect broad trends and were useful for structured evaluation, they remain heuristic and may not perfectly partition datasets along theoretically grounded axes.
    \item \textbf{Incomplete metric space.} We focused on eleven key metrics spanning seven dataset axes. However, other factors, such as temporal drift, feature sparsity, label noise, or task-specific priors, may also play significant roles in real-world tabular performance but are not explicitly captured in our current benchmark.
    \item \textbf{Algorithm and budget scope.} Although we include 13 widely used models spanning various architectural paradigms, our benchmark does not exhaustively cover the growing ecosystem of tabular algorithms. In addition, hyperparameter search spaces and tuning budgets, while extensive and standardized where possible, may still influence final performance in model-specific ways.
\end{itemize}

We hope future work will expand upon these aspects, incorporating more datasets, theoretical guidance for axis definition, and additional models or evaluation protocols to further enhance the benchmarking landscape. Our study demonstrates for the first time that multi-dimensional analysis and evaluation of tabular models is not only feasible but also highly informative. We believe this framework opens promising avenues for future research, where overcoming current limitations may lead to more precise, theory-grounded, and robust insights into the interplay between data characteristics and model design.

\end{document}